\documentclass[10pt,twocolumn,letterpaper]{article}

\usepackage[pagenumbers]{wacv}
\usepackage[accsupp]{axessibility}

\usepackage{graphicx}
\usepackage{subcaption}
\usepackage{booktabs}
\usepackage{siunitx}
\usepackage{multirow}
\usepackage{array}
\usepackage{amsmath}
\usepackage{amssymb}

\graphicspath{{figures/}}

\newcommand{\ccOffPsnrA}{15.869} \newcommand{\ccOffSsimA}{0.771}  \newcommand{\ccOffLpipsA}{0.299}
\newcommand{\ccOnPsnrA}{15.976}  \newcommand{\ccOnSsimA}{0.785}   \newcommand{\ccOnLpipsA}{0.288}
\newcommand{\ccOffPsnrB}{17.115} \newcommand{\ccOffSsimB}{0.796}  \newcommand{\ccOffLpipsB}{0.277}
\newcommand{\ccOnPsnrB}{17.174}  \newcommand{\ccOnSsimB}{0.804}   \newcommand{\ccOnLpipsB}{0.270}
\newcommand{\ccSeamBeforeA}{0.159} \newcommand{\ccSeamAfterA}{0.101}
\newcommand{\ccSeamBeforeB}{0.151} \newcommand{\ccSeamAfterB}{0.103}
\newcommand{\ccAddedMs}{2.3}   \newcommand{\ccTotalMs}{13.1}

\usepackage{tikz}
\usetikzlibrary{arrows.meta,calc}

\definecolor{wacvblue}{rgb}{0.21,0.49,0.74}
\usepackage[pagebackref,breaklinks,colorlinks,allcolors=wacvblue]{hyperref}

\usepackage[capitalize]{cleveref}
\crefname{section}{Sec.}{Secs.}
\crefname{figure}{Fig.}{Figs.}
\crefname{table}{Tab.}{Tabs.}
\Crefname{section}{Section}{Sections}
\Crefname{figure}{Figure}{Figures}
\Crefname{table}{Table}{Tables}

\def\wacvPaperID{322}
\def\confName{WACV}
\def\confYear{2027}

\title{GPU-Native Real-Time Textured Mesh Reconstruction with\\ Seam-Aware Multi-View Radiometric Correction}

\author{Hiroyuki Sato\\
MinD in a Device Co., Ltd., Tokyo, Japan\\
{\tt\small h.sato@mindinadevice.com}
}

\begin{document}
\maketitle
\begin{abstract}
We present a GPU-native volumetric-video system for real-time reconstruction of
graphics-compatible textured triangle meshes from multiple RGB-D sensors.
Existing TSDF systems accelerate fusion and mesh extraction, but extending
them to per-frame textured meshing adds CPU mesh processing, CPU-side texturing
work, or extra CPU--GPU transfers between stages. After input staging, our
method keeps the reconstruction and
texturing stages (TSDF fusion, Marching Cubes, connected-component noise
removal, view-based texture mapping, inpainting, and radiometric seam
correction) resident on the GPU; no intermediate mesh or image data returns to
the host between these stages. A
fixed-size-array representation of mesh topology gives constant-time neighbor
access on the GPU and makes the downstream mesh-level
operations data-parallel. Because we texture each face from a single camera,
exposure seams that vary smoothly across each image remain; we estimate a per-camera
low-frequency gain field and bake it into the atlas on the GPU. On the CMU
Panoptic dataset, our method is faster
than the baselines at every tested voxel
resolution and number of sensors, achieves peak speedups of more than an order of magnitude
over CUDA-accelerated nvblox and CPU-based Open3D, and approaches an offline
textured-meshing baseline in
perceptual re-rendering quality.
\end{abstract}

\section{Introduction}
\label{sec:intro}

Volumetric video~\cite{collet2015high} enables telepresence and AR/VR
avatar generation, and its capture with multiple RGB-D sensors has been an
active area of research~\cite{sterzentsenko2018low,app131810199}. The dominant scene
representation in current systems is the colored point cloud, but its data
volume is large and, under realistic bandwidth budgets, real-time transmission
remains challenging~\cite{10.1145/3768981}. Neural representations such as
NeRF~\cite{mildenhall2020nerf} and 3D Gaussian
Splatting~\cite{kerbl3Dgaussians} achieve high-quality novel-view synthesis,
but their per-frame reconstruction cost and storage footprint hinder real-time
use. A textured triangle mesh provides a graphics-compatible representation in
which geometry and appearance budgets can be controlled separately: 
mesh simplification adjusts geometry, while atlas resolution and texture
compression adjust appearance. Commodity GPUs and standard graphics engines
support this representation natively through their rasterization pipelines.

\begin{figure}[t]
    \centering
    \includegraphics[width=.95\linewidth]{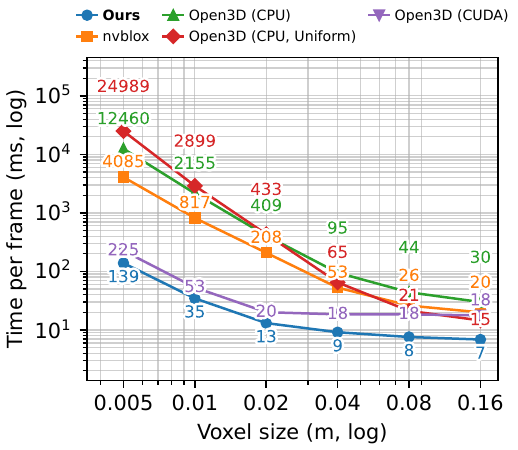}
    \caption{Per-frame processing time versus voxel size on the
    \textit{office2} sequence of CMU Panoptic~\cite{Joo_2015_ICCV,Joo_2017_TPAMI}
    with five Kinect inputs. Both axes use log scale. ``Ours'' includes input
    upload, preprocessing, and radiometric seam correction (\cref{sec:colorcorr}) and remains real-time at
    \SI{0.02}{\meter}. Open3D (CPU) is run without noise removal here and in
    \cref{fig:bench_kinect_vs_time}; \cref{tab:runtime_breakdown} includes it.}
    \label{fig:bench_voxel_vs_time}
\end{figure}

\begin{figure*}[t!]
  \centering
  \includegraphics[width=\textwidth]{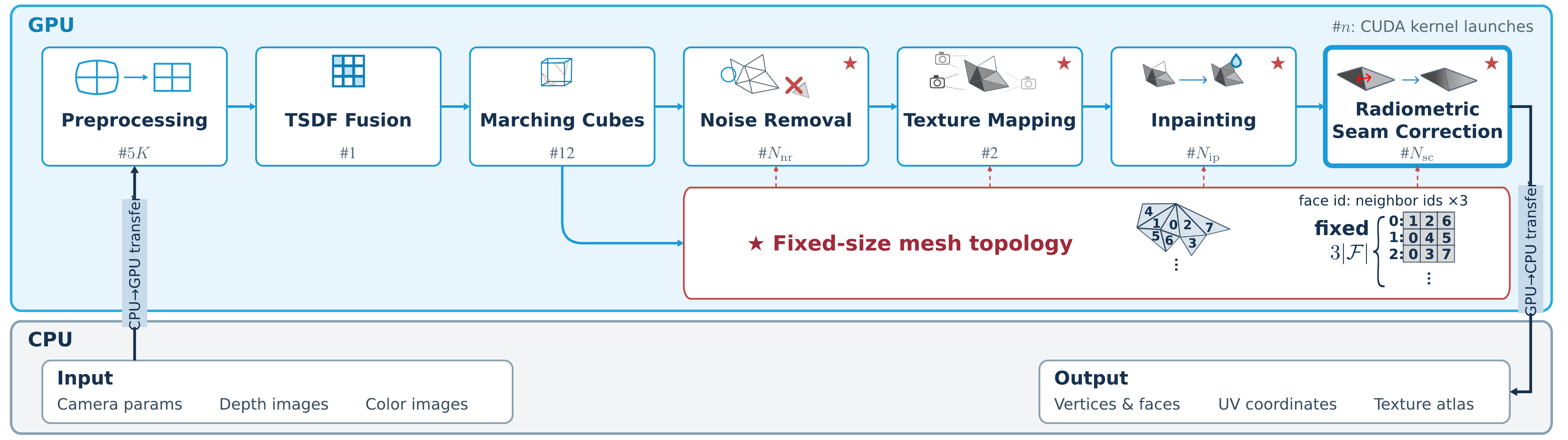}
  \caption{Our GPU-native textured-meshing pipeline. Each frame uploads the
  raw depth and color images in one batch; Preprocessing undistorts them and
  registers depth to color on the GPU. Apart from a few scalar counters and flags only the
  final mesh, UV coordinates, and corrected atlas return to the CPU. \#$n$ gives
  the main CUDA kernel launches per stage ($K$ cameras); $N_{\mathrm{nr}}$ and
  $N_{\mathrm{ip}}$ are the label-propagation and inpainting iterations (capped
  at $100$ and $5$) and $N_{\mathrm{sc}}=55$. Red stars mark the stages that build on the
  fixed-size-array mesh topology (\cref{sec:topology}); the seam correction
  (\cref{sec:colorcorr}) reuses this adjacency and writes the corrected atlas
  before output.}
  \label{fig:pipeline}
\end{figure*}

Producing this representation every frame from multiple RGB-D streams is costly.
Truncated Signed Distance Function (TSDF) fusion~\cite{curless1996volumetric}
is widely used for
3D reconstruction because it is robust and
parallel~\cite{newcombe2011kinectfusion,niessner2013real,10.1145/2461912.2461919,10.1145/3054739}.
GPU implementations such as nvblox~\cite{millane2024nvblox} and
Open3D~\cite{Zhou2018} accelerate TSDF fusion and Marching Cubes
(MC)~\cite{10.1145/37402.37422} based mesh extraction; Open3D's CPU pipeline
adds connected-component noise removal (NR). These systems do not support texture
mapping in our setting, and they do not provide the per-frame textured mesh
pipeline that volumetric video requires. Extending them to textured meshing adds
mesh filtering, texture processing, and host-device movement. Runtime then rises
sharply as voxel resolution increases (\cref{fig:bench_voxel_vs_time}), the
regime demanding high visual quality.

Real-time texturing raises an appearance challenge. When each face uses one
camera, adjacent faces may meet along visible exposure seams whose magnitude
varies smoothly across each camera's image.
Texture mapping methods reduce such seams by globally optimizing view labels,
colors, or texture alignment~\cite{lempitsky2007seamless,Waechter2014Texturing,fu2018texture},
but at offline cost. Real-time systems instead texture the scene online, fusing
images into progressively updated texture
representations~\cite{lee2020texturefusion,10.1145/3503926}
or adjusting patch colors against a reference~\cite{han2022real}. They scan
static scenes with one moving sensor, rather than reconstructing each dynamic
frame from simultaneous views and correcting their seams.

We target indoor volumetric video with high-density, high-frame-rate multi-view
RGB-D input. Our system
(\cref{fig:pipeline}) targets a bounded observation volume, and after one
batched input upload keeps the reconstruction and texturing stages (TSDF fusion, mesh
generation, noise removal, texture mapping, and radiometric seam correction) on
the GPU. At high resolution, data movement dominates fusion arithmetic, so GPU
residency lets per-frame textured meshing, topology-aware cleanup, inpainting,
and radiometric correction fit within the \SI{33}{\milli\second} budget of
\SI{30}{fps} capture. We make four contributions: (1)~a GPU-native
volumetric-video system that generates graphics-compatible textured meshes from
multiple RGB-D sensors while keeping mesh-level processing and texturing on the
GPU; (2)~a fixed-size-array representation of mesh topology that enables
efficient post-MC mesh processing on the GPU; (3)~GPU-resident texture
processing, including noise removal and inpainting, built on this
representation; and (4)~a real-time, on-GPU radiometric seam correction stage
that estimates a per-camera low-frequency gain field and bakes it into the
texture atlas, reducing inter-camera exposure seams, including mismatches
that vary across each camera's image.

As shown in \cref{fig:bench_voxel_vs_time}, our method runs
$89.7\times$/$29.4\times$ faster than Open3D (CPU)/nvblox at \SI{0.005}{\meter} voxels.
At \SI{0.02}{\meter} it runs in real time (\SI{13.1}{\milli\second} per frame
including input upload and preprocessing, within the
\SI{33}{\milli\second} budget of \SI{30}{fps} sensors), and its
reconstruction quality approaches the offline textured-meshing baseline in
LPIPS under held-out re-rendering. Code is available at
\url{https://github.com/unclearness/grtmesh}. Implementation details,
protocols, and further results are in the appendix, and the supplementary
video is on the project page, \url{https://unclearness.github.io/grtmesh/}.

\section{Related Work}
\label{sec:related}

\paragraph{TSDF-based real-time reconstruction.}
Volumetric TSDF fusion~\cite{curless1996volumetric} underlies dense
reconstruction, from real-time KinectFusion~\cite{newcombe2011kinectfusion},
voxel hashing~\cite{niessner2013real}, and
BundleFusion~\cite{10.1145/3054739} to offline scene
reconstruction~\cite{10.1145/2461912.2461919}. Our baselines
nvblox~\cite{millane2024nvblox} and Open3D~\cite{Zhou2018} implement TSDF fusion
and MC-based~\cite{10.1145/37402.37422} mesh extraction with per-vertex color
(Open3D adds CPU connected-component noise removal). They provide mapping
pipelines with per-vertex color rather than per-frame texture-mapped meshes;
Open3D integrates one RGB-D image at a time.

\paragraph{Multi-view RGB-D human and scene capture.}
For multiple RGB-D sensors, per-frame methods fuse the views without tracking,
by merging per-view meshes~\cite{alexiadis2012real} or by moving-least-squares
fusion of the point clouds~\cite{10897380}; another line performs non-rigid
fusion and tracking. Function4D~\cite{yu2021function4d} requires training on large datasets.
The single-sensor DoubleFusion~\cite{yu2018doublefusion} and
SparseFusion~\cite{9113759} further rely on the SMPL body template~\cite{SMPL:2015}.
Fusion4D~\cite{10.1145/2897824.2925969}, the engine behind the Holoportation
telepresence system~\cite{10.1145/2984511.2984517}, and its successor
Motion2Fusion~\cite{dou2017motion2fusion} are compelling but compute-heavy, and
an FFT-based method~\cite{alexiadis2016integrated}
scales worse than TSDF fusion with resolution.
Our method is object-agnostic and reconstructs everything visible
without foreground/background segmentation.

\paragraph{High-quality offline reconstruction and texturing.}
Screened Poisson Surface Reconstruction
(SPSR)~\cite{kazhdan2006poisson,kazhdan2013screened} produces high-quality
watertight meshes but requires a large sparse solve. Global texture mapping
optimizations~\cite{lempitsky2007seamless,Waechter2014Texturing,fu2018texture,PatchBasedTextureMapping}
yield seamless atlases by leveling colors along seams~\cite{lempitsky2007seamless},
combining global color adjustment with local blending~\cite{Waechter2014Texturing},
or correcting texture misalignment from global to local~\cite{fu2018texture}; related work jointly
optimizes appearance with pose, texture coordinates, or
lighting~\cite{zhou2014color,jeon2016texture,maier2017intrinsic3d}. These
methods, like motion-graph synthesis~\cite{prada2016motion} and spatiotemporal
atlas parameterization~\cite{prada2017spatiotemporal}, run offline.
We cast related view selection and seam leveling as per-face,
data-parallel GPU computations, with a per-camera low-frequency field baked into
the atlas (\cref{sec:colorcorr}).

\paragraph{Real-time online texturing.}
MeshReduce~\cite{10494107} produces real-time textured meshes through distributed
processing. TextureFusion~\cite{lee2020texturefusion} and
NormalFusion~\cite{ha2021normalfusion} store appearance in a
texture-tile voxel grid rather than a standard UV atlas, while
TextureMe~\cite{10.1145/3503926} maintains a conventional-texture-compatible but
fragmented atlas of right-triangular patches. All three update a \emph{static}
single-sensor scan online, registering each new frame to reduce
geometric or photometric misalignment. This temporal accumulation assumes a
persistent surface and does not fit per-frame meshes for dynamic volumetric
video. TextureFusion and TextureMe report degraded texture under severe exposure
change, view-dependent or specular lighting, and automatic white balance;
NormalFusion assumes Lambertian surfaces under static illumination. Han~\etal~\cite{han2022real} texture a globally consistent
reconstruction online with MRF view selection and CPU per-patch color transfer.
It is closest to ours, but adjusts each patch as a whole; we fit a per-camera
spatial field in fixed time on the GPU, avoiding CPU-side optimization.

\section{Method}
\label{sec:method}

\cref{fig:pipeline} shows our pipeline. We implement the host (CPU) side in C++
and the device (GPU) side in CUDA~\cite{cuda_programming_guide}. For each frame,
we upload the raw color and depth images in one batch; the GPU undistorts them,
registers depth to color, builds the color atlas, runs the reconstruction and
texturing stages, and writes the textured mesh back to the CPU. To minimize
synchronization and transfer overhead, we coalesce data where possible and use
pinned memory.

\subsection{TSDF Fusion and Mesh Generation}
\label{sec:tsdf}

To reconstruct a surface from a stream of depth images we use projective
TSDF-based volumetric fusion~\cite{newcombe2011kinectfusion}. For speed
the voxel grid uses a uniform (dense) layout rather than a hash- or
tree-based sparse structure. To process the multi-view depth input, we implement
the entire fusion step as a single CUDA kernel in which each thread owns one
voxel. The thread projects its grid point into each view's depth image in turn,
computes the corresponding signed distance, and integrates the per-view TSDF
contributions. Because each voxel is processed independently, the kernel needs no
atomic operations and reaches high parallel efficiency.

We then extract a triangle mesh
\(
\mathcal{M}=(\mathcal{V},\mathcal{F})
\)
from the TSDF volume with MC, where
\(
\mathcal{V}=\{\mathbf{v}_i \in \mathbb{R}^3\}
\)
is the set of vertex positions and
\(
\mathcal{F}=\{f_j=(v_{j0}, v_{j1}, v_{j2})\}
\)
is the set of triangular faces, each defined by three vertex indices. Running MC
in parallel on the GPU requires determining the output size in advance and
resolving vertices shared between adjacent cells; following common
practice~\cite{dyken2008high}, we therefore implement vertex generation and face
generation as two separate CUDA kernels in a multi-pass scheme.

\subsection{Fixed-Size-Array Mesh Topology}
\label{sec:topology}

The plain vertex/face list produced by MC carries no face adjacency: locating
the neighbors of a face requires an additional $O(|\mathcal{F}|)$ search, which
makes it a bottleneck for downstream mesh processing. We therefore introduce a fixed-size-array adjacency
data structure suited to fast construction and lookup on the GPU. For every edge
of a face $f$ we define an undirected edge key
$\mathrm{key}(v_i,v_j)=(\min(v_i,v_j),\max(v_i,v_j))$ and pack it into a 64-bit
integer. For all $3|\mathcal{F}|$ edges we form (key, face-ID) pairs and sort
them by key with a radix sort; faces that share an identical key are then
linked, which determines the neighbor set $\mathcal{N}(f)$ of each face.

Because a manifold mesh has at most three edge-adjacent neighbors per face, we
store $\mathcal{N}(f)$ in a fixed-size array of length three. Half-edge
structures~\cite{botsch2010polygon} also give constant-time access but store
several references per edge for general polygons; our layout keeps only the
three indices that triangle adjacency needs. Inside a CUDA kernel it yields
regular, coalesced memory access and suppresses warp divergence. Construction
consists of $O(|\mathcal{F}|)$ data-parallel steps (edge generation, a radix sort
over fixed-length keys, and neighbor registration) and completes on the GPU.
Boundary slots remain invalid. For non-manifold edge runs with more than two
incident faces, the sorted-edge pass registers one valid face pair and leaves the
remaining incidences disconnected for downstream propagation. As
the red stars in \cref{fig:pipeline} indicate, the subsequent stages (noise
removal, texture mapping, inpainting, and seam correction) all build on this
representation.

\subsection{Noise Removal}
\label{sec:nr}

Small isolated mesh fragments, caused by depth noise, occlude other surfaces
during texture mapping and degrade visual quality. To suppress them we run a
connected-component analysis on the GPU using a label-propagation
scheme~\cite{hawick2010parallel} built on the data structure of
\cref{sec:topology}. Each face is initialized with its own face index as a
label; we then propagate the minimum label over the adjacent faces (with
pointer jumping) until convergence or an iteration cap ($100$ in all
experiments, which bounds latency). Faces sharing a common label form a connected component; we count
the faces per component and remove components below a size threshold in a
compaction step. Casting noise removal as an independent per-face label update
lets it exploit the parallelism of the GPU.

\begin{figure*}[t]
\centering
\includegraphics[width=\textwidth]{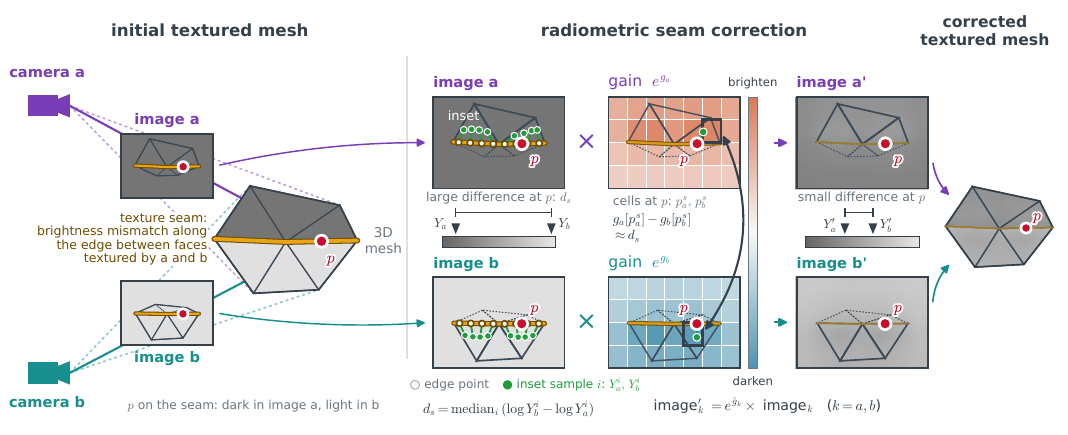}
\caption{Radiometric seam correction. Left: cameras $a$ and $b$ see the same surface at
different exposures (dark in image $a$, light in image $b$); faces above the
seam are textured from $a$ and faces below it from $b$, so brightness mismatches
along the \emph{texture seam}. Center: both images are sampled at eight
corresponding positions per seam edge (four drawn), each inset toward the
opposite vertex of that camera's own face ($\circ$ edge point, $\bullet$
sample); the median log-luminance difference $d_s$ constrains
the per-camera log-gain grids $g_k$, which live in each camera's image space.
The cells $p_a^s,p_b^s$ at the seam midpoint $p$ are coupled by
$g_a[p_a^s]-g_b[p_b^s]\approx d_s$, smoothness spreads the correction to
neighboring cells, and $e^{\hat g_k}$ multiplies the camera image (red
brightens, blue darkens, as in \cref{fig:colorcorr}). Right: the corrected
images and mesh; the correction is strongest at the seam and fades with
distance, and the mismatch at $p$ is reduced. Two cameras are shown; with $K$ cameras each
seam couples the grids of its two cameras in the same way, and all $K$ grids
are solved jointly.}
\label{fig:seam_algorithm}
\end{figure*}

\subsection{Texture Mapping}
\label{sec:texture}

Following prior work~\cite{Waechter2014Texturing,fu2018texture}, we adopt a
view-selection approach driven by the projected geometry of each face. We use a
fixed atlas layout that stacks the input RGB
images~\cite{lempitsky2007seamless,10494107}, giving each camera a known atlas
band. The texture mapping kernels compute UV coordinates and best-view
assignments directly against this layout on the GPU.
For each triangular face we test candidate cameras using its three vertices.
We reject a camera if a vertex has invalid observed depth or its camera-space
depth differs from the depth at its projection by more than a tolerance.
The default is \SI{10}{\centi\meter}, fixed across voxel sizes.
Among valid cameras we select the view that minimizes the angle between the
face normal and viewing direction, excluding angles above $80^{\circ}$ to
avoid projective distortion. Projecting the face into that camera's RGB image
yields local UV coordinates. Unassigned faces go to the inpainting stage below; the scaling
and offset of the atlas layout then give the final mesh UVs. We avoid MRF-style view-label
optimization because the face graph changes every frame, and a data-dependent
global solve would make latency hard to bound. Independent per-face assignment
maps directly onto the GPU; the seam correction stage below handles the
radiometric discontinuities it leaves.

\subsection{Inpainting}
\label{sec:inpainting}

The texture mapping stage of \cref{sec:texture} leaves some faces unassigned due
to occlusion. We fill them with a fast inpainting procedure that again uses the
data structure of \cref{sec:topology}. The procedure is iterative: whenever an
unassigned face has an already-assigned neighbor, we propagate the neighbor's UV
coordinates to it. To prevent artifacts, we copy the value of a single vertex of
the neighbor and fill the entire triangle with a uniform color. The procedure
terminates once no face changes state in an iteration or after a fixed cap of
five iterations, which bounds the latency. Each iteration is an independent
per-face GPU operation.

\subsection{Radiometric Seam Correction}
\label{sec:colorcorr}

View-selection texturing assigns one camera to each face. Adjacent faces that
use different cameras form a texture seam, where exposure differences can
produce a visible brightness step (\cref{fig:seam_algorithm}, left). Our
correction uses these seams as sparse measurements of radiometric mismatch and
spreads their evidence through a coarse gain field in each camera's image space
(\cref{fig:seam_algorithm}, center). The field varies spatially because a
single exposure offset per camera cannot represent mismatches that change
across the image, such as those left by lens vignetting, locally adaptive ISP
processing (e.g., local tone mapping), and view-dependent shading. Seams
constrain only differences between cameras: the field reduces these relative
mismatches, while shading common to all cameras is not observable.

\paragraph{Measure mismatch at seams.}
We identify seams by comparing camera IDs across the existing face adjacency.
At eight corresponding positions along each shared edge, we sample the two
source images after a small inward inset into their respective projected faces
(the edge points and inset samples in \cref{fig:seam_algorithm}).
Sampling a $1/8$-resolution linear-RGB proxy suppresses texture detail and
sensitivity to small misalignments. We exclude sharp folds and degenerate
projected faces, reject dark or invalid samples, and summarize each seam by
$d_s=\operatorname{median}_i(\log Y_b^i-\log Y_a^i)$, where $Y_a^i,Y_b^i$
are the sampled luminances. The weight $w_s$ is the fraction of valid samples.
Thus, each accepted seam supplies one robust brightness-difference constraint
between its two source cameras.

\paragraph{Estimate smooth per-camera fields.}
Each camera $k$ has a $24\times16$ log-gain grid $g_k$. We associate each seam
with one nearest cell in each grid, at the inset projection of its midpoint;
call these cells $p_a^s$ and $p_b^s$ (the outlined cells in
\cref{fig:seam_algorithm}). Since multiplying luminance by $e^g$ adds
$g$ in log space, matching the corrected samples requires
$g_{a(s)}[p_a^s]-g_{b(s)}[p_b^s]\approx d_s$. We minimize
\begin{equation}
\begin{aligned}
E(\mathbf g)={}&\sum_{s\in\mathcal S}w_s
 \bigl(g_{a(s)}[p_a^s]-g_{b(s)}[p_b^s]-d_s\bigr)^2\\
 &+\lambda_{\mathrm{sm}}\sum_k\sum_{(p,q)\in\mathcal E}
       \bigl(g_k[p]-g_k[q]\bigr)^2
  +\lambda_{\mathrm a}\sum_k\|g_k\|^2,
\end{aligned}
\label{eq:cc_energy}
\end{equation}
where $a(s)$ and $b(s)$ are the two source cameras of seam $s$, $g_k[p]$ is
the value of $g_k$ at cell $p$, $\mathcal S$ is the seam set, and $\mathcal E$
contains each undirected four-neighbor grid edge once. The smoothness term
($\lambda_{\mathrm{sm}}$) propagates seam evidence into unobserved cells; the
anchor term ($\lambda_{\mathrm a}$) favors unit gain ($g=0$). Together with the coarse
grid, this favors gradual radiometric changes rather than independent
per-face offsets. The fields are estimated jointly: seams couple cameras
and smoothness couples cells within each camera.

\paragraph{Solve and apply.}
We use twelve matrix-free weighted Jacobi iterations ($\omega=0.7$). Each
update combines neighboring grid values with the seam constraints, clamps the
updated log-gain to $\pm0.15$, and then removes the global mean across cameras
to control overall exposure drift. Nearest-cell assignment keeps the data term
local during the solve. At bake time, we instead interpolate the solved field
\emph{bilinearly} and clamp the interpolated log-gain to $\pm0.15$.
The bake is then a single multiplication: each full-resolution linear-RGB
atlas pixel is scaled by $\exp(\hat g_k)$ (\cref{fig:seam_algorithm}, right)
before conversion back to sRGB. A scalar gain scales all channels
equally. The vertically stacked camera images give a direct mapping from each
atlas pixel to its camera's grid, without changing mesh UVs.
Camera exposure changes little between frames, so an exponential moving
average of the grids ($\alpha=0.15$, in all experiments) suppresses
frame-to-frame flicker of the correction ($8\times$ smaller frame-to-frame
change of the grid).

All stages run on the GPU and reuse the face adjacency, camera assignments,
and projected coordinates already produced by texturing. Persistent buffers
and a fixed iteration count keep the correction stage bounded.

\section{Experiments}
\label{sec:experiments}

We evaluate the runtime and visual quality of our method on the Panoptic Studio
dataset~\cite{Joo_2015_ICCV,Joo_2017_TPAMI}, in which ten Kinect~v2
sensors~\cite{KinectV2} are arranged on a hemisphere and observe a volume of
roughly $6\,\mathrm{m}\times 6\,\mathrm{m}$ and about $2\,\mathrm{m}$ in height.
We use two sequences: \textit{office2}, in which a single performer interacts
with a variety of objects, and \textit{band2}, in which three performers play
musical instruments. 
Runtimes are measured on \textit{office2}, frames 400--2399;
held-out re-rendering and seam metrics use 100 frames.

\begin{figure}[t]
  \centering
  \includegraphics[width=.95\linewidth]{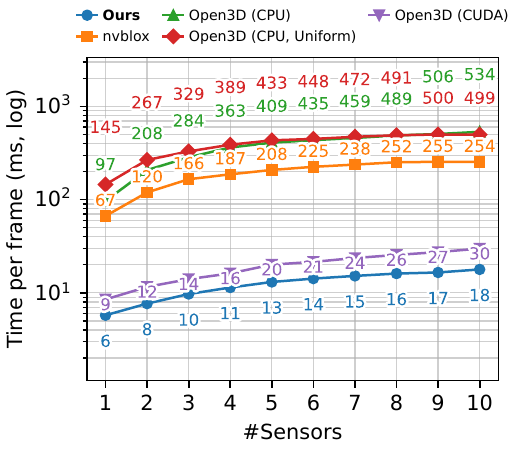}
  \caption{Per-frame processing time versus the number of sensors. Measured on
  the \textit{office2} sequence of CMU Panoptic at a voxel size of
  \SI{0.02}{\meter}. The vertical axis is the processing time per frame (ms, log
  scale). Our method's processing time increases more slowly with the number of
  sensors than the baselines. ``Ours'' is the full method,
  including the radiometric seam correction stage (\cref{sec:colorcorr}); it
  stays below all baselines at every sensor count.}
  \label{fig:bench_kinect_vs_time}
\end{figure}

\begin{table}[t]
\centering
\footnotesize
\setlength{\tabcolsep}{3.0pt}
\renewcommand{\arraystretch}{1.0}
\caption{Per-stage runtime breakdown (ms) on \textit{office2} (voxel
\SI{0.02}{\meter}, five Kinect inputs; about $0.59$M output faces per
frame). ``Ours w/o sc'' disables the seam correction, and ``Ours
(plain)'' drops all texture stages for comparison with the per-vertex-color
baselines nvblox, Open3D (CPU; Scalable variant with noise removal), and Open3D
(CUDA). Inpainting
includes the adjacency rebuild (1.3 of 1.7~ms). Columns are separate
runs; shared stages vary by up to ${\sim}0.3$~ms with GPU timing.}
\label{tab:runtime_breakdown}
\sisetup{
  table-number-alignment = center,
  detect-weight = true
}
\newcommand{\NA}{\multicolumn{1}{c}{N.A.}}
\begin{tabular}{
l
S[table-format=2.1]
S[table-format=2.1]
S[table-format=2.1]
S[table-format=3.1]
S[table-format=4.1]
S[table-format=2.1]
}
\toprule
Stage & {Ours} & {\shortstack{Ours\\w/o sc}} & {\shortstack{Ours\\(plain)}} & {nvblox} & {\shortstack{Open3D\\(CPU)}} & {\shortstack{Open3D\\(CUDA)}} \\
\midrule
CPU$\rightarrow$GPU  & \bfseries 2.0 & 1.9 & 1.9 & 3.3   & \NA   & 4.4 \\
Preprocessing$^{\dagger}$ & \bfseries 0.3 & 0.3 & 0.3 & \NA & \NA & \NA \\
TSDF Fusion          & \bfseries 0.4 & 0.4 & 0.4 & 36.3  & 95.7  & 8.2 \\
Marching Cubes       & \bfseries 0.5 & 0.5 & 0.6 & 28.1  & 299.6 & 5.7 \\
Noise Removal        & \bfseries 3.2 & 2.9 & 3.4 & \NA   & 957.0 & \NA \\
Texture Mapping      & \bfseries 0.2 & 0.1 & {--} & \NA  & \NA   & \NA \\
Inpainting           & \bfseries 1.7 & 1.7 & {--} & \NA  & \NA   & \NA \\
GPU$\rightarrow$CPU  & \bfseries 2.9 & 2.8 & 1.1 & 86.3  & \NA   & 1.8 \\
Seam Corr.\        & \bfseries 1.9 & {--} & {--} & \NA & \NA & \NA \\
Other                & \bfseries 0.2 & 0.2 & 0.1 & 54.2  & 7.2   & 0.0 \\
\midrule
Total                & {\bfseries \ccTotalMs} & 10.8 & 7.7 & 208.2 & 1359.4 & 20.1 \\
Speedup (full)$^{\ddagger}$ & & & & \multicolumn{1}{c}{15.9$\times$} & \multicolumn{1}{c}{103.8$\times$} & \multicolumn{1}{c}{1.5$\times$} \\
Speedup (plain)$^{\ddagger}$ & & & & \multicolumn{1}{c}{27.0$\times$} & \multicolumn{1}{c}{176.5$\times$} & \multicolumn{1}{c}{2.6$\times$} \\
\bottomrule
\end{tabular}
\\[1.5pt]
{\scriptsize $^{\dagger}$ Undistortion and depth-to-color registration, on
the GPU in ours; baselines run it on the CPU, untimed. $^{\ddagger}$ Baseline
time divided by Ours or Ours (plain), respectively. Rows are rounded
independently and need not sum to the totals.}
\end{table}

\begin{figure*}[t]
  \centering
  \setlength{\tabcolsep}{1.5pt}
  \renewcommand{\arraystretch}{0.9}
  \setlength{\fboxsep}{0pt}
  \begin{tabular}{c c c c c c c}
    & \textbf{GT}
    & \textbf{Open3D w/o nr}
    & \textbf{Open3D w/ nr}
    & \textbf{nvblox}
    & \textbf{Ours}
    & \textbf{Offline} \\
    \rotatebox{90}{\quad\quad\quad\quad office2} &
    \includegraphics[width=0.155\textwidth]{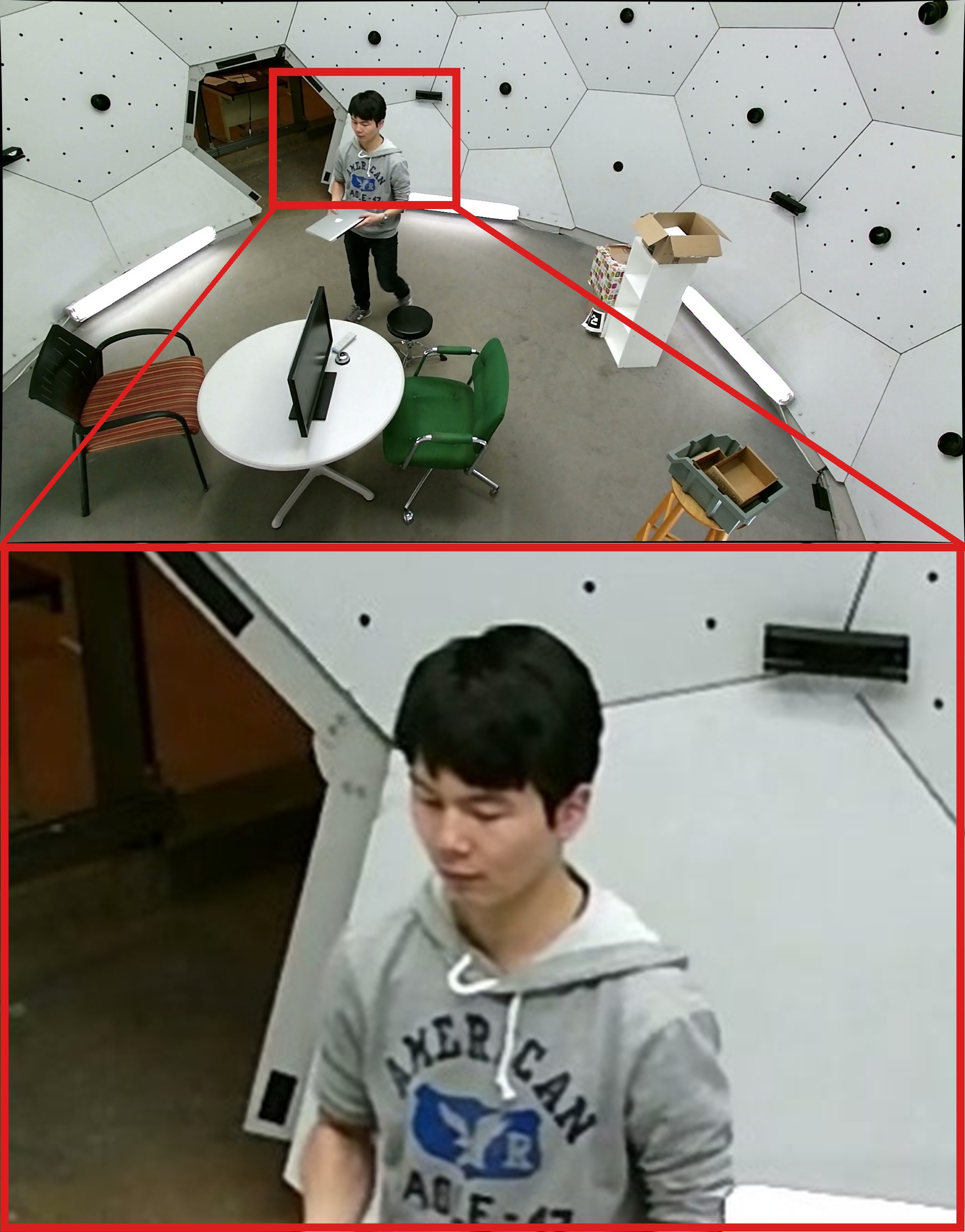} &
    \includegraphics[width=0.155\textwidth]{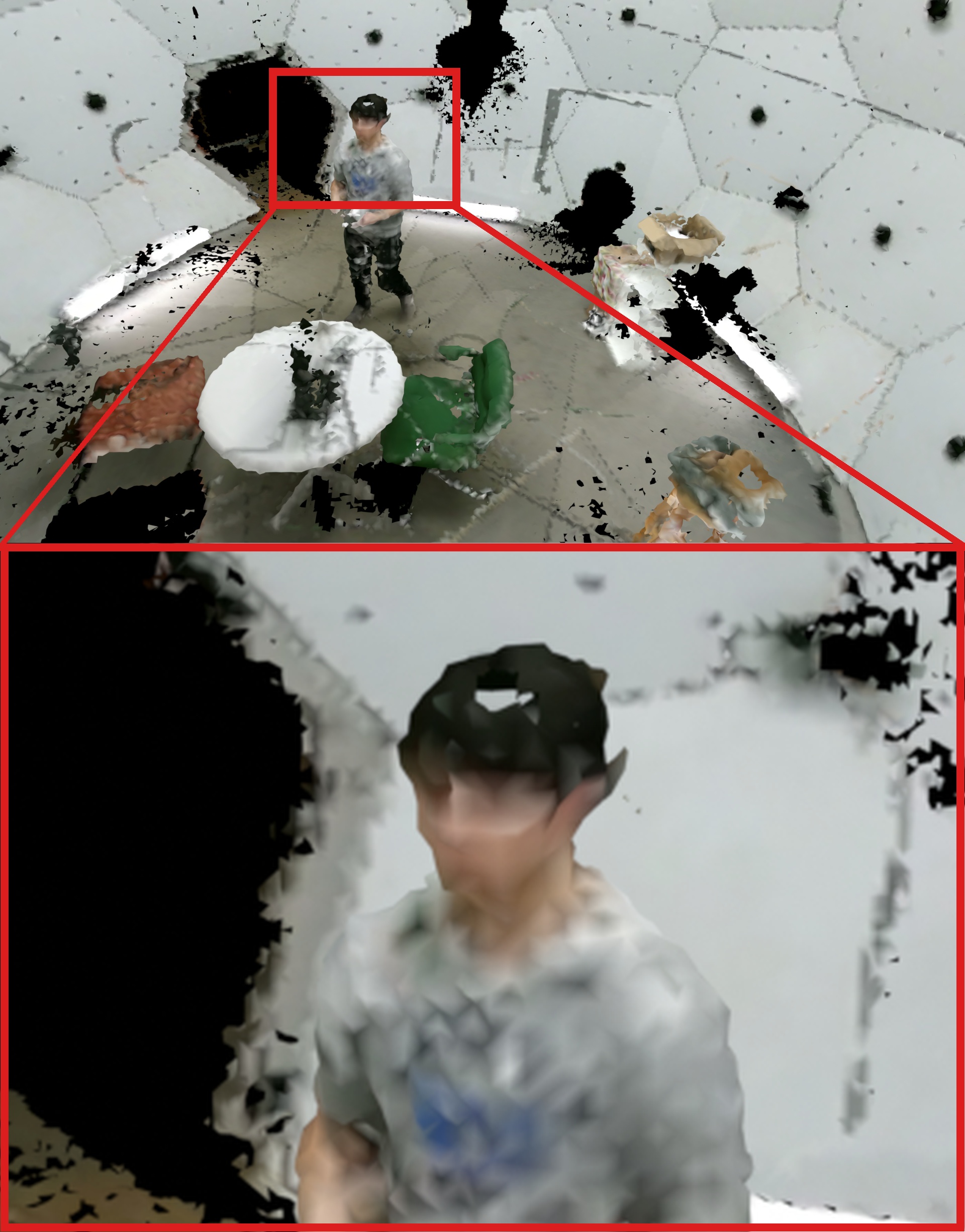} &
    \includegraphics[width=0.155\textwidth]{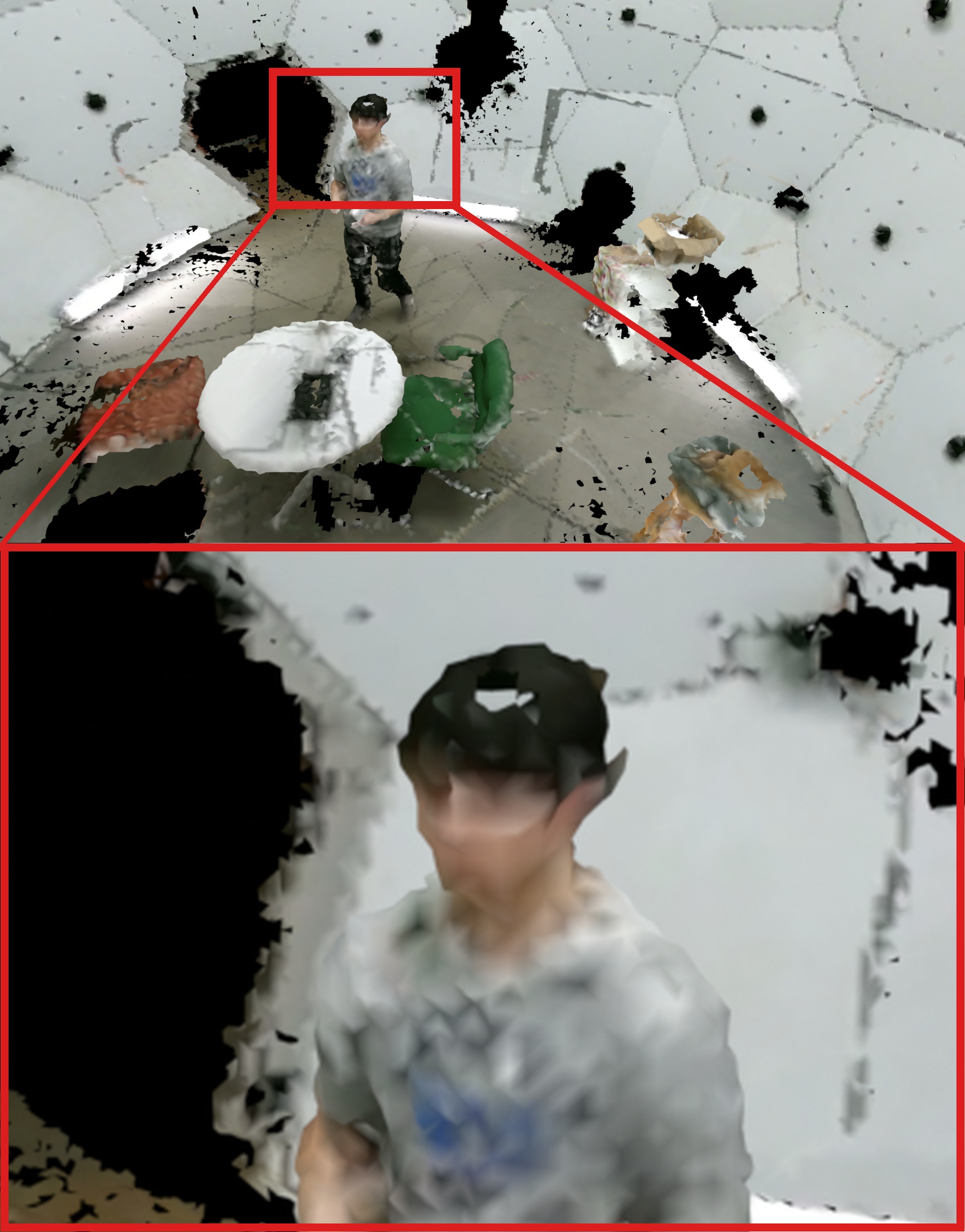} &
    \includegraphics[width=0.155\textwidth]{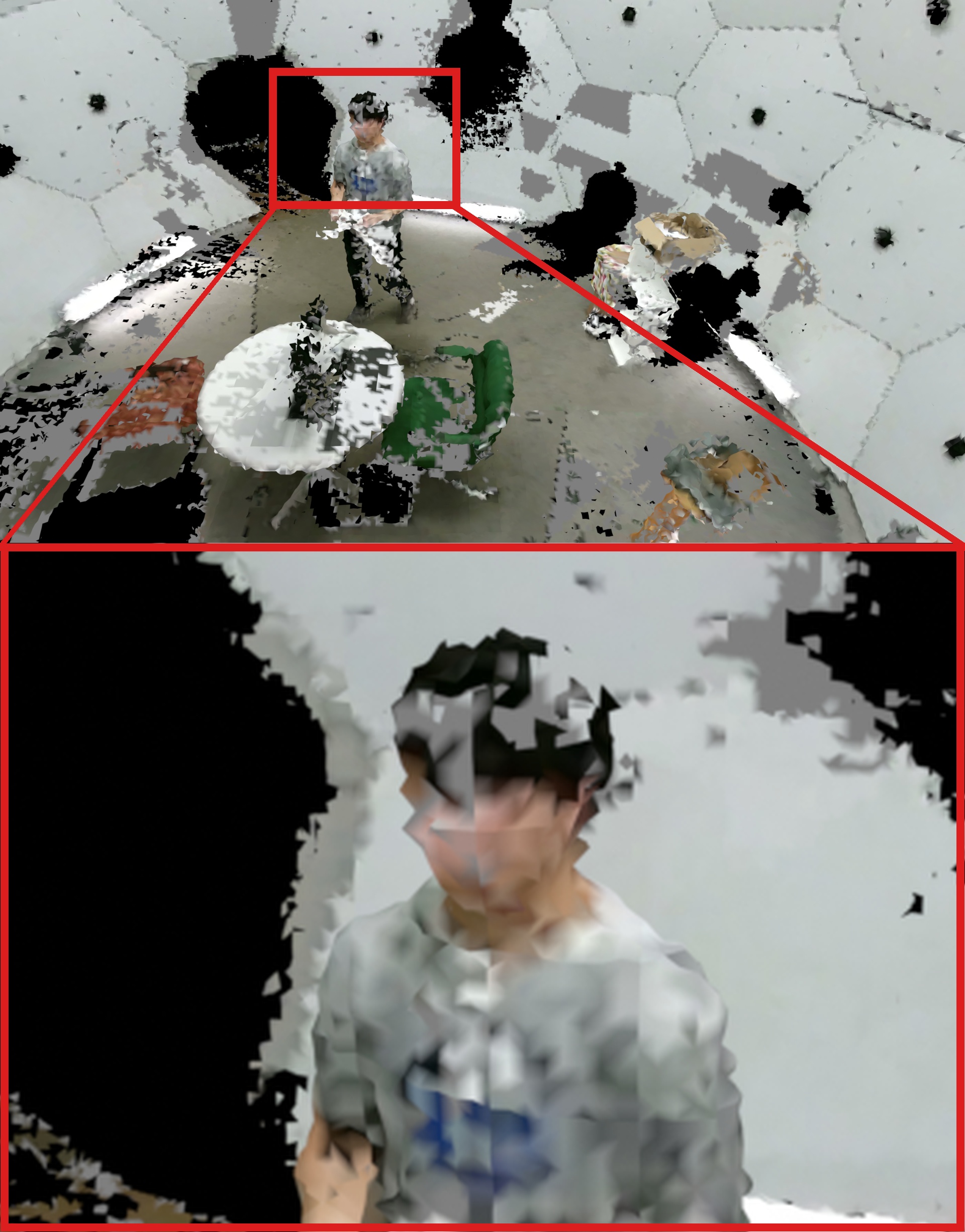} &
    \includegraphics[width=0.155\textwidth]{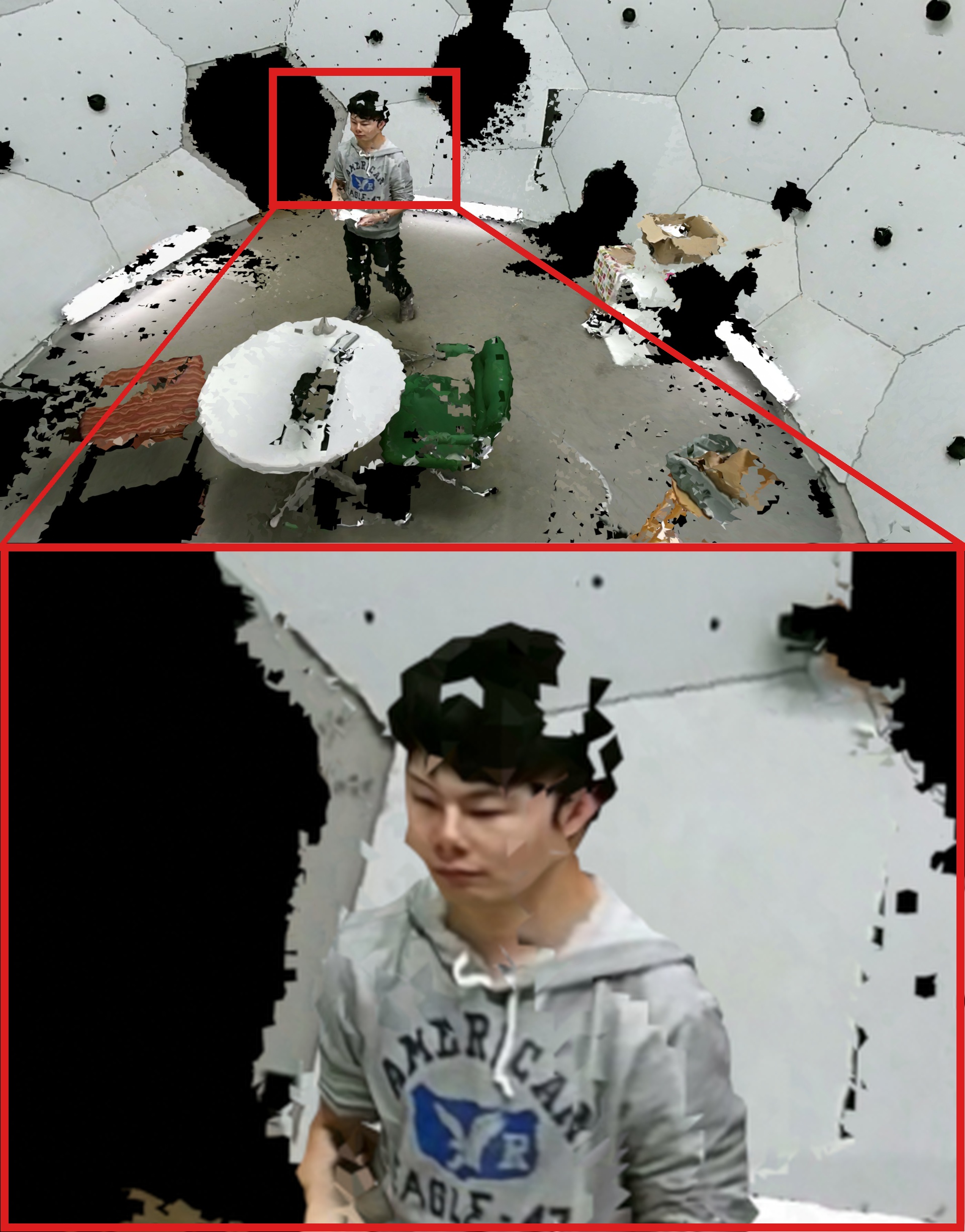} &
    \includegraphics[width=0.155\textwidth]{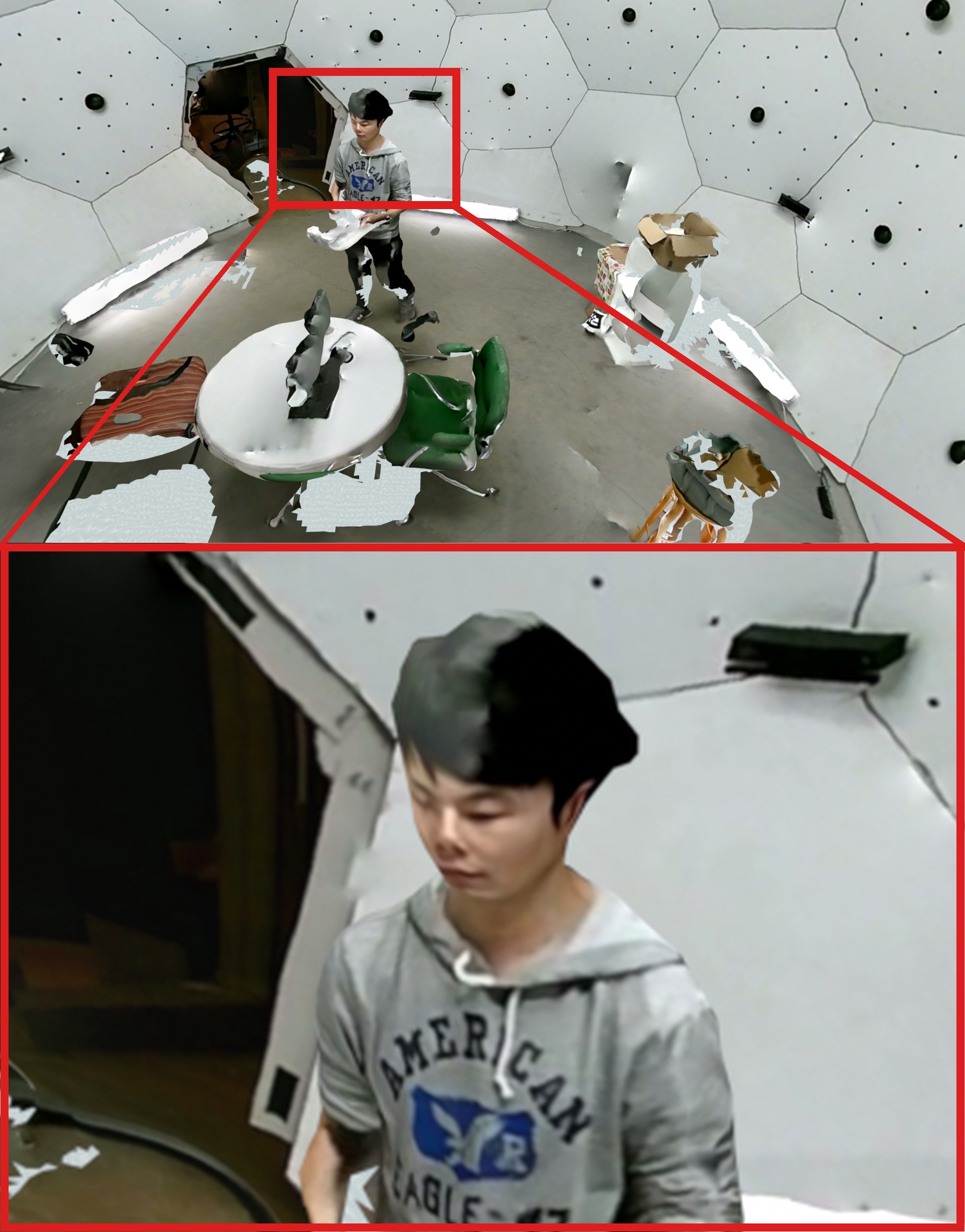} \\[-0.6mm]
    \rotatebox{90}{\quad\quad\quad\quad band2} &
    \includegraphics[width=0.155\textwidth]{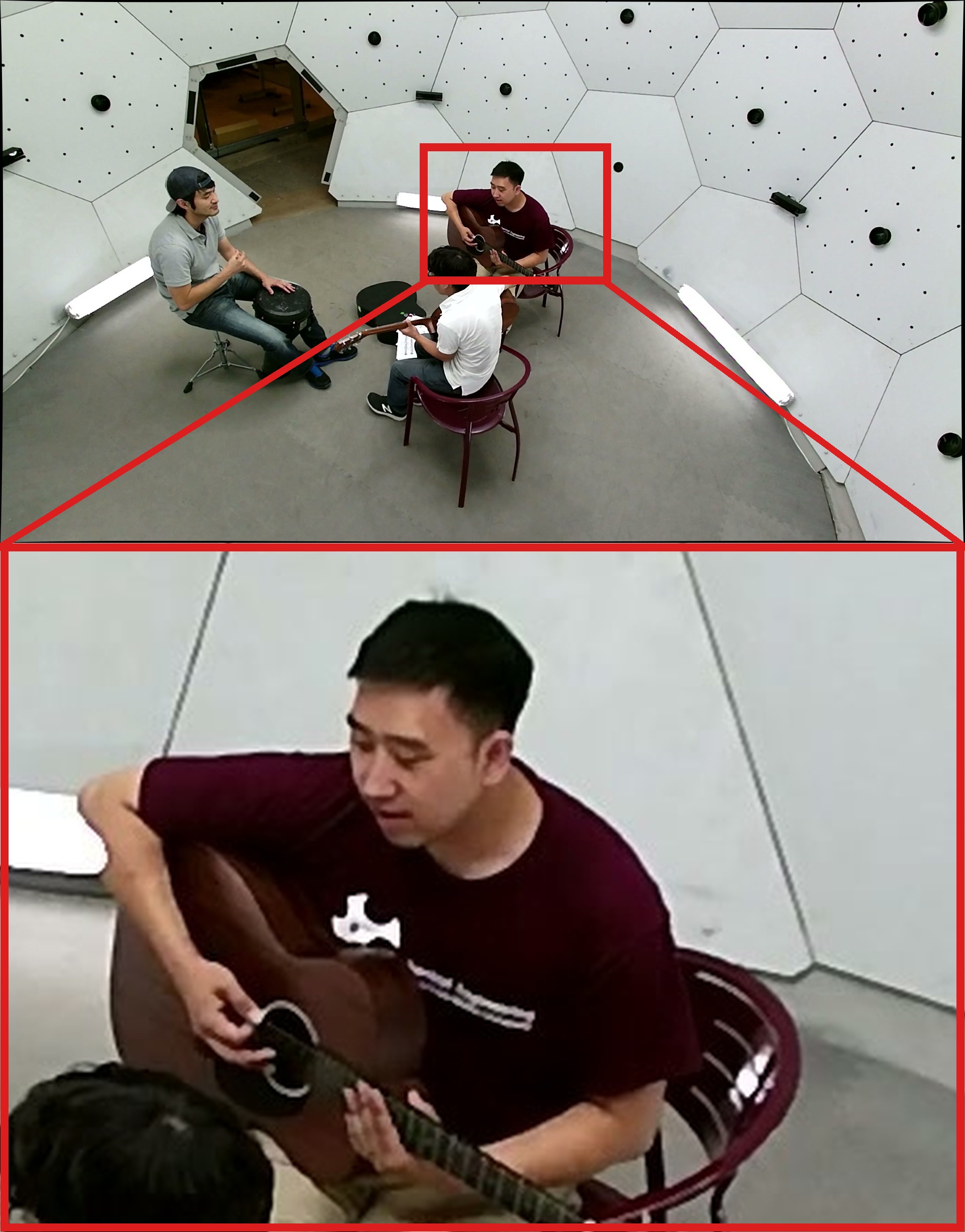} &
    \includegraphics[width=0.155\textwidth]{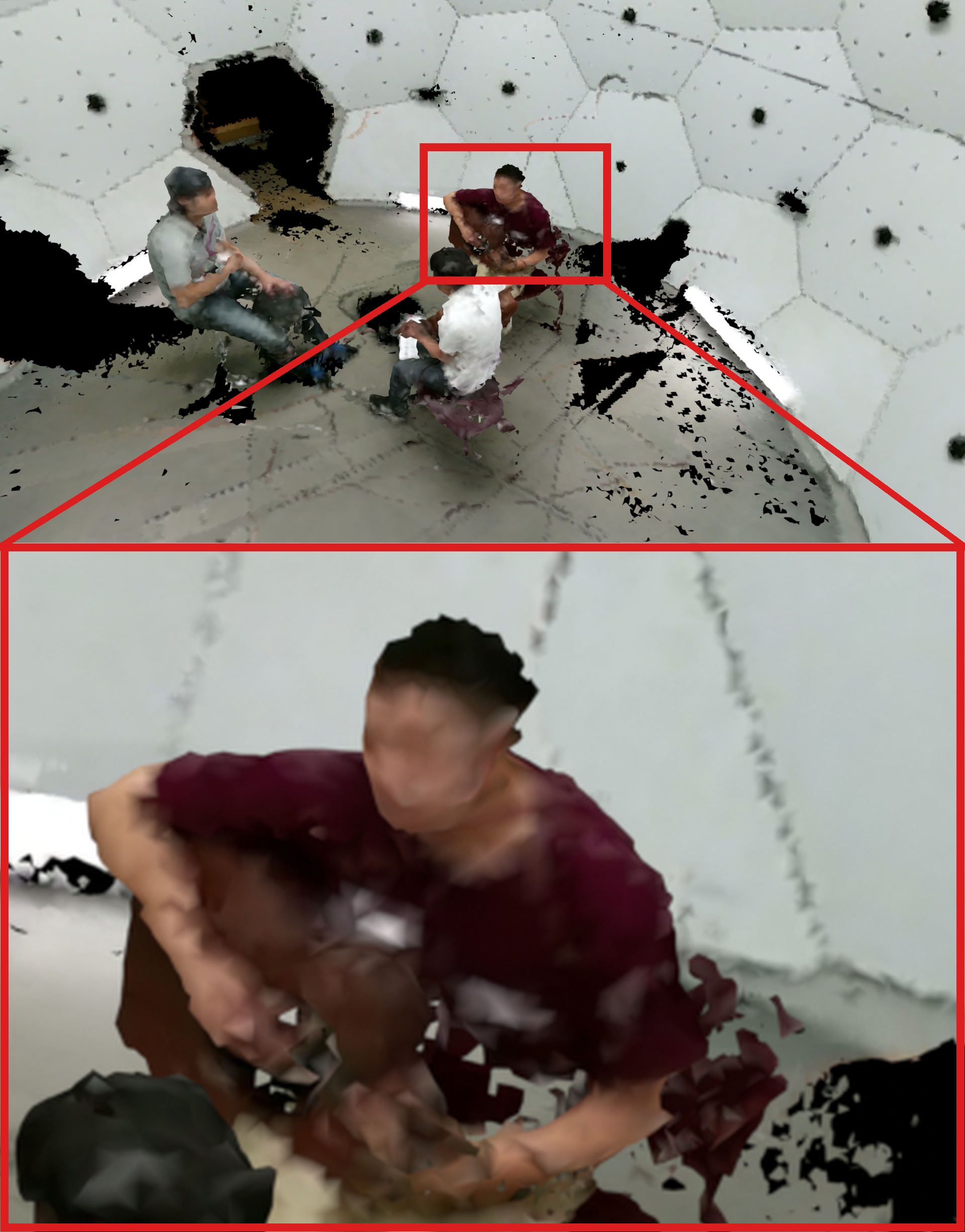} &
    \includegraphics[width=0.155\textwidth]{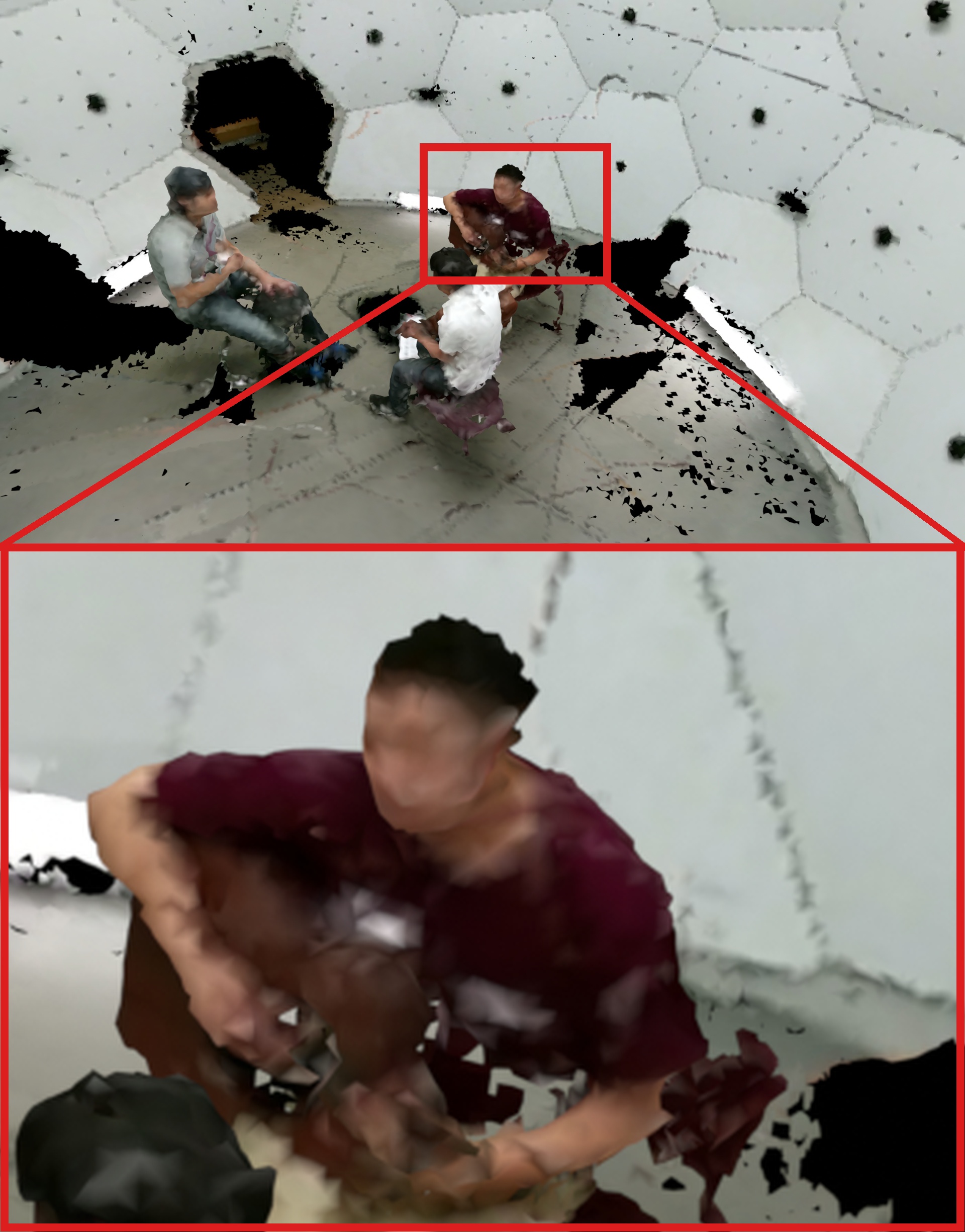} &
    \includegraphics[width=0.155\textwidth]{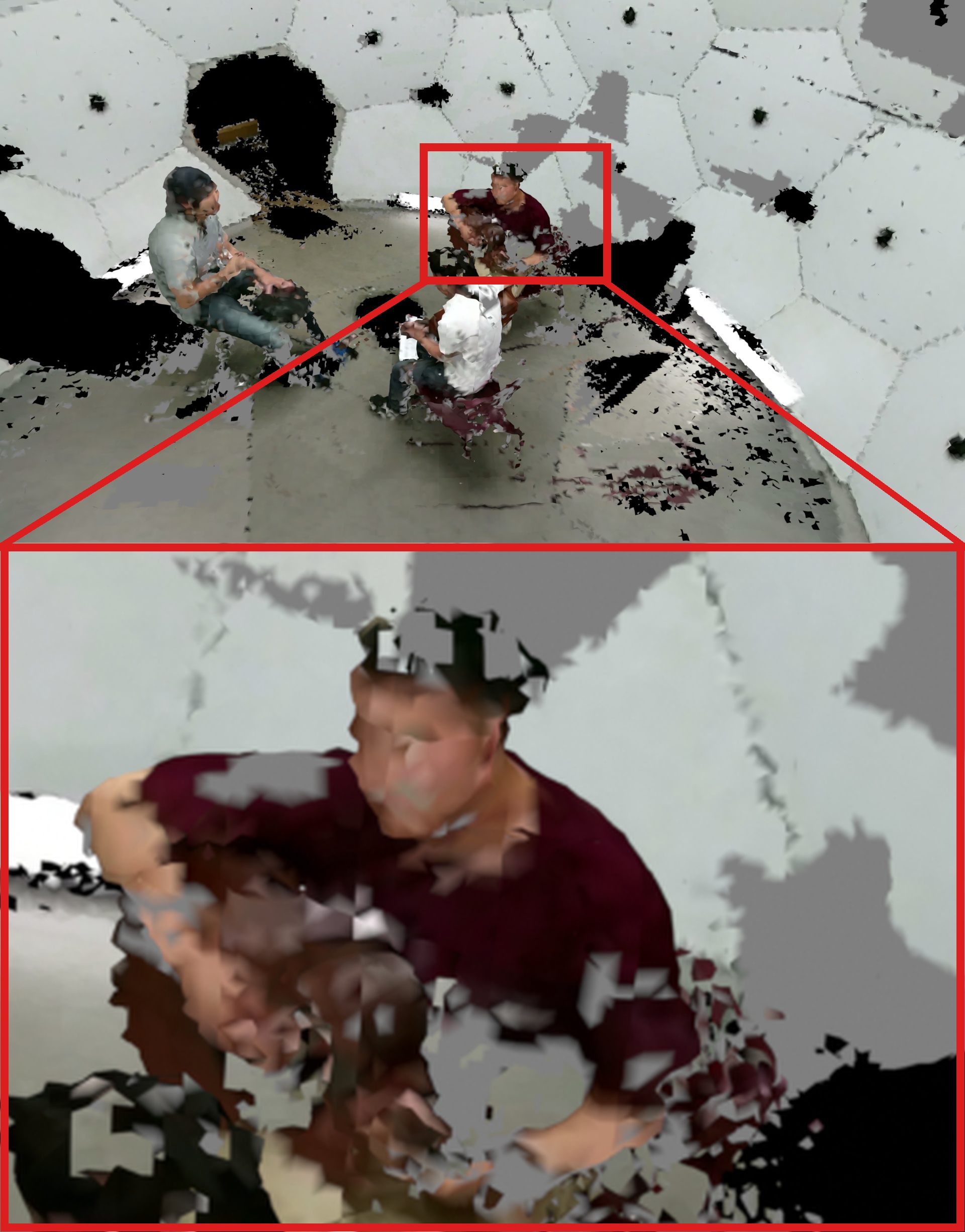} &
    \includegraphics[width=0.155\textwidth]{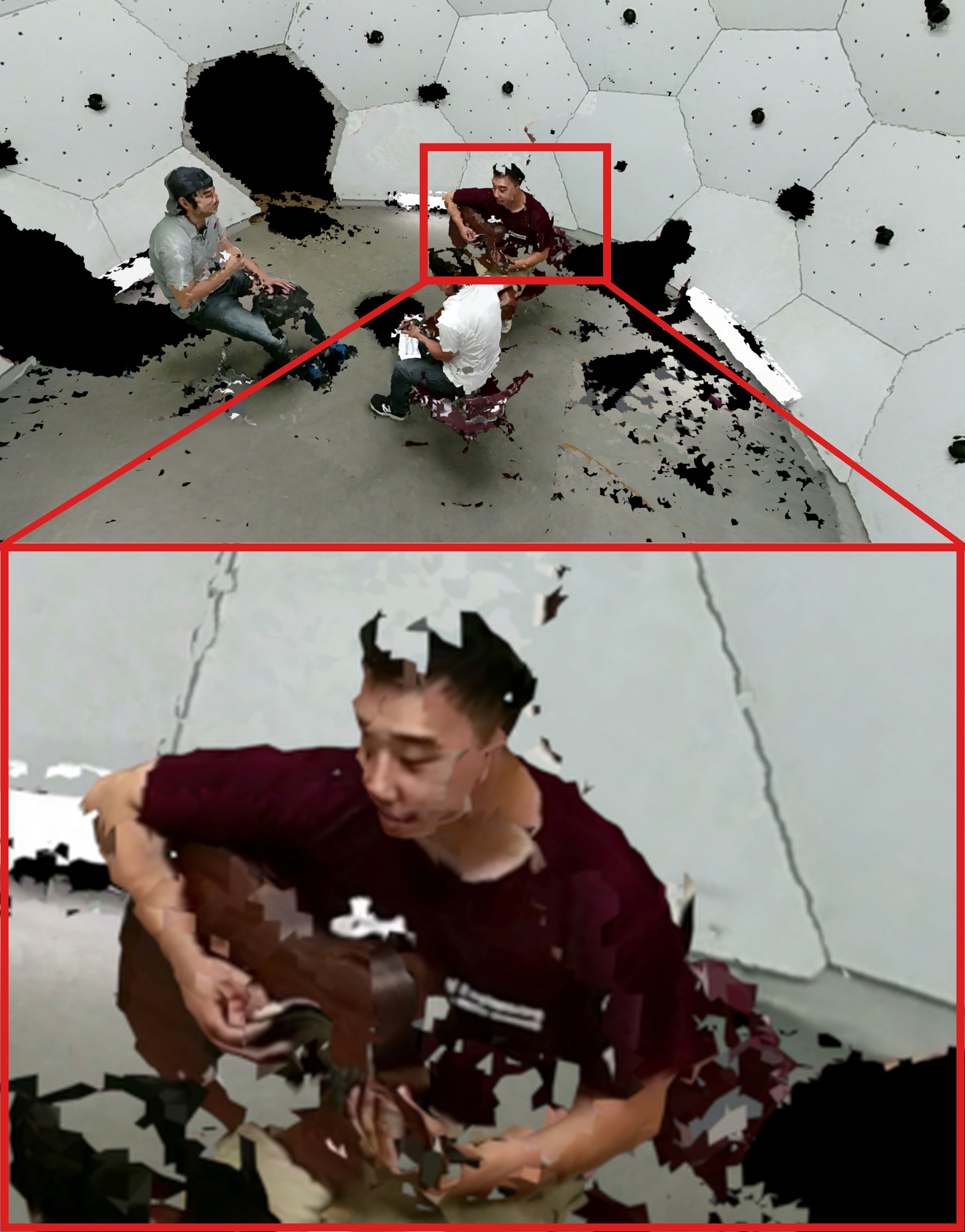} &
    \includegraphics[width=0.155\textwidth]{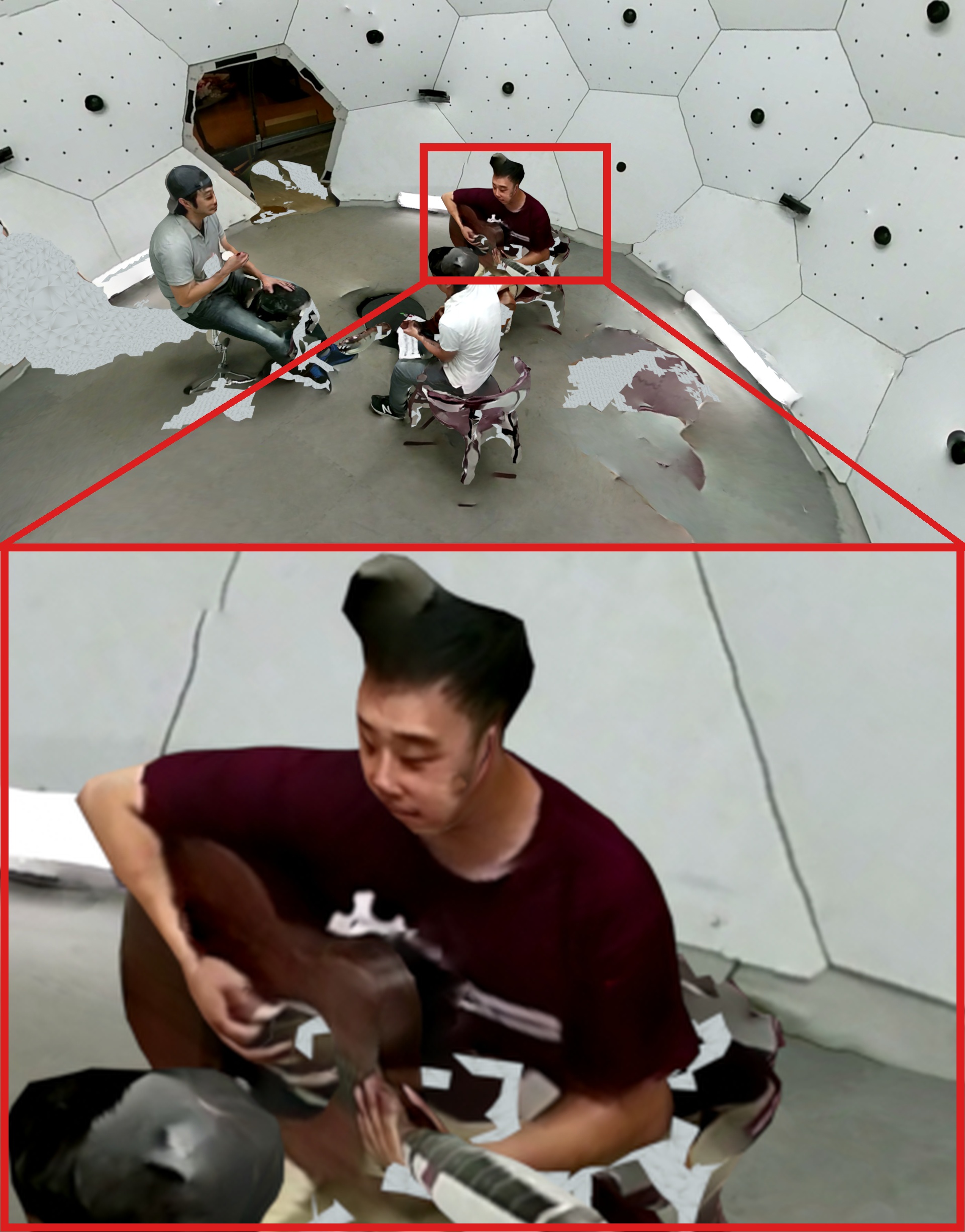} \\
  \end{tabular}
  \caption{Qualitative re-rendering comparison on the \textit{office2} and
  \textit{band2} sequences of CMU Panoptic (camera 9, frames 933 and 403,
  respectively). Meshes reconstructed from five Kinect
  inputs at a voxel size of \SI{0.02}{\meter} are re-rendered from the RGB
  viewpoints of a different set of five Kinects that were \emph{not} used as
  input. ``Ours'' includes the radiometric seam correction
  (\cref{sec:colorcorr}). Compared with the per-vertex-color outputs of Open3D
  and nvblox, our method reproduces facial and clothing textures more clearly,
  approaching the offline result at real-time speed. Quantitative scores are in
  \cref{tab:panoptic_quant}.}
  \label{fig:panoptic_qual}
\end{figure*}

\begin{table}[t]
\centering
\scriptsize
\setlength{\tabcolsep}{3pt}
\renewcommand{\arraystretch}{1.0}
\caption{Quantitative re-rendering quality on \textit{office2} and \textit{band2}
(held-out five-Kinect evaluation, voxel \SI{0.02}{\meter}), measured against the
GT images with 5\% trimmed mean PSNR, SSIM~\cite{wang2004image}, and
LPIPS~\cite{zhang2018unreasonable}. ``Ours'' is the full method including
radiometric seam correction and ``Ours w/o sc'' the ablation; ``Ours (vtx
color)'' replaces the texture atlas with per-vertex color on the same geometry,
isolating the atlas contribution. Metrics are computed on the common valid-pixel
mask, the intersection of covered pixels across all eight evaluated variants. The qualitative
comparison is in \cref{fig:panoptic_qual}.}
\label{tab:panoptic_quant}
\begin{tabular}{@{}l ccc ccc@{}}
  \toprule
  & \multicolumn{3}{c}{\textbf{office2}} & \multicolumn{3}{c}{\textbf{band2}} \\
  \cmidrule(lr){2-4} \cmidrule(lr){5-7}
  Method & PSNR$\uparrow$ & SSIM$\uparrow$ & LPIPS$\downarrow$
         & PSNR$\uparrow$ & SSIM$\uparrow$ & LPIPS$\downarrow$ \\
  \midrule
  Open3D w/o nr & \textbf{16.483} & 0.783 & 0.353 & \textbf{17.635} & 0.807 & 0.329 \\
  Open3D w/ nr & 16.408 & 0.784 & 0.353 & 17.590 & 0.808 & 0.330 \\
  nvblox & 15.604 & 0.759 & 0.378 & 16.549 & 0.785 & 0.355 \\
  Ours (vtx color) & 16.160 & 0.787 & 0.337 & 17.320 & 0.808 & 0.319 \\
  Ours w/o sc & \ccOffPsnrA & \ccOffSsimA & \ccOffLpipsA & \ccOffPsnrB & \ccOffSsimB & \ccOffLpipsB \\
  Ours & \ccOnPsnrA & \ccOnSsimA & \ccOnLpipsA & \ccOnPsnrB & \ccOnSsimB & \ccOnLpipsB \\
  Offline & 15.191 & \textbf{0.798} & \textbf{0.239} & 16.206 & \textbf{0.829} & \textbf{0.196} \\
  \bottomrule
\end{tabular}
\end{table}

\begin{figure}[t]
\centering
\begin{subfigure}[t]{0.32\linewidth}
\centering
\includegraphics[width=\linewidth]{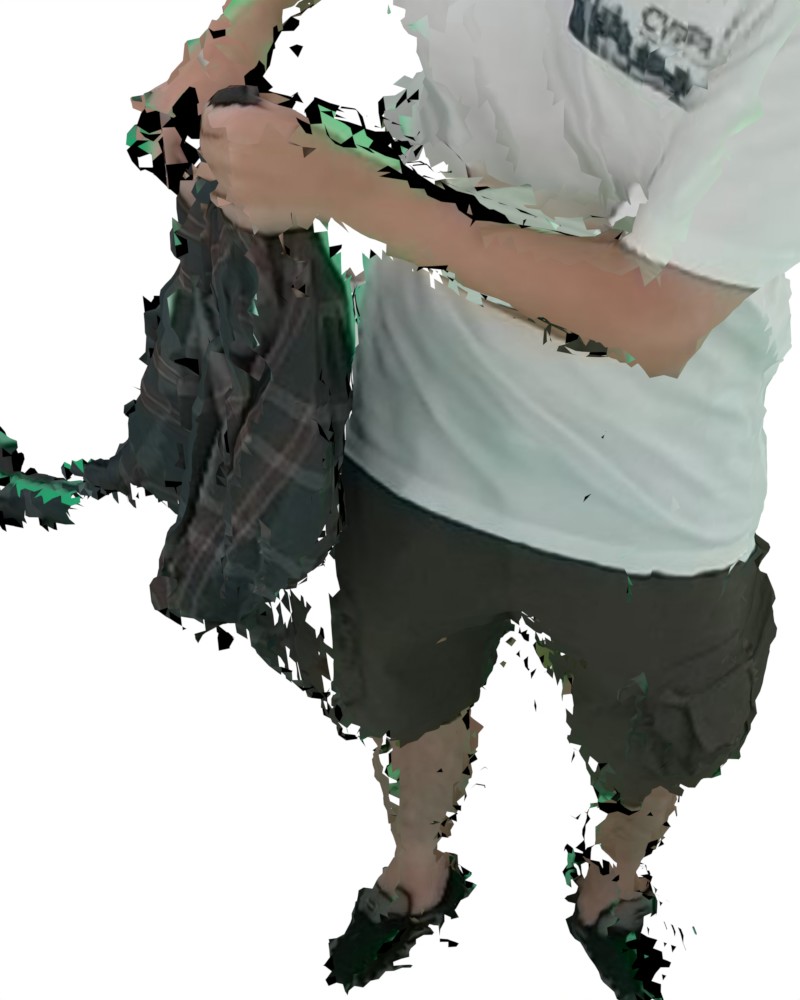}
\caption{\footnotesize Texturing}
\end{subfigure}
\hfill
\begin{subfigure}[t]{0.32\linewidth}
\centering
\includegraphics[width=\linewidth]{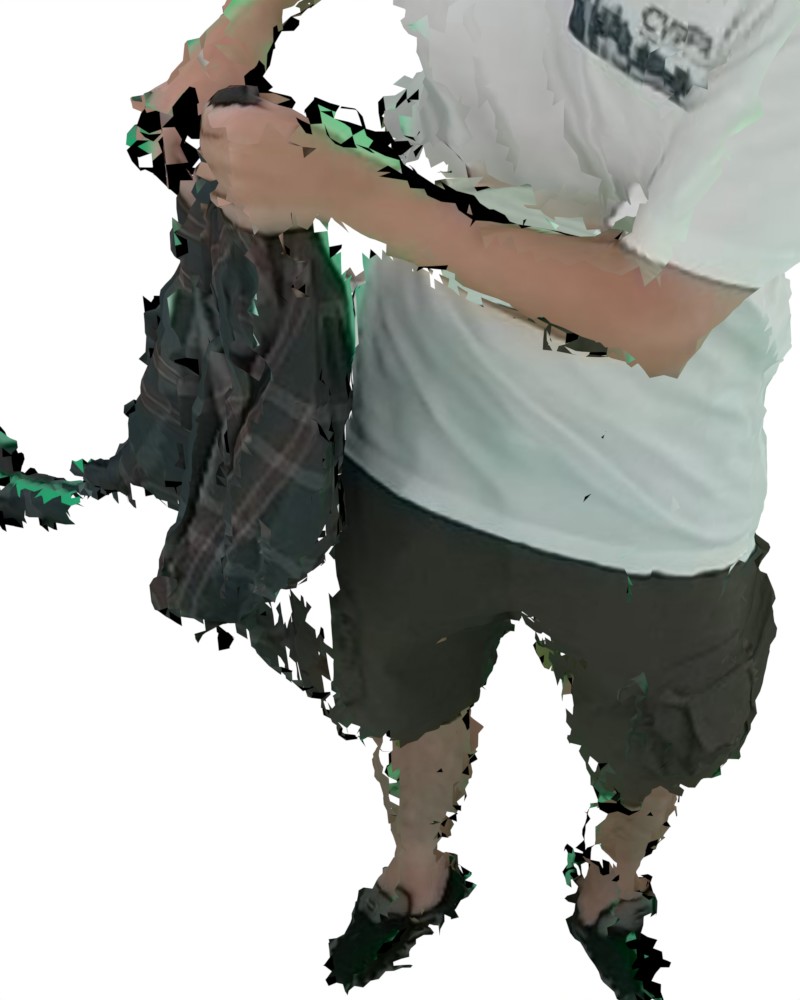}
\caption{\footnotesize + noise removal}
\end{subfigure}
\hfill
\begin{subfigure}[t]{0.32\linewidth}
\centering
\includegraphics[width=\linewidth]{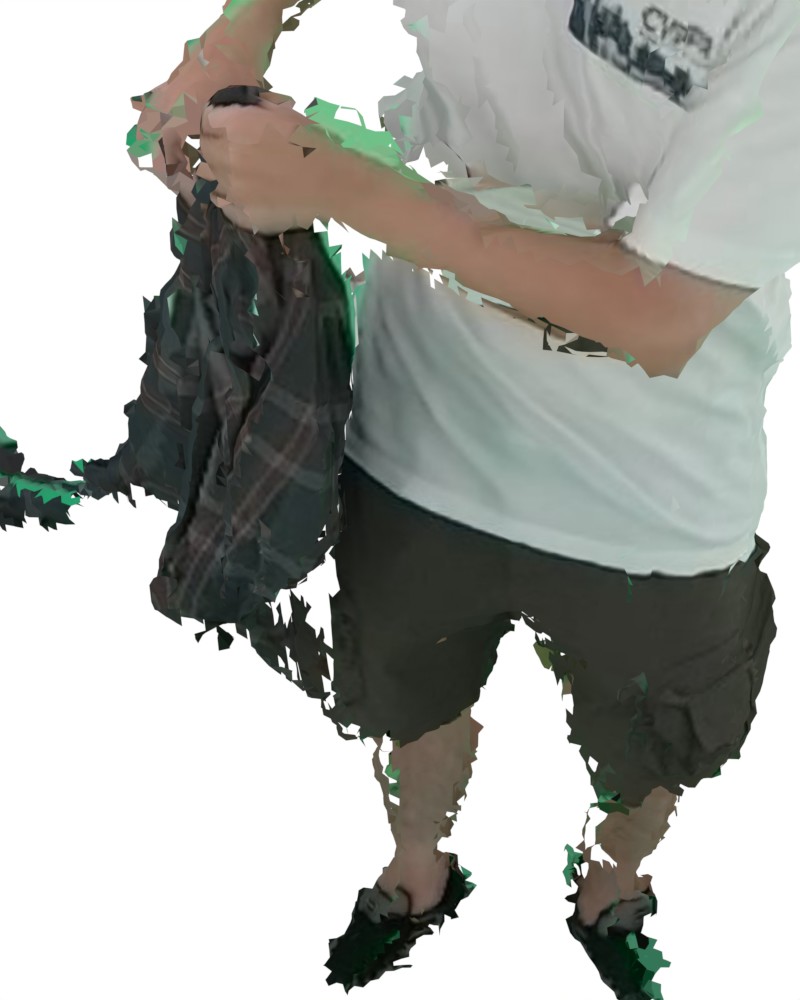}
\caption{\footnotesize + noise removal + inpainting}
\end{subfigure}
\caption{Effect of noise removal and inpainting on a foreground-only in-house
capture. The subject swings a shirt, producing noisy depth near the cloth and
hands. (a) uses texture mapping without noise removal
or inpainting; it leaves isolated fragments and untextured faces. (b) Noise removal
suppresses the fragments, and (c) inpainting fills many remaining untextured faces.
See the supplementary video for temporal behavior.}
\label{fig:own}
\end{figure}

\subsection{Runtime Evaluation}
\label{sec:exp_runtime}

\cref{fig:bench_voxel_vs_time} reports processing time as a function of voxel
size and \cref{fig:bench_kinect_vs_time} as a function of the number of
sensors.\footnote{All runtime measurements were performed on a system with an
AMD Ryzen Threadripper 9970X, \SI{256}{GB} of RAM, and an NVIDIA RTX PRO 6000
Blackwell Max-Q.} We compare against the GPU implementation nvblox and
Open3D, both of which represent appearance with per-vertex color.
nvblox targets incremental mapping rather than per-frame full textured-mesh
output, so this comparison measures the cost of adapting such systems to our
volumetric-video protocol.
On the CPU we run Open3D's Scalable~\cite{10.1145/2461912.2461919} (plain
``Open3D'' below) and Uniform variants, and for the former we also report the
effect of connected-component noise removal (w/ nr, w/o nr); Open3D (CUDA), its
GPU path, performs fusion and MC only. Anticipating
volumetric-video applications that capture an entire space, we reconstruct
everything visible without distinguishing foreground from background. Sensor
input arrives on the CPU and the mesh returns to it for transmission, so the
reported runtime includes the CPU--GPU transfers.

Our method is faster than the baselines across the tested settings. In particular, at a
voxel size of \SI{0.02}{\meter} the full method runs at \SI{13.1}{\milli\second}
per frame (\SI{10.9}{\milli\second} excluding input upload and preprocessing),
within the \SI{33}{\milli\second} (\SI{30}{fps}) budget of typical
RGB-D sensors and thus real-time (the texture-free ``Ours (plain)'' pipeline of
\cref{tab:runtime_breakdown} is faster still, at \SI{7.7}{\milli\second}). At a
voxel size of \SI{0.005}{\meter} it attains a $29.4\times$ speedup over nvblox and
an $89.7\times$ speedup over Open3D on the CPU ($179.9\times$ over its Uniform variant and
$1.6\times$ over Open3D (CUDA), the closest baseline); peak
GPU memory at this setting is about \SI{24.1}{GB}, of which \SI{16.6}{GB} are
static grid and mesh buffers.
\cref{tab:runtime_breakdown} gives the per-stage breakdown; it attributes much
of nvblox's runtime to CPU--GPU transfer and GPU-side container overhead, which
our dense GPU-resident design avoids.

\subsection{Visual Quality Evaluation}
\label{sec:exp_quality}

\cref{fig:panoptic_qual} and \cref{tab:panoptic_quant} show the qualitative and
quantitative evaluation. We reconstruct from five Kinects and use the remaining
five as evaluation viewpoints. We set the voxel size to \SI{0.02}{\meter}, at which our
method runs in real time. Prior online texturing systems are not direct
baselines here: they assume persistent single-sensor scans or CPU-side global
patch optimization, whereas our input is a fresh multi-view mesh each frame. We
therefore bracket the texture stage
with two references on the same protocol: a real-time per-vertex-color variant
on identical geometry and an offline texture-optimization reference. For the
offline reference we use
SPSR~\cite{kazhdan2013screened} followed by
mvs-texturing~\cite{Waechter2014Texturing}. Our method reproduces sharp facial
and clothing textures and ranks second to Offline in LPIPS. PSNR follows a
different order: Offline scores lowest and Open3D highest. This pattern is
consistent with the perception--distortion tradeoff~\cite{blau2018perception}.
On identical geometry, replacing our atlas with per-vertex color
(``Ours (vtx color)'') raises PSNR/SSIM but worsens LPIPS, isolating the
atlas's benefit to perceptual similarity. These scores compare appearance
on shared covered pixels; they do not measure missing surface coverage. The
offline method takes about \SI{13}{\second}
per frame (about \SI{3}{\second} for SPSR and \SI{10}{\second} for mvs-texturing),
about $993\times$ slower than ours. Among the compared methods, only ours
has an explicit temporal component: the others texture each frame
independently, whereas our gain field is smoothed over time
(\cref{sec:colorcorr}), which suppresses frame-to-frame changes of the
correction. The supplementary video shows the temporal behavior and
further qualitative comparisons.

\begin{figure*}[t]
\centering
\footnotesize
\setlength{\tabcolsep}{1pt}
\renewcommand{\arraystretch}{0.9}
\setlength{\fboxsep}{0pt}
\begin{tabular}{@{}ccc@{}}
  \includegraphics[width=0.26\linewidth]{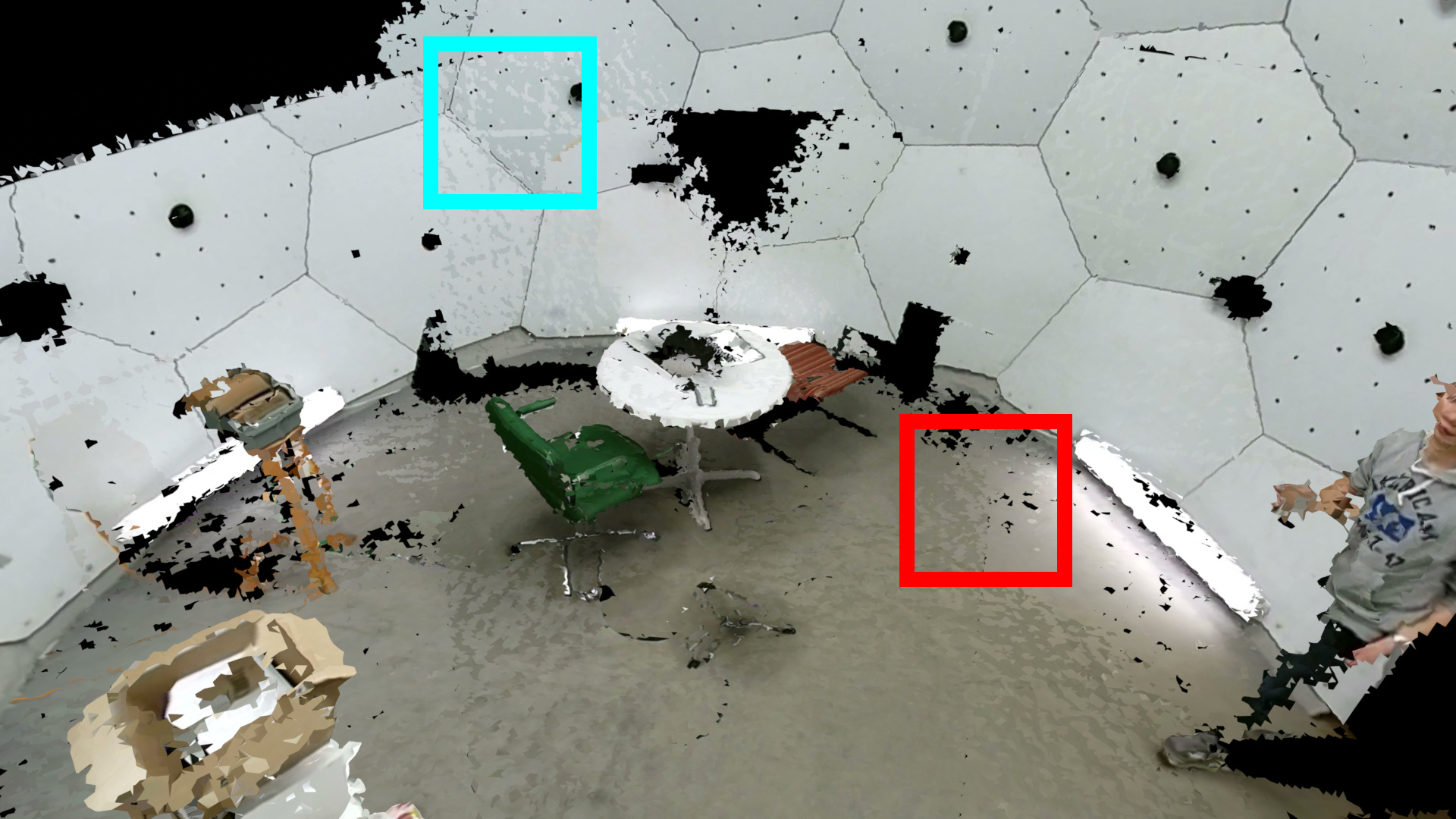} &
  \includegraphics[width=0.26\linewidth]{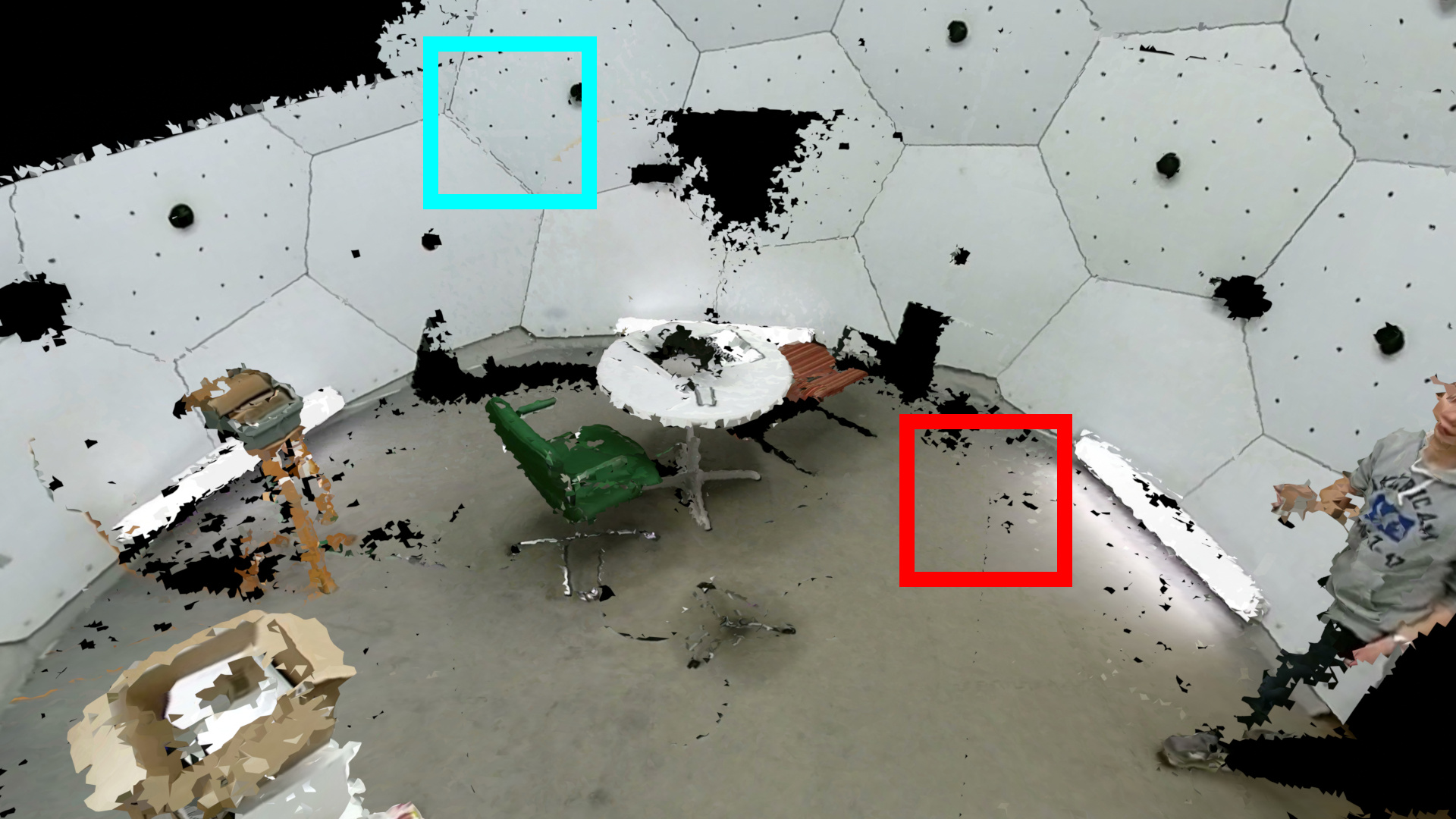} &
  \includegraphics[width=0.26\linewidth]{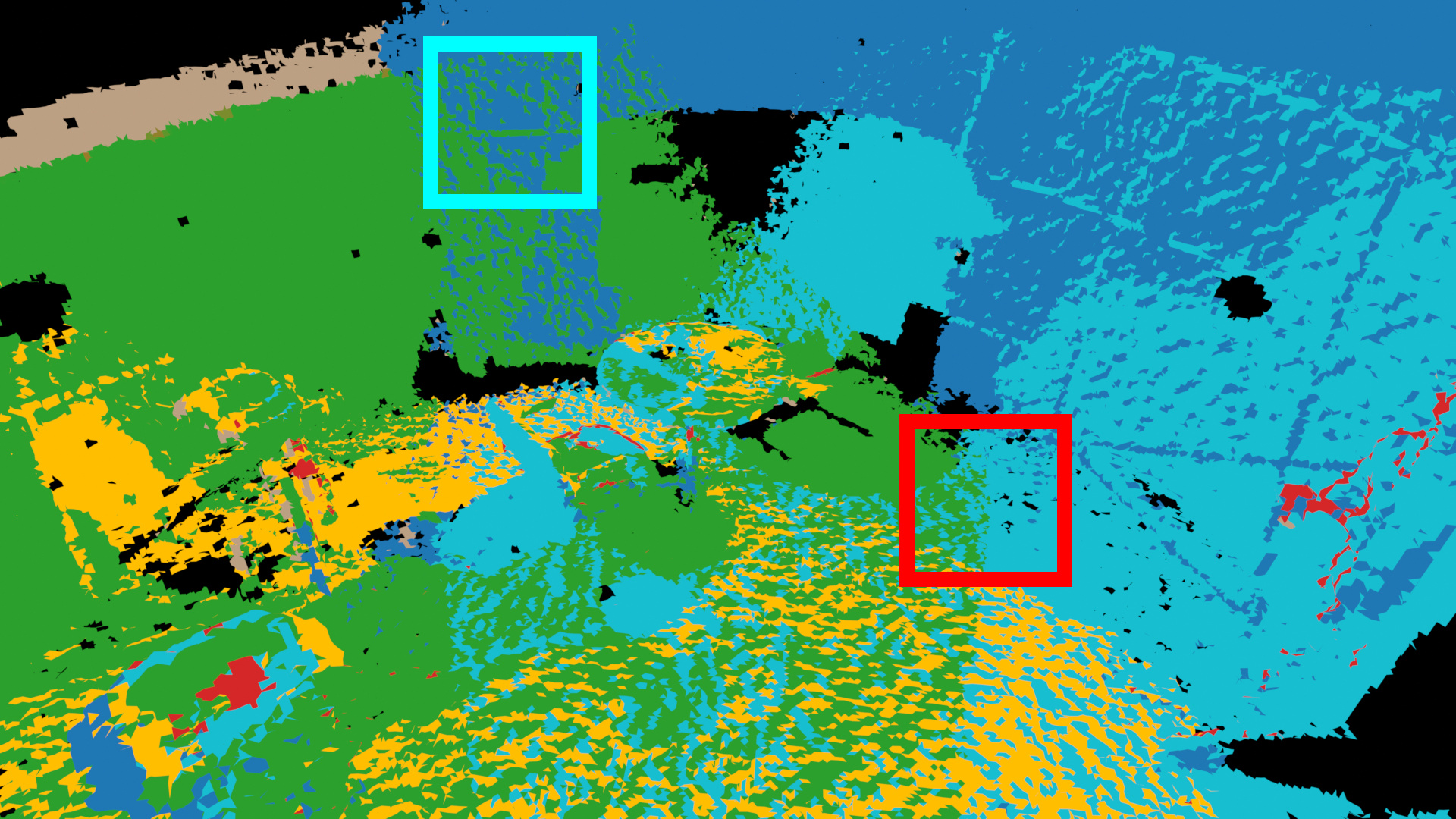} \\
  w/o seam corr. & w/ seam corr. & source camera (1 color $=$ 1 input view) \\
\end{tabular}
\\[1pt]
(a) Held-out (novel) view: full renders and the source-camera mosaic
(cyan box: wall-panel junction; red box: floor)
\\[3pt]
\begingroup
\setlength{\fboxrule}{1pt}
\definecolor{regioncyan}{RGB}{0,255,255}
\begin{tabular}{@{}ccc@{\hskip 12pt}ccc@{}}
  \multicolumn{3}{c}{wall (cyan)} & \multicolumn{3}{c}{floor (red)} \\
  \fcolorbox{regioncyan}{white}{\includegraphics[width=\dimexpr0.13\linewidth-2\fboxrule\relax]{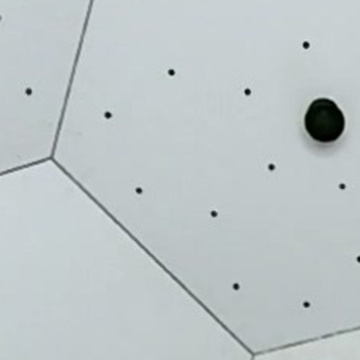}} &
  \fcolorbox{regioncyan}{white}{\includegraphics[width=\dimexpr0.13\linewidth-2\fboxrule\relax]{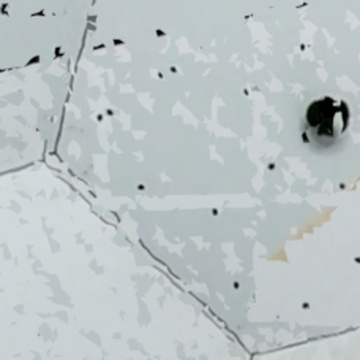}} &
  \fcolorbox{regioncyan}{white}{\includegraphics[width=\dimexpr0.13\linewidth-2\fboxrule\relax]{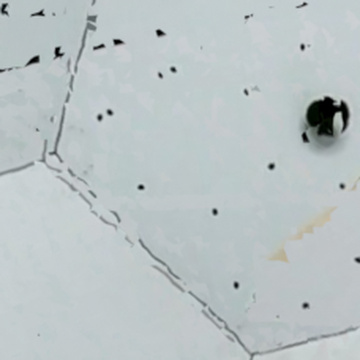}} &
  \fcolorbox{red}{white}{\includegraphics[width=\dimexpr0.13\linewidth-2\fboxrule\relax]{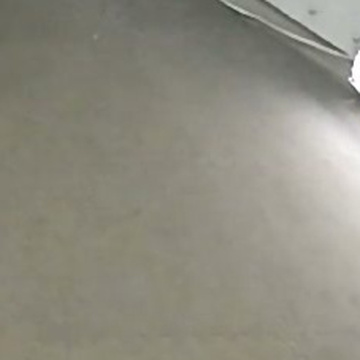}} &
  \fcolorbox{red}{white}{\includegraphics[width=\dimexpr0.13\linewidth-2\fboxrule\relax]{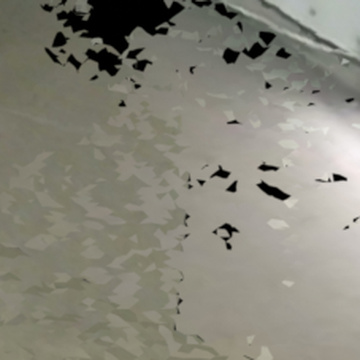}} &
  \fcolorbox{red}{white}{\includegraphics[width=\dimexpr0.13\linewidth-2\fboxrule\relax]{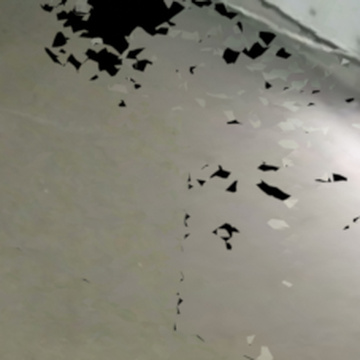}} \\
  GT & w/o & w/ & GT & w/o & w/ \\
\end{tabular}
\endgroup
\\[1pt]
(b) Zoom-ins (GT / w/o / w/ seam corr.) of the two boxed regions
\\[3pt]
\includegraphics[width=0.85\linewidth]{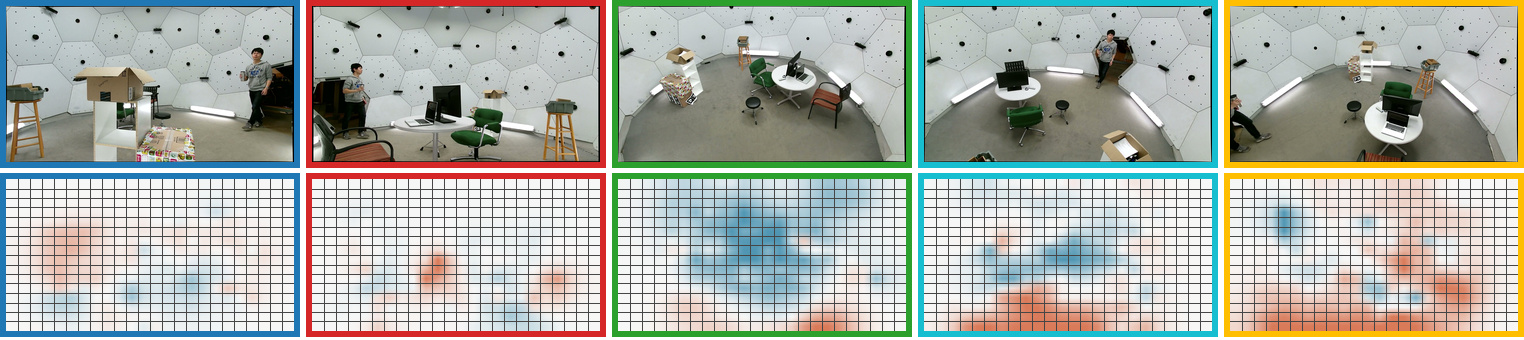}
\\[1pt]
(c) Each input view: its image (top) and low-frequency gain $\exp(g_k)$ on a
$24\times16$ grid (bottom)
\caption{Radiometric seam correction on \textit{office2}, rendered from a
held-out viewpoint. \textbf{(a)} Full view without and with correction, and the
source-camera mosaic, where each color marks the input camera that textures that
face, so inter-camera seams lie at the color boundaries. \textbf{(b)} Two zoomed
regions, a wall-panel junction (cyan) and the floor (red), each as GT, without,
and with correction. \textbf{(c)} Each input camera's $24\times16$ gain field
$\exp(g_k)$ (red brightens, blue darkens). Seam residuals are in
\cref{tab:seam}.}
\label{fig:colorcorr}
\end{figure*}

\begin{table}[t]
\centering
\scriptsize
\setlength{\tabcolsep}{6pt}
\renewcommand{\arraystretch}{1.0}
\caption{Mean absolute log-luminance mismatch at the seams between input
cameras (\SI{0.02}{\meter}, frames of \cref{tab:panoptic_quant}), inside the
solver (Constr.) and on the written atlas (Output).}
\label{tab:seam}
\begin{tabular}{lcccc}
\toprule
 & \multicolumn{2}{c}{\textit{office2}} & \multicolumn{2}{c}{\textit{band2}} \\
\cmidrule(lr){2-3}\cmidrule(lr){4-5}
Seam residual$\downarrow$ & Constr. & Output & Constr. & Output \\
\midrule
Before (Ours w/o sc) & \ccSeamBeforeA & 0.162 & \ccSeamBeforeB & 0.150 \\
After (Ours)         & \ccSeamAfterA  & 0.108 & \ccSeamAfterB  & 0.109 \\
\midrule
Reduction & \SI{36.5}{\percent} & \SI{33.3}{\percent} & \SI{31.8}{\percent} & \SI{27.3}{\percent} \\
\bottomrule
\end{tabular}
\end{table}

\subsection{Effect of Radiometric Seam Correction}
\label{sec:exp_seam}

\cref{fig:colorcorr} and \cref{tab:seam} evaluate the radiometric
seam correction of \cref{sec:colorcorr} on \textit{office2} and \textit{band2}
under the same held-out protocol. Its effect on re-rendering quality is already
in \cref{tab:panoptic_quant}, where ``Ours'' improves over ``Ours w/o sc'' in
PSNR, SSIM, and LPIPS on both sequences; one gain per camera reaches only
$30$--$72\%$ of these gains. At the seams themselves,
the correction reduces the solver's seam residual by
\SI{31.8}{}--\SI{36.5}{\percent} and that of the written atlas by
\SI{27.3}{}--\SI{33.3}{\percent} (\cref{tab:seam}); \cref{fig:colorcorr}(a,b) shows the equalized
seams as a smooth, low-frequency difference with no blotching.

The log-gain clamp ($\pm0.15$) bounds the applied gain to approximately
$[0.86,1.16]$ times the input linear RGB, and the solved fields
remain smooth and low-frequency (\cref{fig:colorcorr}(c)). With the stage
enabled, the frame time rises from \SI{10.8}{\milli\second} to \ccTotalMs~ms
(\ccAddedMs~ms end to end, mostly the \SI{1.9}{\milli\second} Seam Corr.\
row of \cref{tab:runtime_breakdown}), within the real-time budget.
Finally, correcting all
three color channels instead of luminance only changes the held-out metrics
negligibly (\cref{sec:supp_rgb}):
exposure dominates the inter-camera differences, so we keep the luminance-only
field, which preserves chroma and solves a third as many parameters. The
RGB variant uses the same solver; on our in-house capture with
independent white balance it is on par.

\subsection{Effect of Noise Removal and Inpainting}
\label{sec:exp_ablation}

\cref{fig:own} shows the effect of our noise removal and inpainting on an
in-house dataset captured with five Orbbec Femto Bolt
sensors~\cite{OrbbecFemtoBolt}. On a dynamic scene with heavy depth noise, both
stages suppress artifacts, in particular temporal flickering; see the
supplementary video. Unlike the Panoptic runs, this auxiliary test
reconstructs the foreground only.

\section{Conclusion}
\label{sec:conclusion}

We presented a GPU-native volumetric-video system for graphics-compatible
textured meshes from multiple RGB-D sensors. A fixed-size-array adjacency
keeps mesh post-processing and texturing data-parallel on the
GPU, and an on-GPU low-frequency gain field baked into the atlas reduces
inter-camera exposure seams, including mismatches that vary across each image.
The system reaches
\SI{13.1}{\milli\second} per frame at \SI{0.02}{\meter} voxels and ranks second
only to an offline baseline in LPIPS. At \SI{0.005}{\meter} voxels, the full
method is $89.7\times$/$29.4\times$ faster than Open3D (CPU, w/o nr)/nvblox.

\paragraph{Limitations and future work.}
Detail is bounded by the voxel size; finer voxels cost memory and time. Visibility is tested per vertex, not inside triangles that span a depth
discontinuity. The low-frequency gain field leaves sharp local seams from
geometric or calibration error; gradient-domain leveling is a natural
extension.

{
    \small
    \bibliographystyle{ieeenat_fullname}
    \bibliography{main}

\begin{thebibliography}{54}
\providecommand{\natexlab}[1]{#1}
\providecommand{\url}[1]{\texttt{#1}}
\expandafter\ifx\csname urlstyle\endcsname\relax
  \providecommand{\doi}[1]{doi: #1}\else
  \providecommand{\doi}{doi: \begingroup \urlstyle{rm}\Url}\fi

\bibitem[Alexiadis et~al.(2013)Alexiadis, Zarpalas, and
  Daras]{alexiadis2012real}
Dimitrios~S Alexiadis, Dimitrios Zarpalas, and Petros Daras.
\newblock Real-time, full 3-d reconstruction of moving foreground objects from
  multiple consumer depth cameras.
\newblock \emph{IEEE Transactions on Multimedia}, 15\penalty0 (2):\penalty0
  339--358, 2013.

\bibitem[Alexiadis et~al.(2017)Alexiadis, Chatzitofis, Zioulis, Zoidi, Louizis,
  Zarpalas, and Daras]{alexiadis2016integrated}
Dimitrios~S Alexiadis, Anargyros Chatzitofis, Nikolaos Zioulis, Olga Zoidi,
  Georgios Louizis, Dimitrios Zarpalas, and Petros Daras.
\newblock An integrated platform for live 3d human reconstruction and motion
  capturing.
\newblock \emph{IEEE Transactions on Circuits and Systems for Video
  Technology}, 27\penalty0 (4):\penalty0 798--813, 2017.

\bibitem[Bi et~al.(2017)Bi, Kalantari, and
  Ramamoorthi]{PatchBasedTextureMapping}
Sai Bi, Nima~Khademi Kalantari, and Ravi Ramamoorthi.
\newblock Patch-based optimization for image-based texture mapping.
\newblock \emph{ACM Transactions on Graphics (Proceedings of SIGGRAPH 2017)},
  36\penalty0 (4), 2017.

\bibitem[Blau and Michaeli(2018)]{blau2018perception}
Yochai Blau and Tomer Michaeli.
\newblock The perception-distortion tradeoff.
\newblock In \emph{Proceedings of the IEEE conference on computer vision and
  pattern recognition}, pages 6228--6237, 2018.

\bibitem[Botsch et~al.(2010)Botsch, Kobbelt, Pauly, Alliez, and
  L{\'e}vy]{botsch2010polygon}
Mario Botsch, Leif Kobbelt, Mark Pauly, Pierre Alliez, and Bruno L{\'e}vy.
\newblock \emph{Polygon mesh processing}.
\newblock CRC press, 2010.

\bibitem[{Carnegie Mellon University Graphics Lab}()]{CMUMocap}
{Carnegie Mellon University Graphics Lab}.
\newblock {CMU} graphics lab motion capture database.
\newblock \url{http://mocap.cs.cmu.edu}.
\newblock Accessed 2026-10-09.

\bibitem[Collet et~al.(2015)Collet, Chuang, Sweeney, Gillett, Evseev,
  Calabrese, Hoppe, Kirk, and Sullivan]{collet2015high}
Alvaro Collet, Ming Chuang, Pat Sweeney, Don Gillett, Dennis Evseev, David
  Calabrese, Hugues Hoppe, Adam Kirk, and Steve Sullivan.
\newblock High-quality streamable free-viewpoint video.
\newblock \emph{ACM Transactions on Graphics (ToG)}, 34\penalty0 (4):\penalty0
  69:1--69:13, 2015.

\bibitem[Curless and Levoy(1996)]{curless1996volumetric}
Brian Curless and Marc Levoy.
\newblock A volumetric method for building complex models from range images.
\newblock In \emph{Proceedings of the 23rd annual conference on Computer
  graphics and interactive techniques}, pages 303--312, 1996.

\bibitem[Dai et~al.(2017)Dai, Nie\ss{}ner, Zollh\"{o}fer, Izadi, and
  Theobalt]{10.1145/3054739}
Angela Dai, Matthias Nie\ss{}ner, Michael Zollh\"{o}fer, Shahram Izadi, and
  Christian Theobalt.
\newblock Bundlefusion: Real-time globally consistent 3d reconstruction using
  on-the-fly surface reintegration.
\newblock \emph{ACM Trans. Graph.}, 36\penalty0 (3), 2017.

\bibitem[Dou et~al.(2016)Dou, Khamis, Degtyarev, Davidson, Fanello, Kowdle,
  Escolano, Rhemann, Kim, Taylor, Kohli, Tankovich, and
  Izadi]{10.1145/2897824.2925969}
Mingsong Dou, Sameh Khamis, Yury Degtyarev, Philip Davidson, Sean~Ryan Fanello,
  Adarsh Kowdle, Sergio~Orts Escolano, Christoph Rhemann, David Kim, Jonathan
  Taylor, Pushmeet Kohli, Vladimir Tankovich, and Shahram Izadi.
\newblock Fusion4d: real-time performance capture of challenging scenes.
\newblock \emph{ACM Trans. Graph.}, 35\penalty0 (4), 2016.

\bibitem[Dou et~al.(2017)Dou, Davidson, Fanello, Khamis, Kowdle, Rhemann,
  Tankovich, and Izadi]{dou2017motion2fusion}
Mingsong Dou, Philip Davidson, Sean~Ryan Fanello, Sameh Khamis, Adarsh Kowdle,
  Christoph Rhemann, Vladimir Tankovich, and Shahram Izadi.
\newblock {Motion2Fusion}: Real-time volumetric performance capture.
\newblock \emph{ACM Transactions on Graphics (ToG)}, 36\penalty0 (6):\penalty0
  246:1--246:16, 2017.

\bibitem[Dyken et~al.(2008)Dyken, Ziegler, Theobalt, and Seidel]{dyken2008high}
Christopher Dyken, Gernot Ziegler, Christian Theobalt, and Hans-Peter Seidel.
\newblock High-speed marching cubes using histopyramids.
\newblock \emph{Computer Graphics Forum}, 27\penalty0 (8):\penalty0 2028--2039,
  2008.

\bibitem[Eck et~al.(2023)Eck, Wechner, Pankratz, Yu, Lazarovici, and
  Navab]{app131810199}
Ulrich Eck, Michael Wechner, Frieder Pankratz, Kevin Yu, Marc Lazarovici, and
  Nassir Navab.
\newblock Real-time {3D} reconstruction pipeline for room-scale, immersive,
  medical teleconsultation.
\newblock \emph{Applied Sciences}, 13\penalty0 (18):\penalty0 10199, 2023.

\bibitem[Fu et~al.(2018)Fu, Yan, Yang, Liao, and Xiao]{fu2018texture}
Yanping Fu, Qingan Yan, Long Yang, Jie Liao, and Chunxia Xiao.
\newblock Texture mapping for 3d reconstruction with rgb-d sensor.
\newblock In \emph{Proceedings of the IEEE conference on computer vision and
  pattern recognition}, pages 4645--4653, 2018.

\bibitem[Ghosh et~al.(2025)Ghosh, Shin, Zhang, Ye, Jin, Madhyastha, Netravali,
  Ortega, Rao, Rowe, and Govindan]{10.1145/3768981}
Rajrup Ghosh, Christina~Suyong Shin, Lei Zhang, Muyang Ye, Tao Jin, Harsha~V.
  Madhyastha, Ravi Netravali, Antonio Ortega, Sanjay Rao, Anthony Rowe, and
  Ramesh Govindan.
\newblock Livo: Toward bandwidth-adaptive fully-immersive volumetric video
  conferencing.
\newblock \emph{Proc. ACM Netw.}, 3\penalty0 (CoNEXT4), 2025.

\bibitem[Ha et~al.(2021)Ha, Lee, Meuleman, and Kim]{ha2021normalfusion}
Hyunho Ha, Joo~Ho Lee, Andreas Meuleman, and Min~H Kim.
\newblock {NormalFusion}: Real-time acquisition of surface normals for
  high-resolution {RGB-D} scanning.
\newblock In \emph{Proceedings of the IEEE/CVF Conference on Computer Vision
  and Pattern Recognition}, pages 15970--15979, 2021.

\bibitem[Han et~al.(2022)Han, Gu, Zhong, Quan, and Fang]{han2022real}
Lei Han, Siyuan Gu, Dawei Zhong, Shuxue Quan, and Lu Fang.
\newblock Real-time globally consistent dense 3d reconstruction with online
  texturing.
\newblock \emph{IEEE Transactions on Pattern Analysis and Machine
  Intelligence}, 44\penalty0 (3):\penalty0 1519--1533, 2022.

\bibitem[Hawick et~al.(2010)Hawick, Leist, and Playne]{hawick2010parallel}
Kenneth~A Hawick, Arno Leist, and Daniel~P Playne.
\newblock Parallel graph component labelling with gpus and cuda.
\newblock \emph{Parallel Computing}, 36\penalty0 (12):\penalty0 655--678, 2010.

\bibitem[Jeon et~al.(2016)Jeon, Jung, Kim, and Lee]{jeon2016texture}
Junho Jeon, Yeongyu Jung, Haejoon Kim, and Seungyong Lee.
\newblock Texture map generation for 3d reconstructed scenes.
\newblock \emph{The Visual Computer}, 32\penalty0 (6):\penalty0 955--965, 2016.

\bibitem[Jin et~al.(2024)Jin, Dasari, Smith, Apicharttrisorn, Seshan, and
  Rowe]{10494107}
Tao Jin, Mallesham Dasari, Connor Smith, Kittipat Apicharttrisorn, Srinivasan
  Seshan, and Anthony Rowe.
\newblock Meshreduce: Scalable and bandwidth efficient 3d scene capture.
\newblock In \emph{2024 IEEE Conference Virtual Reality and 3D User Interfaces
  (VR)}, pages 20--30, 2024.

\bibitem[Joo et~al.(2015)Joo, Liu, Tan, Gui, Nabbe, Matthews, Kanade, Nobuhara,
  and Sheikh]{Joo_2015_ICCV}
Hanbyul Joo, Hao Liu, Lei Tan, Lin Gui, Bart Nabbe, Iain Matthews, Takeo
  Kanade, Shohei Nobuhara, and Yaser Sheikh.
\newblock Panoptic studio: A massively multiview system for social motion
  capture.
\newblock In \emph{ICCV}, 2015.

\bibitem[Joo et~al.(2019)Joo, Simon, Li, Liu, Tan, Gui, Banerjee, Godisart,
  Nabbe, Matthews, Kanade, Nobuhara, and Sheikh]{Joo_2017_TPAMI}
Hanbyul Joo, Tomas Simon, Xulong Li, Hao Liu, Lei Tan, Lin Gui, Sean Banerjee,
  Timothy~Scott Godisart, Bart Nabbe, Iain Matthews, Takeo Kanade, Shohei
  Nobuhara, and Yaser Sheikh.
\newblock Panoptic studio: A massively multiview system for social interaction
  capture.
\newblock \emph{IEEE Transactions on Pattern Analysis and Machine
  Intelligence}, 41\penalty0 (1):\penalty0 190--204, 2019.

\bibitem[Kazhdan and Hoppe(2013)]{kazhdan2013screened}
Michael Kazhdan and Hugues Hoppe.
\newblock Screened poisson surface reconstruction.
\newblock \emph{ACM Transactions on Graphics (ToG)}, 32\penalty0 (3):\penalty0
  1--13, 2013.

\bibitem[Kazhdan et~al.(2006)Kazhdan, Bolitho, and Hoppe]{kazhdan2006poisson}
Michael Kazhdan, Matthew Bolitho, and Hugues Hoppe.
\newblock Poisson surface reconstruction.
\newblock In \emph{Eurographics Symposium on Geometry Processing}, pages
  61--70, 2006.

\bibitem[Kerbl et~al.(2023)Kerbl, Kopanas, Leimk{\"u}hler, and
  Drettakis]{kerbl3Dgaussians}
Bernhard Kerbl, Georgios Kopanas, Thomas Leimk{\"u}hler, and George Drettakis.
\newblock 3d gaussian splatting for real-time radiance field rendering.
\newblock \emph{ACM Transactions on Graphics}, 42\penalty0 (4), 2023.

\bibitem[Kim et~al.(2022)Kim, Kim, Nam, Park, and Lee]{10.1145/3503926}
Jungeon Kim, Hyomin Kim, Hyeonseo Nam, Jaesik Park, and Seungyong Lee.
\newblock Textureme: High-quality textured scene reconstruction in real time.
\newblock \emph{ACM Trans. Graph.}, 41\penalty0 (3), 2022.

\bibitem[Lee et~al.(2020)Lee, Ha, Dong, Tong, and Kim]{lee2020texturefusion}
Joo~Ho Lee, Hyunho Ha, Yue Dong, Xin Tong, and Min~H Kim.
\newblock Texturefusion: High-quality texture acquisition for real-time rgb-d
  scanning.
\newblock In \emph{Proceedings of the IEEE/CVF Conference on Computer Vision
  and Pattern Recognition}, pages 1272--1280, 2020.

\bibitem[Lempitsky and Ivanov(2007)]{lempitsky2007seamless}
Victor Lempitsky and Denis Ivanov.
\newblock Seamless mosaicing of image-based texture maps.
\newblock In \emph{2007 IEEE conference on computer vision and pattern
  recognition}, pages 1--6. IEEE, 2007.

\bibitem[Loper et~al.(2015)Loper, Mahmood, Romero, Pons-Moll, and
  Black]{SMPL:2015}
Matthew Loper, Naureen Mahmood, Javier Romero, Gerard Pons-Moll, and Michael~J.
  Black.
\newblock {SMPL}: A skinned multi-person linear model.
\newblock \emph{ACM Trans. Graphics (Proc. SIGGRAPH Asia)}, 34\penalty0
  (6):\penalty0 248:1--248:16, 2015.

\bibitem[Lorensen and Cline(1987)]{10.1145/37402.37422}
William~E. Lorensen and Harvey~E. Cline.
\newblock Marching cubes: A high resolution 3d surface construction algorithm.
\newblock \emph{SIGGRAPH Comput. Graph.}, 21\penalty0 (4):\penalty0 163–169,
  1987.

\bibitem[Maier et~al.(2017)Maier, Kim, Cremers, Kautz, and
  Nie{\ss}ner]{maier2017intrinsic3d}
Robert Maier, Kihwan Kim, Daniel Cremers, Jan Kautz, and Matthias Nie{\ss}ner.
\newblock Intrinsic3d: High-quality 3d reconstruction by joint appearance and
  geometry optimization with spatially-varying lighting.
\newblock In \emph{Proceedings of the IEEE international conference on computer
  vision}, pages 3114--3122, 2017.

\bibitem[Meerits et~al.(2018)Meerits, Thomas, Nozick, and Saito]{10897380}
Siim Meerits, Diego Thomas, Vincent Nozick, and Hideo Saito.
\newblock {FusionMLS}: Highly dynamic {3D} reconstruction with consumer-grade
  {RGB-D} cameras.
\newblock \emph{Computational Visual Media}, 4\penalty0 (4):\penalty0 287--303,
  2018.

\bibitem[{Microsoft}(2014)]{KinectV2}
{Microsoft}.
\newblock Kinect for windows v2: Sensor and data sources overview.
\newblock
  \url{https://learn.microsoft.com/en-us/shows/visual-studio-connect-event-2014/716},
  2014.

\bibitem[Mildenhall et~al.(2020)Mildenhall, Srinivasan, Tancik, Barron,
  Ramamoorthi, and Ng]{mildenhall2020nerf}
Ben Mildenhall, Pratul~P. Srinivasan, Matthew Tancik, Jonathan~T. Barron, Ravi
  Ramamoorthi, and Ren Ng.
\newblock Nerf: Representing scenes as neural radiance fields for view
  synthesis.
\newblock In \emph{ECCV}, 2020.

\bibitem[Millane et~al.(2024)Millane, Oleynikova, Wirbel, Steiner, Ramasamy,
  Tingdahl, and Siegwart]{millane2024nvblox}
Alexander Millane, Helen Oleynikova, Emilie Wirbel, Remo Steiner, Vikram
  Ramasamy, David Tingdahl, and Roland Siegwart.
\newblock nvblox: Gpu-accelerated incremental signed distance field mapping.
\newblock In \emph{2024 IEEE International Conference on Robotics and
  Automation (ICRA)}, pages 2698--2705. IEEE, 2024.

\bibitem[Newcombe et~al.(2011)Newcombe, Izadi, Hilliges, Molyneaux, Kim,
  Davison, Kohli, Shotton, Hodges, and Fitzgibbon]{newcombe2011kinectfusion}
Richard~A Newcombe, Shahram Izadi, Otmar Hilliges, David Molyneaux, David Kim,
  Andrew~J Davison, Pushmeet Kohli, Jamie Shotton, Steve Hodges, and Andrew
  Fitzgibbon.
\newblock Kinectfusion: Real-time dense surface mapping and tracking.
\newblock In \emph{2011 10th IEEE international symposium on mixed and
  augmented reality}, pages 127--136. Ieee, 2011.

\bibitem[Nie{\ss}ner et~al.(2013)Nie{\ss}ner, Zollh{\"o}fer, Izadi, and
  Stamminger]{niessner2013real}
Matthias Nie{\ss}ner, Michael Zollh{\"o}fer, Shahram Izadi, and Marc
  Stamminger.
\newblock Real-time 3d reconstruction at scale using voxel hashing.
\newblock \emph{ACM Transactions on Graphics (ToG)}, 32\penalty0 (6):\penalty0
  1--11, 2013.

\bibitem[{NVIDIA Corporation}()]{cuda_programming_guide}
{NVIDIA Corporation}.
\newblock {CUDA Programming Guide}.
\newblock \url{https://docs.nvidia.com/cuda/cuda-programming-guide/}.
\newblock Accessed: 2026-03-12.

\bibitem[{Orbbec Inc.}(2023)]{OrbbecFemtoBolt}
{Orbbec Inc.}
\newblock Femto bolt: Time-of-flight rgb-d camera.
\newblock \url{https://www.orbbec.com/products/tof-camera/femto-bolt/}, 2023.
\newblock Time-of-flight RGB-D sensor.

\bibitem[Orts-Escolano et~al.(2016)Orts-Escolano, Rhemann, Fanello, Chang,
  Kowdle, Degtyarev, Kim, Davidson, Khamis, Dou, Tankovich, Loop, Cai, Chou,
  Mennicken, Valentin, Pradeep, Wang, Kang, Kohli, Lutchyn, Keskin, and
  Izadi]{10.1145/2984511.2984517}
Sergio Orts-Escolano, Christoph Rhemann, Sean Fanello, Wayne Chang, Adarsh
  Kowdle, Yury Degtyarev, David Kim, Philip~L. Davidson, Sameh Khamis, Mingsong
  Dou, Vladimir Tankovich, Charles Loop, Qin Cai, Philip~A. Chou, Sarah
  Mennicken, Julien Valentin, Vivek Pradeep, Shenlong Wang, Sing~Bing Kang,
  Pushmeet Kohli, Yuliya Lutchyn, Cem Keskin, and Shahram Izadi.
\newblock Holoportation: Virtual 3d teleportation in real-time.
\newblock In \emph{Proceedings of the 29th Annual Symposium on User Interface
  Software and Technology}, page 741–754, New York, NY, USA, 2016.
  Association for Computing Machinery.

\bibitem[Prada et~al.(2016)Prada, Kazhdan, Chuang, Collet, and
  Hoppe]{prada2016motion}
Fabi{\'a}n Prada, Misha Kazhdan, Ming Chuang, Alvaro Collet, and Hugues Hoppe.
\newblock Motion graphs for unstructured textured meshes.
\newblock \emph{ACM Transactions on Graphics (ToG)}, 35\penalty0 (4):\penalty0
  108:1--108:14, 2016.

\bibitem[Prada et~al.(2017)Prada, Kazhdan, Chuang, Collet, and
  Hoppe]{prada2017spatiotemporal}
Fabi{\'a}n Prada, Misha Kazhdan, Ming Chuang, Alvaro Collet, and Hugues Hoppe.
\newblock Spatiotemporal atlas parameterization for evolving meshes.
\newblock \emph{ACM Transactions on Graphics (ToG)}, 36\penalty0 (4):\penalty0
  58:1--58:12, 2017.

\bibitem[Sch\"ops et~al.(2017)Sch\"ops, Sch\"onberger, Galliani, Sattler,
  Schindler, Pollefeys, and Geiger]{schoeps2017cvpr}
Thomas Sch\"ops, Johannes~L. Sch\"onberger, Silvano Galliani, Torsten Sattler,
  Konrad Schindler, Marc Pollefeys, and Andreas Geiger.
\newblock A multi-view stereo benchmark with high-resolution images and
  multi-camera videos.
\newblock In \emph{Conference on Computer Vision and Pattern Recognition
  (CVPR)}, 2017.

\bibitem[Sterzentsenko et~al.(2018)Sterzentsenko, Karakottas, Papachristou,
  Zioulis, Doumanoglou, Zarpalas, and Daras]{sterzentsenko2018low}
Vladimiros Sterzentsenko, Antonis Karakottas, Alexandros Papachristou, Nikolaos
  Zioulis, Alexandros Doumanoglou, Dimitrios Zarpalas, and Petros Daras.
\newblock A low-cost, flexible and portable volumetric capturing system.
\newblock In \emph{2018 14th International Conference on Signal-Image
  Technology \& Internet-Based Systems (SITIS)}, pages 200--207. IEEE, 2018.

\bibitem[Waechter et~al.(2014)Waechter, Moehrle, and
  Goesele]{Waechter2014Texturing}
Michael Waechter, Nils Moehrle, and Michael Goesele.
\newblock Let there be color! --- {L}arge-scale texturing of {3D}
  reconstructions.
\newblock In \emph{Proceedings of the European Conference on Computer Vision}.
  Springer, 2014.

\bibitem[Wang et~al.(2025)Wang, Chen, Karaev, Vedaldi, Rupprecht, and
  Novotny]{wang2025vggt}
Jianyuan Wang, Minghao Chen, Nikita Karaev, Andrea Vedaldi, Christian
  Rupprecht, and David Novotny.
\newblock Vggt: Visual geometry grounded transformer.
\newblock In \emph{Proceedings of the IEEE/CVF Conference on Computer Vision
  and Pattern Recognition}, 2025.

\bibitem[Wang et~al.(2004)Wang, Bovik, Sheikh, and Simoncelli]{wang2004image}
Zhou Wang, Alan~C Bovik, Hamid~R Sheikh, and Eero~P Simoncelli.
\newblock Image quality assessment: from error visibility to structural
  similarity.
\newblock \emph{IEEE transactions on image processing}, 13\penalty0
  (4):\penalty0 600--612, 2004.

\bibitem[Yu et~al.(2018)Yu, Zheng, Guo, Zhao, Dai, Li, Pons-Moll, and
  Liu]{yu2018doublefusion}
Tao Yu, Zerong Zheng, Kaiwen Guo, Jianhui Zhao, Qionghai Dai, Hao Li, Gerard
  Pons-Moll, and Yebin Liu.
\newblock Doublefusion: Real-time capture of human performances with inner body
  shapes from a single depth sensor.
\newblock In \emph{Proceedings of the IEEE conference on computer vision and
  pattern recognition}, pages 7287--7296, 2018.

\bibitem[Yu et~al.(2021)Yu, Zheng, Guo, Liu, Dai, and Liu]{yu2021function4d}
Tao Yu, Zerong Zheng, Kaiwen Guo, Pengpeng Liu, Qionghai Dai, and Yebin Liu.
\newblock Function4d: Real-time human volumetric capture from very sparse
  consumer rgbd sensors.
\newblock In \emph{Proceedings of the IEEE/CVF conference on computer vision
  and pattern recognition}, pages 5746--5756, 2021.

\bibitem[Zhang et~al.(2018)Zhang, Isola, Efros, Shechtman, and
  Wang]{zhang2018unreasonable}
Richard Zhang, Phillip Isola, Alexei~A Efros, Eli Shechtman, and Oliver Wang.
\newblock The unreasonable effectiveness of deep features as a perceptual
  metric.
\newblock In \emph{Proceedings of the IEEE conference on computer vision and
  pattern recognition}, pages 586--595, 2018.

\bibitem[Zhou and Koltun(2013)]{10.1145/2461912.2461919}
Qian-Yi Zhou and Vladlen Koltun.
\newblock Dense scene reconstruction with points of interest.
\newblock \emph{ACM Trans. Graph.}, 32\penalty0 (4), 2013.

\bibitem[Zhou and Koltun(2014)]{zhou2014color}
Qian-Yi Zhou and Vladlen Koltun.
\newblock Color map optimization for 3d reconstruction with consumer depth
  cameras.
\newblock \emph{ACM Transactions on Graphics (ToG)}, 33\penalty0 (4):\penalty0
  1--10, 2014.

\bibitem[Zhou et~al.(2018)Zhou, Park, and Koltun]{Zhou2018}
Qian-Yi Zhou, Jaesik Park, and Vladlen Koltun.
\newblock {Open3D}: {A} modern library for {3D} data processing.
\newblock \emph{arXiv:1801.09847}, 2018.

\bibitem[Zuo et~al.(2021)Zuo, Wang, Zheng, Yu, Gong, Yang, and Cheng]{9113759}
Xinxin Zuo, Sen Wang, Jiangbin Zheng, Weiwei Yu, Minglun Gong, Ruigang Yang,
  and Li Cheng.
\newblock Sparsefusion: Dynamic human avatar modeling from sparse rgbd images.
\newblock \emph{IEEE Transactions on Multimedia}, 23:\penalty0 1617--1629,
  2021.

\end{thebibliography}
}

\clearpage
\twocolumn[{\centering\Large\bf Appendix: Supplementary Material\par\vspace{1.5em}}]
\appendix
This appendix provides (i) implementation details and the full hyperparameter
list (\cref{sec:supp_hyper}), (ii) the held-out re-rendering evaluation protocol
(\cref{sec:supp_protocol}), (iii) ablations of the seam correction (one gain per camera, luminance vs.\
RGB) on Panoptic, on a synthetic capture with ground truth, and on our
in-house five-camera capture
(\cref{sec:supp_rgb}), (iv) a comparison against applying an offline texturing
method to our own geometry (\cref{sec:supp_offtex}), (v) the
quality/runtime/memory tradeoff of finer voxel sizes (\cref{sec:supp_voxel}),
(vi) an ablation of the depth tolerance in occlusion-aware view selection
(\cref{sec:supp_occl}), (vii) an SfM+MVS application example
(\cref{sec:supp_sfm}), and (viii) a statement on the use of large language
models (\cref{sec:supp_llm}). An
accompanying video (project page: \url{https://unclearness.github.io/grtmesh/}) shows (1) the seam
correction on the $100$ held-out frames of the evaluation in \cref{sec:exp_quality}
(\textit{office2}, frames $400$--$697$ at stride $3$, camera 7), playing the
re-rendering without (left) and with (right) correction; (2) the texture
atlas and the solved gain fields over time; (3) free-view comparisons of our
output with nvblox and the offline baseline on \textit{office2} and
\textit{band2} over $800$ continuous frames; and (4) the noise-removal,
inpainting, and seam-correction ablations on the in-house capture
(\cref{sec:supp_rgb}) with its luminance and RGB gain fields. The video shows
the temporal behavior of the correction; the fixed \SI{10}{\centi\meter}
evaluation is reported in \cref{sec:exp_quality} and the ablations below.

\section{Implementation Details and Hyperparameters}
\label{sec:supp_hyper}

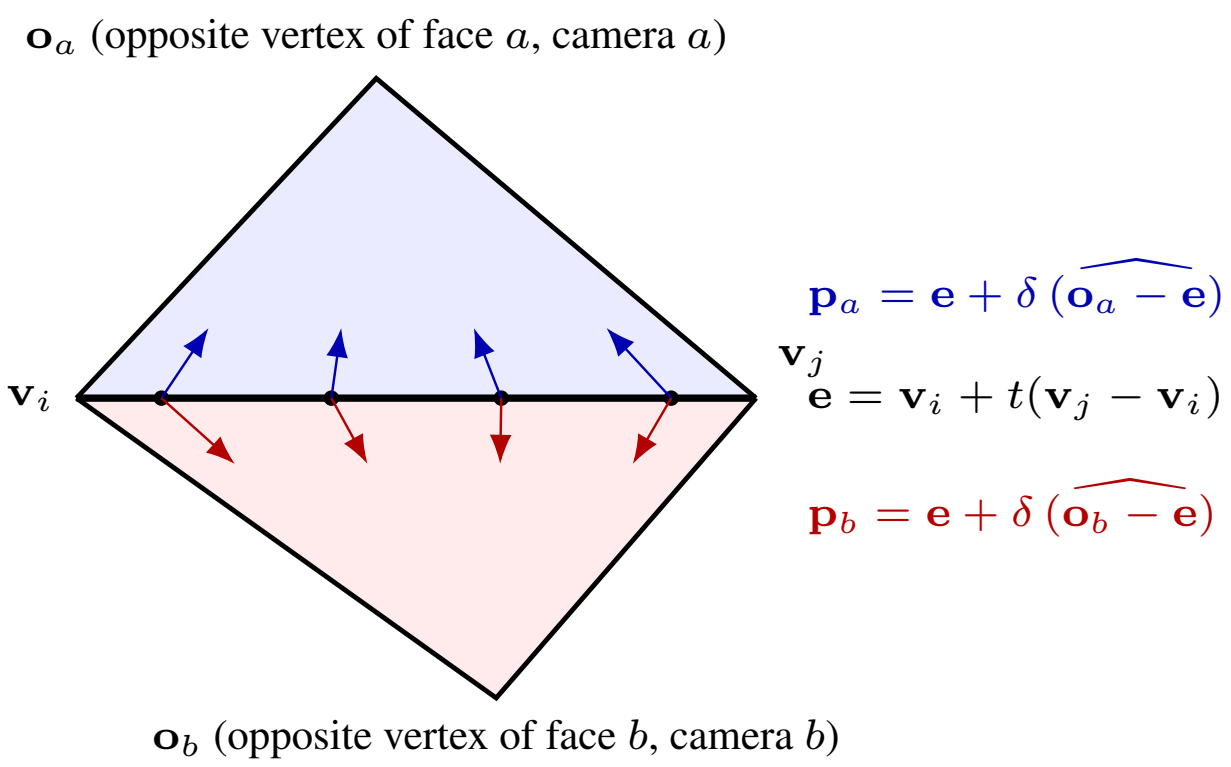
\begin{figure}[t]
\centering
\resizebox{\linewidth}{!}{%
\begin{tikzpicture}[scale=1.15,>=Latex,font=\scriptsize]
  \coordinate (vi) at (0,0);
  \coordinate (vj) at (3.4,0);
  \coordinate (oa) at (1.5,1.6);
  \coordinate (ob) at (2.1,-1.5);
  \fill[blue!8] (vi) -- (vj) -- (oa) -- cycle;
  \fill[red!8]  (vi) -- (vj) -- (ob) -- cycle;
  \draw[thick] (vi) -- (oa) -- (vj);
  \draw[thick] (vi) -- (ob) -- (vj);
  \draw[very thick] (vi) -- (vj);
  \node[left]  at (vi) {$\mathbf{v}_i$};
  \node[above right] at (vj) {$\mathbf{v}_j$};
  \node[above] at (oa) {$\mathbf{o}_a$ (opposite vertex of face $a$, camera $a$)};
  \node[below] at (ob) {$\mathbf{o}_b$ (opposite vertex of face $b$, camera $b$)};
  \foreach \t in {0.125,0.375,0.625,0.875} {
    \coordinate (e) at ($(vi)!\t!(vj)$);
    \fill (e) circle (1.1pt);
    \draw[->,blue!70!black] (e) -- ($(e)!0.22!(oa)$);
    \draw[->,red!70!black]  (e) -- ($(e)!0.22!(ob)$);
  }
  \node[blue!70!black,anchor=west] at (3.55,0.55) {$\mathbf{p}_a=\mathbf{e}+\delta\,\widehat{(\mathbf{o}_a-\mathbf{e})}$};
  \node[red!70!black,anchor=west]  at (3.55,-0.55) {$\mathbf{p}_b=\mathbf{e}+\delta\,\widehat{(\mathbf{o}_b-\mathbf{e})}$};
  \node[anchor=west] at (3.55,0) {$\mathbf{e}=\mathbf{v}_i+t(\mathbf{v}_j-\mathbf{v}_i)$};
\end{tikzpicture}%
}
\caption{Seam-residual sampling geometry (\cref{sec:colorcorr}). A seam is the
edge $\mathbf{v}_i\mathbf{v}_j$ shared by face $a$ (textured from camera $a$)
and face $b$ (camera $b$). For each of the $N=8$ sample points
$\mathbf{e}$ at $t=(i+\tfrac12)/N$ along the edge, both faces are sampled at
corresponding shared-edge positions $\mathbf{e}$, and the image-space sample position
is inset by $\delta$ from $\mathbf{e}$ toward the projection of that face's
opposite vertex ($\mathbf{o}_a$ or $\mathbf{o}_b$), \emph{not} along the edge
normal. The inset is applied independently in each camera's projected
coordinates, with $\delta=0.5$ proxy pixels ($=4$ full-resolution pixels at
the $1/8$ proxy), moving the samples toward the interiors of their respective projected faces.
The inset positions need not correspond to the same 3D point. Dark or invalid
samples are rejected, and
the seam residual $d_s$ is the median over the remaining samples of
$\log Y_b-\log Y_a$; seams whose faces fold sharply
($\mathbf{n}_a\!\cdot\!\mathbf{n}_b<0.5$) or project degenerately are
discarded.}
\label{fig:supp_inset}
\end{figure}

All runtime and quality numbers were obtained on a workstation with an AMD Ryzen
Threadripper 9970X, \SI{256}{GB} of RAM, and an NVIDIA RTX PRO 6000 Blackwell
Max-Q. \cref{tab:supp_hyper} lists the Panoptic defaults of \cref{sec:experiments}; ablations and the
auxiliary capture specify their overrides. We stabilize the seam correction gain
field across frames with an exponential moving average,
$\mathbf{g}_t \leftarrow \alpha\,\mathbf{g}_t + (1-\alpha)\,\mathbf{g}_{t-1}$,
with $\alpha=0.15$ and the zero-mean gauge re-applied after the blend.
\cref{fig:supp_inset} illustrates how the seam residuals are sampled, including
the direction and magnitude of the sub-pixel inset.

\paragraph{Runtime protocol.}
Runtimes are measured on frames $400$--$2399$ (\textit{band2}: $1397$--$3396$) and reported as the mean over
the fastest $95\%$ of the timed frames, which removes start-up and
scheduling outliers. Frames are decoded into host memory before timing; our
times start with the batched upload of the raw color and depth images and
include their undistortion and depth-to-color registration on the GPU
(\SI{2.2}{\milli\second} together at \SI{0.02}{\meter}). The baselines
perform these two steps on the CPU, outside their timed region. Over all 2000
frames without the trimming, the full method takes \SI{13.2}{\milli\second}
on average at \SI{0.02}{\meter} ($95$th percentile \SI{15.2}{\milli\second}),
against \SI{13.1}{\milli\second} in \cref{tab:runtime_breakdown}.

At \SI{0.01}{\meter} and \SI{0.005}{\meter} the baselines are timed on every
fourth and every tenth frame of the same range ($500$ and $200$ frames; nvblox
at \SI{0.005}{\meter}: the first $60$ of these).
nvblox (nvblox\_torch 0.0.9) does not release GPU memory when its mapper is
cleared, so memory grows until the device is full and the following frames
stall. We therefore recreate the mapper outside the timed region whenever
device memory use exceeds $50\%$ ($90\%$ at \SI{0.005}{\meter}). The first frames after each recreation re-grow the memory
pool and stall. At \SI{0.02}{\meter} and coarser, these stalls fall in the
trimmed $5\%$. At \SI{0.01}{\meter} we use frames $18$--$57$ after each
recreation ($291$ of $500$, fastest-$95\%$ mean). At \SI{0.005}{\meter},
memory fills within five timed frames, so we use the single regrowth-free frame
of each segment ($12$ of $60$ timed frames, plain mean). Excluding the stalls favors nvblox. On
exactly the same frames, our method is $23.7\times$ and $29.4\times$ faster
than nvblox at \SI{0.01}{\meter} and \SI{0.005}{\meter}, as with the reported
means.

Open3D (CUDA) is the tensor \texttt{VoxelBlockGrid} of Open3D 0.19 on the GPU
(block resolution $16$, truncation of four voxels, mesh weight threshold $0.5$;
the default of $3$ removes most of the surface). Its time covers the image
upload, block allocation and integration per camera, Marching Cubes, and the
mesh download. Its \SIrange{0.02}{0.16}{\meter} runs were repeated back to back
because the first sweep drifted between jobs; the two sweeps differ by up to
\SI{1.9}{\milli\second} per setting.

\paragraph{Kernel launches per frame.}
The launch counts in the pipeline figure (\cref{fig:pipeline}) were
measured with Nsight Systems on \textit{office2} (ten frames, five Kinects,
\SI{0.02}{\meter}): TSDF fusion $1$; Marching Cubes $12$; noise removal
$N_{\mathrm{nr}}=100$ label-propagation launches (the iteration cap, which the
room-scale meshes reach in every frame; a convergence flag is read back every
ten iterations) plus about $28$ launches for edge sorting, adjacency
construction, and compaction, and $3$ for normal computation and smoothing;
texture mapping $2$ (view selection and UV generation) plus $13$ launches that
rebuild the fixed-size-array adjacency of the compacted mesh (edge sort); inpainting
$N_{\mathrm{ip}}=5$ (the cap, reached in every frame); and seam correction
$N_{\mathrm{sc}}=55$: $12$ Jacobi iterations of four kernels (clear, splat
seams, update, gauge) plus proxy construction, seam detection, residual
sampling, the temporal EMA blend and its gauge re-application, residual
statistics, and the bake ($53$ on the first frame, which has no EMA history).
In total, about $219$ kernel launches per frame. The GPU input preprocessing
adds $25$ more (five per camera: color and depth undistortion, z-buffer
clear, depth reprojection into the color view, and upsampling), $244$ in all.

\paragraph{Noise-removal iteration cap.}
The cap of $100$ label-propagation iterations bounds the latency of noise
removal, but on our room-scale meshes propagation has not yet converged when
it stops. We reran $30$ frames ($400$--$690$, stride $10$) with a cap of
$5000$ and the same convergence check: propagation converges after
$111$--$241$ iterations at \SI{0.02}{\meter} and $241$--$591$ at
\SI{0.01}{\meter} (\textit{office2} and \textit{band2}), and after $791$ to
more than $5000$ at \SI{0.005}{\meter} (\textit{office2}). With the cap, a large
component can still carry several labels, and a part with fewer than $500$
faces is removed with the noise. Compared with the long runs, the cap removes
on average $0.02$--$0.05\%$ and at most $0.12\%$ of the output faces per frame.
The component analysis is therefore an approximation with a bounded number of
iterations; all reported results use the cap.

\begin{figure}[t]
\centering
\footnotesize
\setlength{\tabcolsep}{3pt}
\begin{tabular}{@{}cc@{}}
\begin{tabular}[b]{@{}c@{}}
  \includegraphics[width=0.27\linewidth]{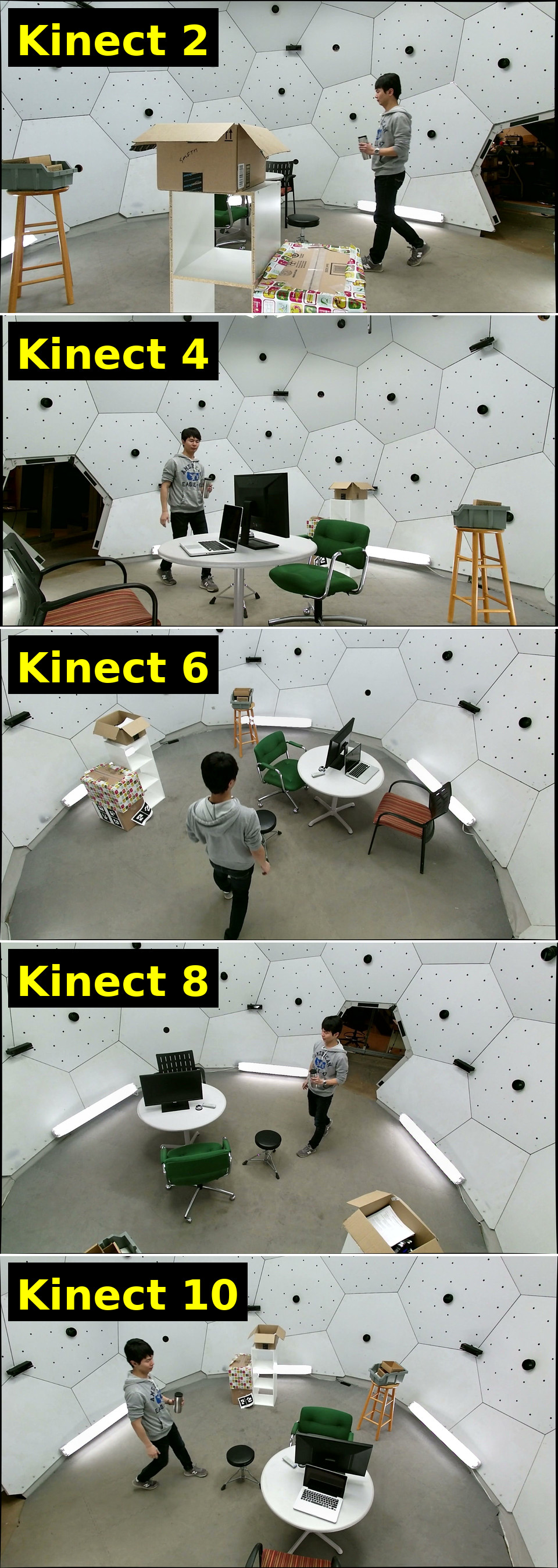} \\
  (a) texture atlas
\end{tabular} &
\begin{tabular}[b]{@{}c@{}}
  \includegraphics[width=0.68\linewidth]{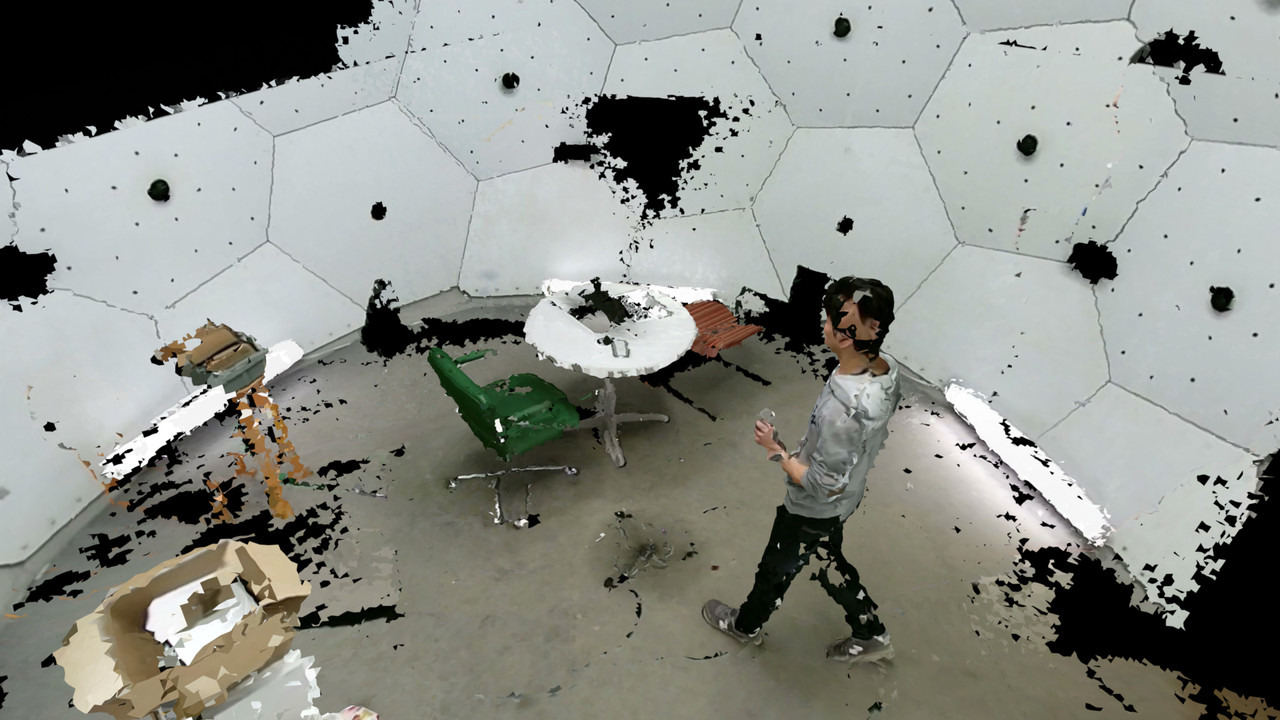} \\
  (b) textured mesh, held-out Kinect 7 \\[3pt]
  \includegraphics[width=0.68\linewidth]{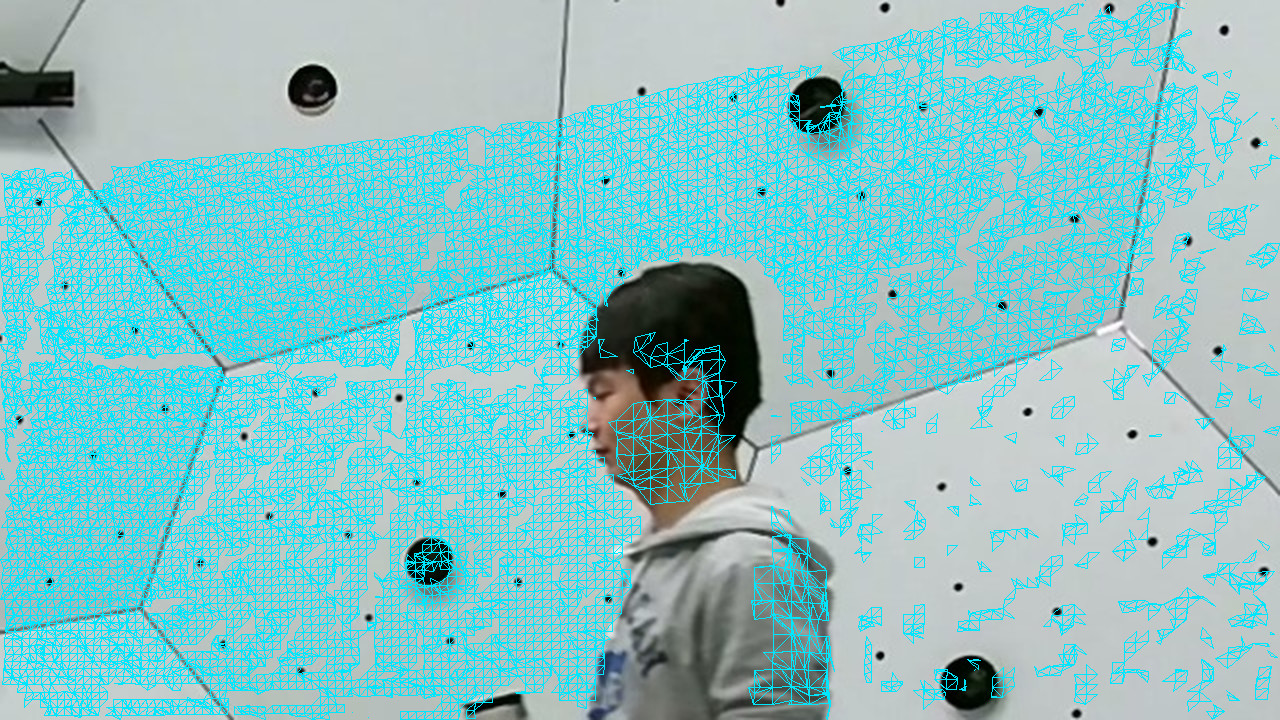} \\
  (c) UV triangles in a crop of the Kinect 2 band
\end{tabular}
\end{tabular}
\caption{Texture atlas of our output (\textit{office2}, frame 445,
\SI{0.02}{\meter}, seam correction on). \textbf{(a)} The atlas is the stack of
the five input color images, one band per Kinect, with the per-camera gain
baked in. \textbf{(b)} The mesh textured from it, rendered from a held-out
camera. \textbf{(c)} A crop of the top band with the UV triangles of the faces
that sample it: UVs are pixel coordinates in the source image, so the mesh
reads the texture exactly where the camera saw it.}
\label{fig:supp_atlas}
\end{figure}

\paragraph{Texture atlas.}
The atlas is the vertical stack of the input color images, one
$1920\times1080$ band per Kinect in input order, i.e.\ $1920\times5400$ pixels
for five Kinects (\cref{fig:supp_atlas}). A face textured from camera $k$
receives as UVs the pixel coordinates of its projection into band $k$;
an inpainted face receives the UV of one vertex of a neighboring face and is
therefore filled with a uniform color; and the seam correction multiplies band
$k$ by $\exp(\hat g_k)$ in place, so the mesh UVs never change. Image regions
that no face samples simply stay in the atlas: the layout needs no chart
packing, and the corrected atlas of the frame shown is a \SI{1.9}{MB} JPEG.
The layout also suits real-time delivery: it is fixed and each band is the
camera image itself, so consecutive atlases keep a constant resolution and are
temporally coherent, and the atlas sequence can be streamed with a standard
video encoder (compression is not evaluated here).

\begin{table}[t]
\centering
\footnotesize
\setlength{\tabcolsep}{4pt}
\renewcommand{\arraystretch}{1.05}
\caption{Default hyperparameters for the Panoptic evaluation of \cref{sec:experiments}. Overrides for ablations and the auxiliary capture are stated in their respective sections.}
\label{tab:supp_hyper}
\resizebox{\linewidth}{!}{%
\begin{tabular}{@{}ll@{}}
\toprule
Stage / parameter & Value \\
\midrule
\multicolumn{2}{@{}l}{\textit{Reconstruction}} \\
Voxel size & \SI{0.02}{\meter} \\
SDF truncation band & $4\times$ voxel size \\
\midrule
\multicolumn{2}{@{}l}{\textit{Mesh post-processing}} \\
Noise removal: min.\ component & $500$ faces \\
Noise removal: max.\ iterations & $100$ \\
Inpainting: max.\ iterations & $5$ \\
Texture view selection: max.\ angle & $80^{\circ}$ \\
Texture view selection: depth tolerance & \SI{10}{\centi\meter} \\
\midrule
\multicolumn{2}{@{}l}{\textit{Radiometric seam correction}} \\
Per-camera gain grid & $24\times16$ \\
Color space & luminance (scalar) \\
Solver & weighted Jacobi \\
Relaxation $\omega$ & $0.7$ \\
Iterations & $12$ \\
Smoothness weight $\lambda_{\mathrm{sm}}$ & $30$ \\
Anchor weight $\lambda_{\mathrm{a}}$ & $0.02$ \\
Log-gain clamp $g_{\max}$ & $0.15$ ($\exp\hat g_k\!\in\![0.86,1.16]$) \\
Residual proxy downscale & $1/8$ (box) \\
Samples per seam & $8$ \\
Sub-pixel inset & $0.5$ proxy px, toward the opposite vertex \\
Fold/veil cutoff $\tau$ ($\mathbf{n}_a\!\cdot\!\mathbf{n}_b$) & $0.5$ \\
Gauge & zero-mean over cameras \\
Temporal EMA $\alpha$ & $0.15$ \\
\bottomrule
\end{tabular}%
}
\end{table}

\subsection{Seam Correction Sampling and Solver}
\label{sec:supp_cc_solver}

\Cref{sec:colorcorr} gives the seam-to-field formulation. Here we detail its
implementation. We read seams directly from the fixed-size-array face adjacency,
processing each face pair once. A candidate is discarded if the face-normal
dot product is below $0.5$, or if either projected triangle has twice-area
below one full-resolution pixel squared (including collapsed inpainted faces).
For each retained seam we sample eight edge parameters
$t_i=(i+\tfrac12)/8$. In each camera, the projected point is moved by $0.5$
proxy pixels toward that face's projected opposite vertex, as illustrated in
\cref{fig:supp_inset}. These positions lie near corresponding shared-edge
locations; after the independent insets they need not be projections of the
same 3D point. Samples outside the proxy or with linear luminance below $0.01$
in either view are rejected. The median log-luminance difference supplies
$d_s$, and the valid-sample fraction supplies $w_s$. There is no additional
robust penalty on the resulting least-squares residual.

\paragraph{Proxy and discrete constraints.}
We build the linear-RGB proxy by $1/8$ box reduction of each input camera band.
This reduces the influence of texture detail and small misalignments on the
residual. One constraint per seam is assigned to the nearest grid cell at the
inset midpoint in each camera. Bilinear assignment would couple multiple
same-camera unknowns in each data term; the nearest-cell construction instead
retains a graph of pairwise differences. Bilinear interpolation is used only
when applying the solved field to the atlas.

\paragraph{Matrix-free update.}
We use the notation $g_k[p]$, $a(s)$, $b(s)$, $p_a^s$, $p_b^s$ of \cref{sec:colorcorr}.
For cell $p$ of camera $k$, let $\mathcal N(p)$ be its four-neighbors in the
same grid and $\mathcal S(p)$ the seams assigned to it, i.e.\ with
$(a(s),p_a^s)=(k,p)$ or $(b(s),p_b^s)=(k,p)$. For such a seam let
$(k'_s,q_s)$ be the other camera and its cell, and let $\sigma_{p,s}$ be $+1$
if $k=a(s)$ and $-1$ if $k=b(s)$. The update is
\begin{equation}
\begin{split}
\mathrm{num}_p &= \lambda_{\mathrm{sm}}\sum_{q\in\mathcal N(p)}g_k[q]
 +\sum_{s\in\mathcal S(p)}w_s\bigl(g_{k'_s}[q_s]+\sigma_{p,s}d_s\bigr),\\
\mathrm{den}_p &= \lambda_{\mathrm a}
 +\lambda_{\mathrm{sm}}|\mathcal N(p)|
 +\sum_{s\in\mathcal S(p)}w_s,\\
g_k^+[p] &=(1-\omega)\,g_k[p]+\omega\,\mathrm{num}_p/\mathrm{den}_p .
\end{split}
\end{equation}
Here $\mathrm{num}_p$ and $\mathrm{den}_p$ are the numerator and denominator
of the weighted average $\mathrm{num}_p/\mathrm{den}_p$; the anchor adds
$\lambda_{\mathrm a}$ to $\mathrm{den}_p$ and nothing to $\mathrm{num}_p$
because its target is $g=0$. Every frame starts
from $g=0$, and all cells are updated from the previous iterate (Jacobi).
We use $\lambda_{\mathrm{sm}}=30$, $\lambda_{\mathrm a}=0.02$,
$\omega=0.7$, and twelve iterations. The anchor makes the unprojected linear
system strictly diagonally dominant. After every update, the implementation first clamps each
value to $[-0.15,0.15]$ and then subtracts the mean over all camera cells.
Consequently the clamp is a bound before gauge removal, not a strict bound on
the final re-centered field. An optional reference-camera mode sets that
camera's grid to zero instead of removing the mean. Luminance uses one scalar
field; RGB mode solves three independent channels.

\paragraph{Temporal filtering and bake.}
Continuous capture uses the EMA specified above, with the gauge re-applied
after blending. On \textit{office2} frames $400$--$699$, it reduces the mean
frame-to-frame change of the (clamped) log-gain grid from $0.0079$ to
$0.0010$, and the $99$th percentile over cells and frames from $0.057$ to
$0.0072$; without it, single cells jump by several percent between frames. Filtering is in camera-grid coordinates and does not require
mesh correspondences across frames. For each atlas pixel, the band layout
identifies its camera and image coordinates. We interpolate that camera's
field bilinearly, clamp the interpolated log-gain again to $[-0.15,0.15]$,
and multiply the original linear RGB by $\exp(\hat g)$, where $\hat g$ denotes
this clamped value. The applied gain therefore remains in approximately
$[0.86,1.16]$, even if the gauge-shifted grid exceeds the update clamp. We then
convert back to sRGB. This leaves image detail in the original texture while
modulating it with a coarse spatial correction; it does not repair geometric
misalignment or sharp local seams.

\paragraph{Constraint and output seam residuals.}
\Cref{tab:seam} reports two seam residuals. The constraint residual
(Constr.) is computed inside the solver: the mean $|d_s|$ over the sampled seams before, and
\mbox{$|d_s-(g_{a(s)}[p_a^s]-g_{b(s)}[p_b^s])|$} after, using the EMA-blended grid at
the nearest cell,
without interpolation or the bake clamp. The output residual (Output) is
measured after the bake: it applies the same seam detection and sampling (\cref{fig:supp_inset}) to the atlas the
system writes (8-bit JPEG), so it includes the bilinear field interpolation,
the clamp, and quantization. ``Before'' reads the correction-off run and
``after'' the full run, each with its own mesh. Both residuals are measured
between the input cameras on the evaluated frames.

\paragraph{GPU reuse.}
Faces, normals, camera labels, and projected coordinates remain on the device
from reconstruction and texturing. The correction also reuses the existing
fixed-size-array adjacency, avoiding another edge-key sort (about
\SI{1.7}{\milli\second} per frame). It needs no new camera
projection and no geometry re-upload. Proxy construction, residual sampling,
field updates, and atlas bake execute on the GPU. Persistent buffers avoid
per-frame allocations, and the iterations require no CPU synchronization.

\section{Held-out Re-rendering Protocol}
\label{sec:supp_protocol}

We reconstruct the textured mesh from one set of Kinects and render it from a
\emph{disjoint} set, so the evaluation views are never used as input.
Concretely, we reconstruct from the even Kinects $\{2,4,6,8,10\}$ and evaluate on
the odd Kinects $\{1,3,5,7,9\}$, at a voxel size of \SI{0.02}{\meter}
(\texttt{cmu-room}). For each held-out camera we render the reconstructed mesh
with its calibrated intrinsics/extrinsics (Blender/EEVEE, emission shading) and
compare to that camera's real, undistorted RGB image.

\paragraph{Shared valid-pixel mask.}
Each method leaves different uncovered (black) pixels where faces are unseen or
removed. Scoring each method only on its own covered pixels would compare
different pixel sets and make the numbers incomparable. We therefore evaluate
every method on the \emph{intersection} of all methods' covered pixels: a pixel
is valid only if every method renders non-black there (a light morphological
opening removes speckle). PSNR, SSIM, and LPIPS are computed on this single
shared mask and averaged with a $5\%$ trimmed mean. PSNR uses only the
masked pixels. SSIM and LPIPS (AlexNet) are computed on the full frame after
every pixel outside the mask is set to the same constant in both images
(black for SSIM and mid-gray for LPIPS, whose input range is $[-1,1]$). All
metrics use sRGB values in $[0,1]$.

\paragraph{Subsets.}
The quality numbers of \cref{sec:exp_quality} and the luminance-vs-RGB ablation use frames $400$--$697$ at stride $3$ ($100$ evaluated frames) per sequence and the same common mask over all eight evaluated variants. The offline-texturing comparison (\cref{sec:supp_offtex}) uses frames $400$--$429$; the voxel-size and occlusion ablations use the $50$-frame subsets specified in their captions. Within each comparison, all methods are evaluated on the same frames and mask.
For the quality and seam numbers of \cref{sec:experiments} and the luminance-vs-RGB ablation, our method reconstructs every frame from $400$ to $699$ in order, and only every third frame is rendered and evaluated. The correction-on variants use the gain-field EMA ($\alpha=0.15$) throughout, as do the runtimes in \cref{tab:runtime_breakdown}; its history starts at frame $400$ (the first field is used as is) without a separate warm-up.

\paragraph{Rounding.}
Differences, ratios, and percentages in the text and in the $\Delta$ columns
of this appendix are computed from unrounded metrics, so they can differ in
the last digit from values recomputed from the rounded table entries.

\section{Seam Correction Variants}
\label{sec:supp_rgb}

Our seam correction estimates a single \emph{luminance} gain field by default,
which preserves chroma and solves a third as many parameters as a per-channel
(RGB) field; the RGB field uses the same solver with three independent channels
and is meant for cameras whose white balance differs. We compare the two
variants on Panoptic, whose Kinects share their white balance, on a
synthetic capture with known per-camera differences, and on our in-house
capture, whose cameras run their own white balance.

\paragraph{Panoptic.}
\cref{tab:supp_rgb} compares the two fields on the same $100$-frame held-out set and common mask as \cref{tab:panoptic_quant}.
Correcting chroma in addition to luminance changes the re-rendering metrics
negligibly (PSNR within $0.002$, SSIM within $0.0002$, LPIPS within $0.0005$ on
both sequences in \cref{tab:supp_rgb}): the inter-camera differences are dominated by exposure, and
the luminance field is sufficient.

\paragraph{Spatial field vs.\ one gain per camera.}
To test whether the spatial field is needed, we also solve a single luminance
gain per camera (a $1\times1$ field) with the same solver, gain bound, and
temporal filter (``constant'' in \cref{tab:supp_rgb}). It reaches only
$72\%$ / $35\%$ of the PSNR gain, $54\%$ / $30\%$ of the SSIM gain, and
$50\%$ / $50\%$ of the LPIPS gain of the $24\times16$ field over no correction
(\textit{office2} / \textit{band2}). Per image pair, the field scores higher
than the constant gain in SSIM and LPIPS on $499$--$500$ of the $500$ pairs of
each sequence and in PSNR on $418$ and $370$. The seam constraint residual
falls by $13.8\%$ / $7.8\%$ with the constant gain against $36.5\%$ /
$31.8\%$ with the field (\cref{tab:seam}). The variation of the gain within each camera
therefore matters beyond a per-camera exposure offset.

\begin{table}[t]
\centering
\footnotesize
\setlength{\tabcolsep}{4pt}
\renewcommand{\arraystretch}{1.05}
\caption{Luminance-only vs.\ per-channel RGB seam correction (held-out,
\SI{0.02}{\meter}, $100$ frames, common mask over all eight evaluated variants). ``off'' is no seam correction; ``constant'' solves one luminance gain per camera ($1\times1$ field) and is scored on the same masks.}
\label{tab:supp_rgb}
\begin{tabular}{@{}l ccc ccc@{}}
\toprule
& \multicolumn{3}{c}{\textbf{office2}} & \multicolumn{3}{c}{\textbf{band2}} \\
\cmidrule(lr){2-4}\cmidrule(lr){5-7}
Variant & PSNR$\uparrow$ & SSIM$\uparrow$ & LPIPS$\downarrow$
        & PSNR$\uparrow$ & SSIM$\uparrow$ & LPIPS$\downarrow$ \\
\midrule
off & 15.869 & 0.7714 & 0.2988 & 17.115 & 0.7960 & 0.2773 \\
constant ($1\times1$) & 15.946 & 0.7786 & 0.2932 & 17.136 & 0.7984 & 0.2737 \\
luminance & 15.976 & 0.7847 & 0.2876 & 17.174 & 0.8038 & 0.2701 \\
RGB & 15.977 & 0.7847 & 0.2873 & 17.176 & 0.8039 & 0.2696 \\
\bottomrule
\end{tabular}
\end{table}

\paragraph{Synthetic capture with ground truth.}
Our in-house capture cannot be released, and neither it nor Panoptic has
images free of inter-camera color differences. We therefore render a
synthetic stand-in in Blender, whose generation scripts are released with the code: a person animated with
CMU motion capture~\cite{CMUMocap} dances for $150$ frames on a
$3\times3$\,m floor, seen by ten $1280\times720$ RGB-D cameras on a circle,
with an empty background. All materials are Lambertian and the lighting is
fixed, so a surface point has the same color in every camera. We then
multiply each camera's linear color by a fixed smooth gain: an exposure
offset (log) of $+0.10$, $-0.10$, $+0.10$, $-0.10$, and $0$ in order around
the circle, assigned separately to the five reconstruction and the five
held-out cameras, so most neighboring cameras of a group differ by about
$22\%$, plus a small linear gradient and a
small white-balance shift; every gain stays within $\pm0.14$ per channel and
is zero-mean over the reconstruction cameras. This is the ideal case for our
model: the differences are multiplicative, low-frequency, and within the
default bound. We reconstruct the foreground at \SI{0.01}{\meter} from the
five even cameras and evaluate on the five odd ones against the unperturbed
renders on all $150$ frames and the shared mask. As for the in-house capture,
the correction uses $|g|\le0.4$ and $\lambda_{\mathrm{sm}}=5$, and the seams
are sampled on a $1/2$ instead of a $1/8$ proxy. This is a limitation of the
default: the $1/8$ proxy suits scenes that fill the image, as in Panoptic,
but here the person covers only a small part of each image, the proxy boxes
at its seams mix in background and other body parts, and the gains solved on
the body do not follow the applied ones. A small subject needs a finer proxy.

\cref{tab:supp_synth} shows that the correction recovers most of the loss
caused by the gains: $94\%$ (luminance) and $97\%$ (RGB) of the SSIM gap and
$87\%$ / $95\%$ of the LPIPS gap to the reconstruction from unperturbed input,
and $57\%$ / $59\%$ of the \SI{0.34}{dB} PSNR gap. The seam constraint residual (\cref{tab:seam}) drops
from $0.091$ to $0.049$, close to the mismatch on the unperturbed input
($0.048$), which reflects geometric and sampling error alone. \cref{fig:supp_synth}(a) shows
seams on the floor and on the shirt: without correction the surface breaks
into brighter and darker patches where the texture switches between cameras;
with either field the patches disappear.

We compare the applied log-gain $h$ with the recovered gain $-g$, the
negated correction field. The correction is relative, so $-g$ is not
expected to equal $h$. A seam constrains only the difference between the two
cameras that meet there: across the seams, the recovered difference matches
the applied one ($r=0.97$, slope $0.94$), which is what removes the seams.
Each camera's own level is not observed; it is set by the zero-mean gauge
and, away from the seams, by the smoothness term and the anchor toward unit
gain. The per-camera gain at the same seams therefore follows the applied one
with less than half its amplitude ($r=0.83$, slope $0.41$;
\cref{fig:supp_synth}(b)), and a gain common to all cameras would not be
recovered at all.

\begin{table}[t]
\centering
\footnotesize
\setlength{\tabcolsep}{4pt}
\renewcommand{\arraystretch}{1.05}
\caption{Synthetic capture (held-out cameras, \SI{0.01}{\meter}, $150$
frames, shared mask). ``clean input'': reconstruction from the unperturbed
renders, the upper bound.}
\label{tab:supp_synth}
\begin{tabular}{@{}l ccc@{}}
\toprule
Variant & PSNR$\uparrow$ & SSIM$\uparrow$ & LPIPS$\downarrow$ \\
\midrule
off         & 22.645 & 0.9404 & 0.1024 \\
luminance   & 22.842 & 0.9509 & 0.0948 \\
RGB         & 22.847 & 0.9513 & 0.0941 \\
\midrule
clean input & 22.988 & 0.9516 & 0.0937 \\
\bottomrule
\end{tabular}
\end{table}

\begin{figure*}[t]
\centering
\footnotesize
\setlength{\tabcolsep}{1pt}
\begin{tabular}{@{}c@{\hskip 4pt}cccc@{}}
  \includegraphics[width=0.2\linewidth]{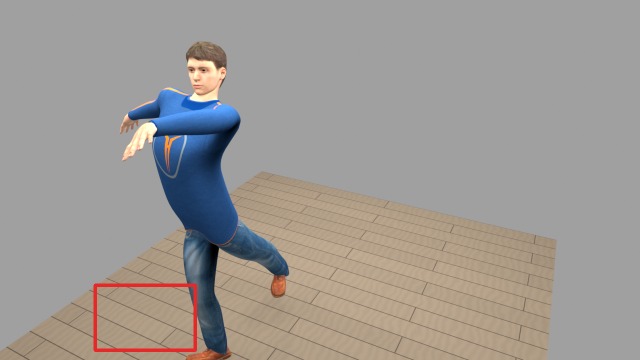} &
  \includegraphics[width=0.17\linewidth]{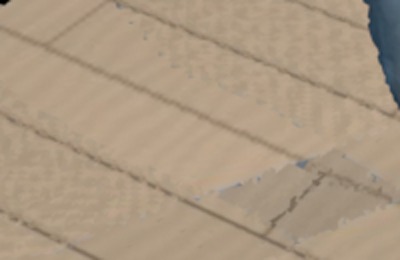} &
  \includegraphics[width=0.17\linewidth]{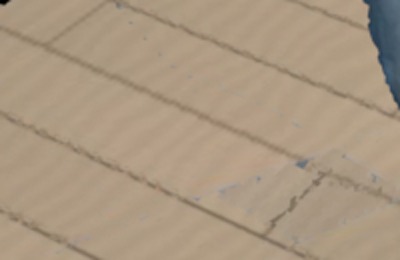} &
  \includegraphics[width=0.17\linewidth]{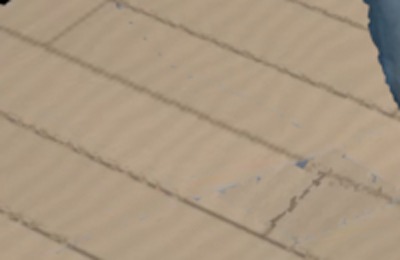} &
  \includegraphics[width=0.17\linewidth]{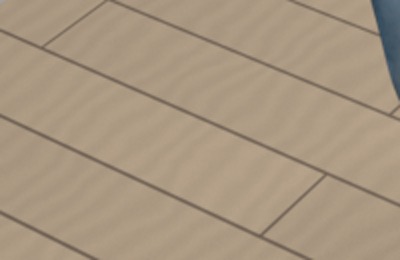} \\
  \includegraphics[width=0.2\linewidth]{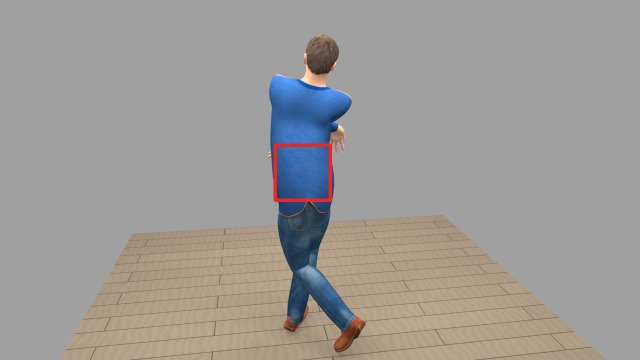} &
  \includegraphics[width=0.17\linewidth]{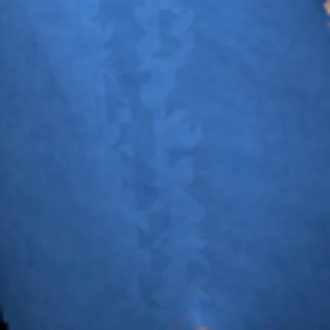} &
  \includegraphics[width=0.17\linewidth]{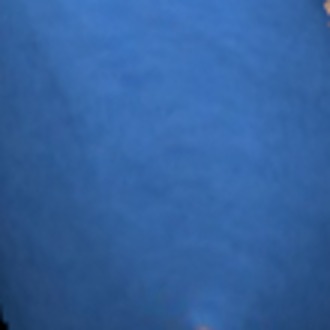} &
  \includegraphics[width=0.17\linewidth]{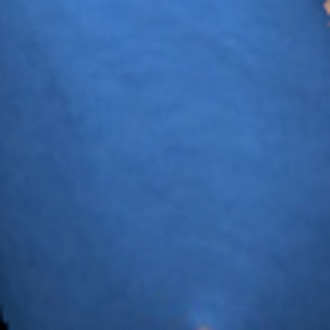} &
  \includegraphics[width=0.17\linewidth]{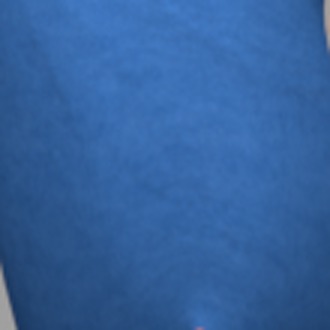} \\
  held-out view (box) & w/o & luminance & RGB & ground truth \\
\end{tabular}
\\[1pt]
(a) Floor (held-out camera 9, frame 135) and shirt (camera 3, frame 70), enlarged
\\[3pt]
\includegraphics[width=0.8\linewidth]{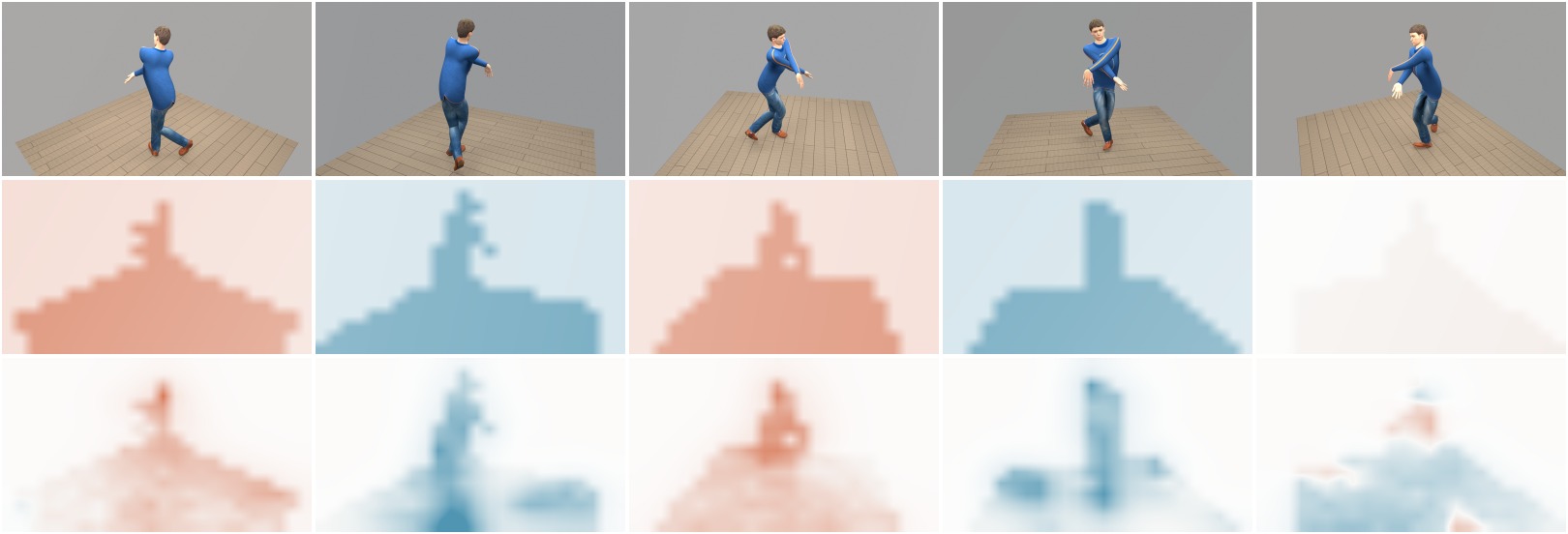}
\\[1pt]
(b) Reconstruction cameras 2, 4, 6, 8, 10 (columns): input, applied exposure
gain $h$, and the recovered gain $-g$ (negated luminance correction field; red
brighter, blue darker, $\pm0.15$; faded: no visible surface)
\caption{Synthetic capture. \textbf{(a)} Without correction the floor and the
shirt split into brighter and darker patches where the texture switches
between cameras; both fields remove them. \textbf{(b)} The correction is
relative: it reproduces the differences between cameras across the seams,
not each camera's applied gain, which is why the two rows differ in
amplitude.}
\label{fig:supp_synth}
\end{figure*}

\begin{figure*}[t]
\centering
\footnotesize
\setlength{\tabcolsep}{1.5pt}
\renewcommand{\arraystretch}{0.9}
\begin{tabular}{@{}cccc@{\hskip 8pt}c@{}}
  \includegraphics[width=0.15\linewidth]{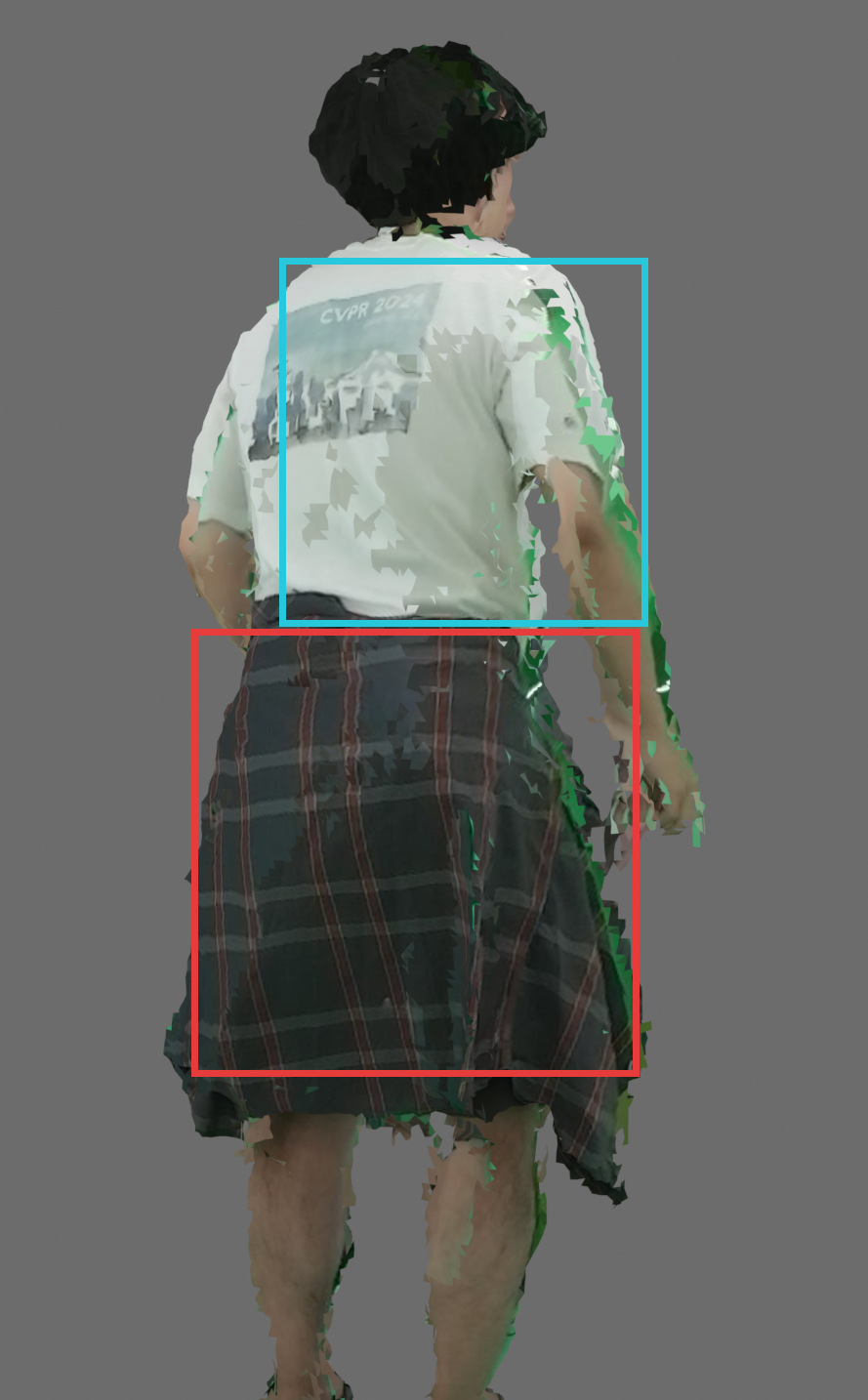} &
  \includegraphics[width=0.15\linewidth]{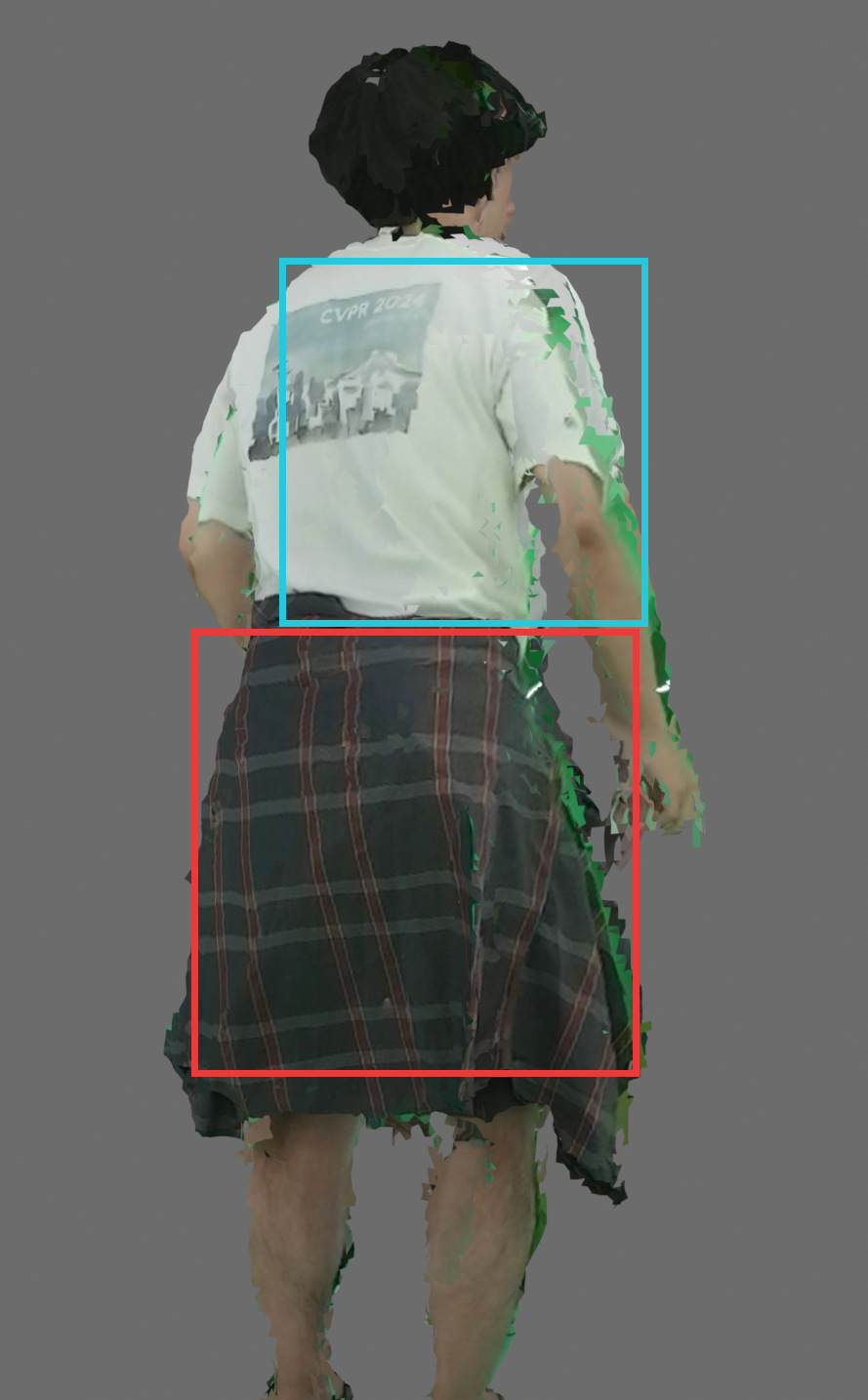} &
  \includegraphics[width=0.15\linewidth]{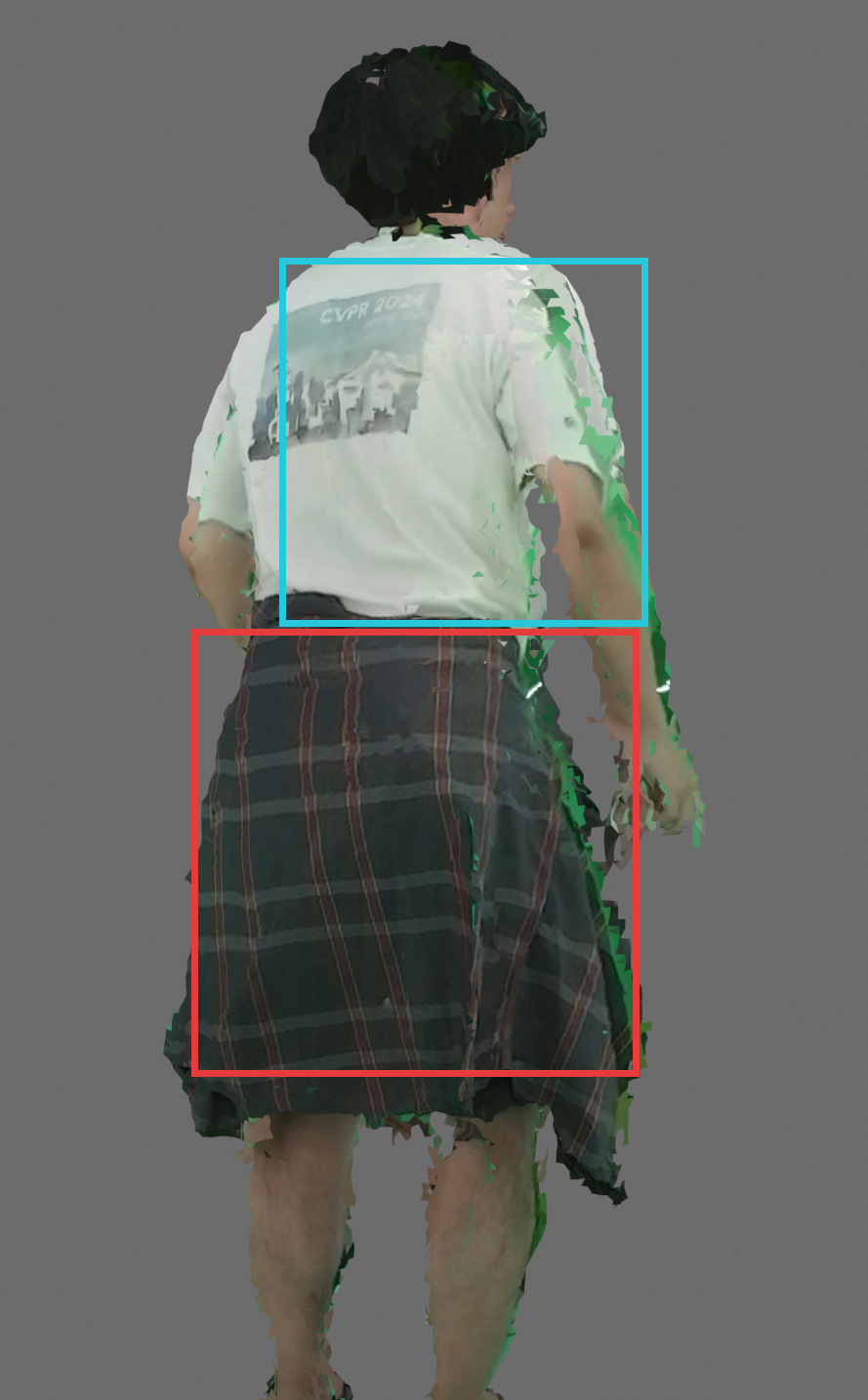} &
  \includegraphics[width=0.15\linewidth]{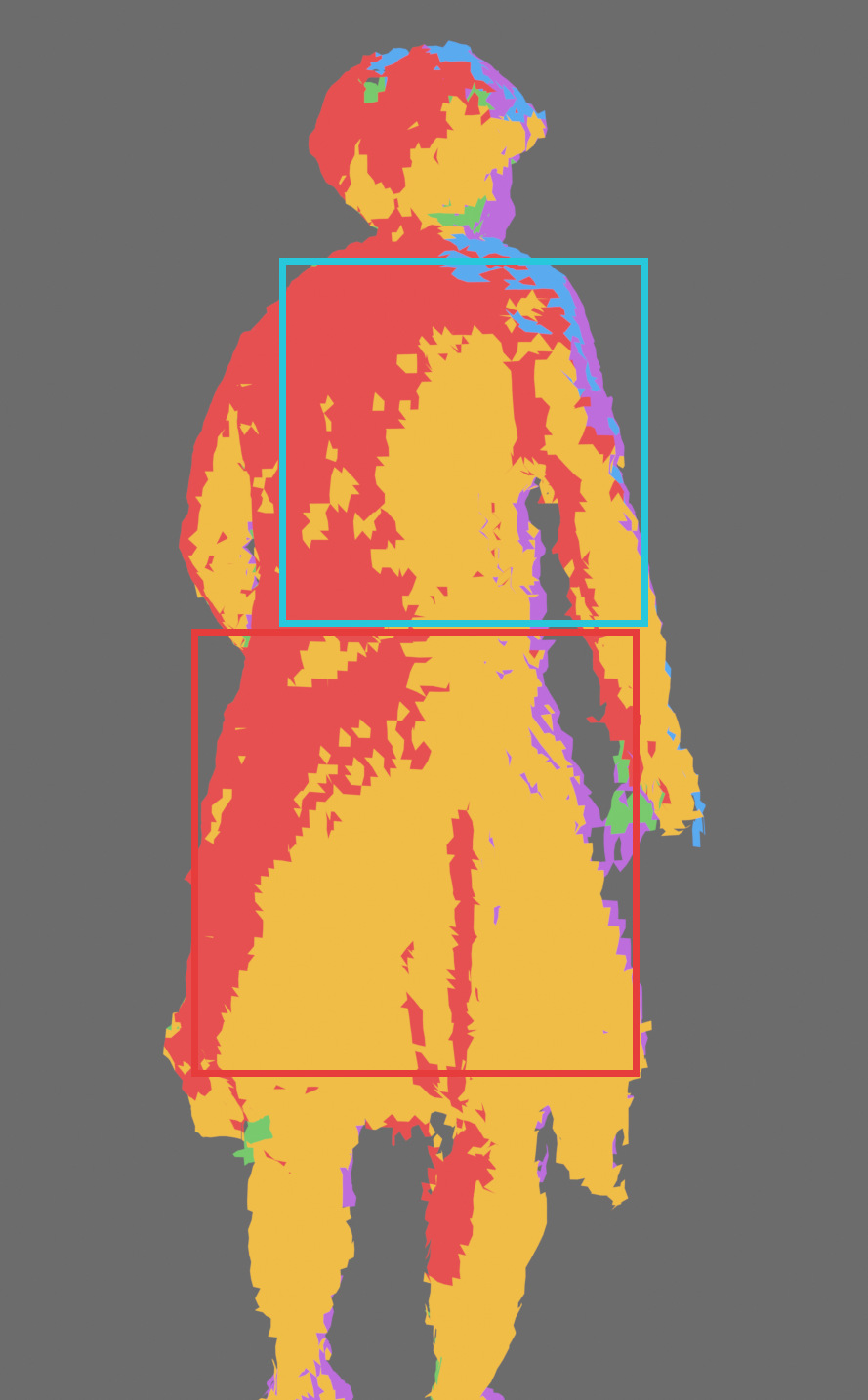} &
  \begin{tabular}[b]{@{}ccc@{}}
    \includegraphics[width=0.105\linewidth]{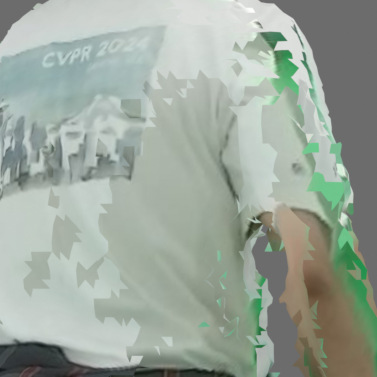} &
    \includegraphics[width=0.105\linewidth]{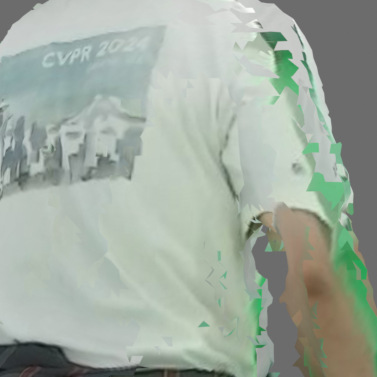} &
    \includegraphics[width=0.105\linewidth]{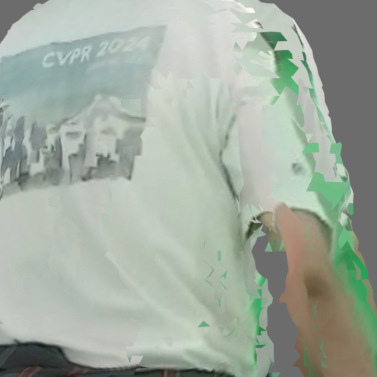} \\
    \multicolumn{3}{c}{T-shirt (cyan): w/o, luminance, RGB} \\
    \includegraphics[width=0.105\linewidth]{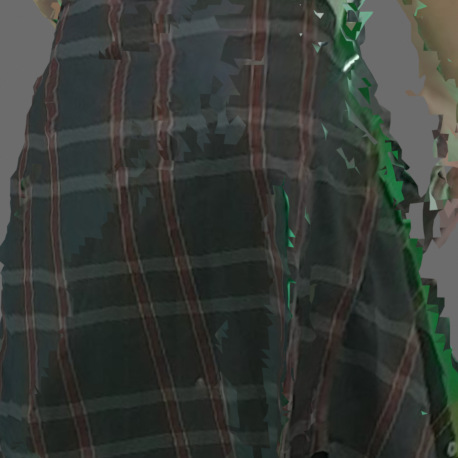} &
    \includegraphics[width=0.105\linewidth]{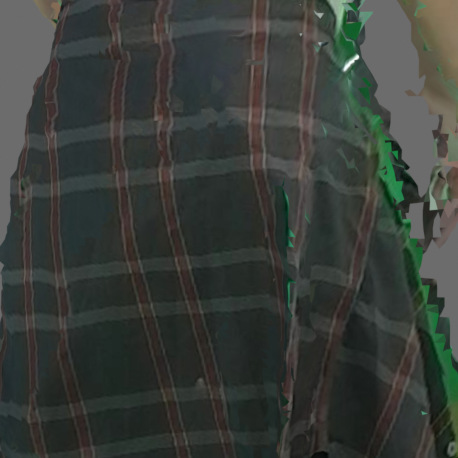} &
    \includegraphics[width=0.105\linewidth]{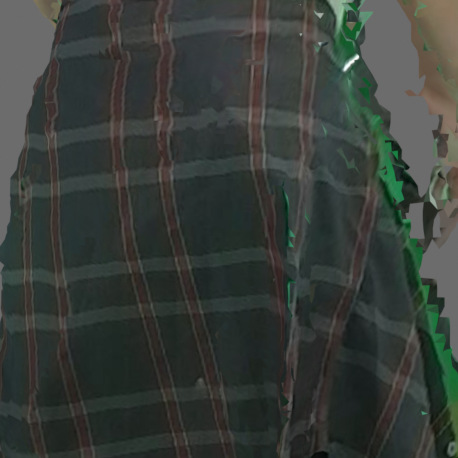} \\
    \multicolumn{3}{c}{waist (red): w/o, luminance, RGB} \\
  \end{tabular} \\
  w/o seam corr. & luminance & RGB & source camera & \\
\end{tabular}
\\[1pt]
(a) Full view without correction, with the luminance field, with the RGB field,
and the source-camera mosaic (1 color $=$ 1 input view; cyan box: T-shirt; red
box: waist); (b) zoom-ins of the two boxed regions in the same order
\\[3pt]
\includegraphics[width=0.85\linewidth]{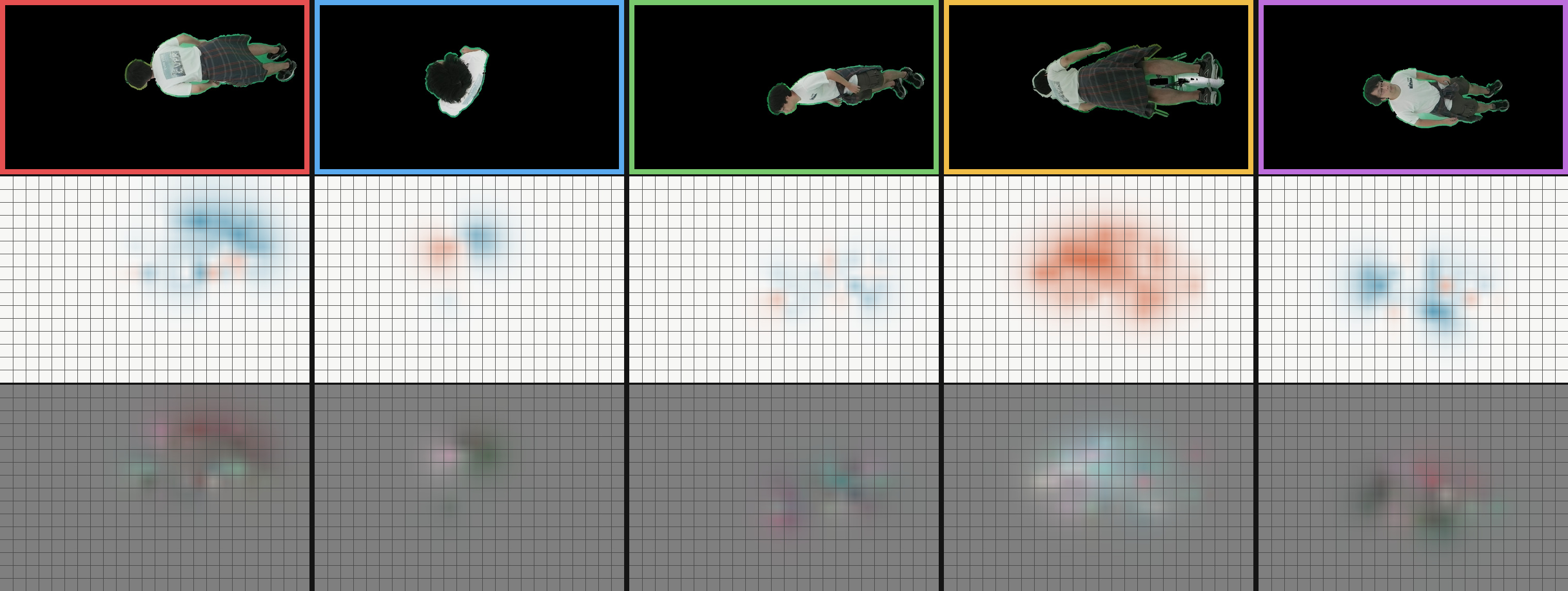}
\\[1pt]
(c) Each input camera (border color as in the mosaic): input view (top), the
luminance gain $\exp(g_k)$ on the $24\times16$ grid (middle; red brightens,
blue darkens, as in \cref{fig:colorcorr}), and the RGB gain shown as the color it
gives a neutral gray (bottom; $0.5\exp(g_k)$ on sRGB values, exaggerating the tint)
\caption{Luminance vs.\ RGB seam correction on the in-house five-camera capture
(turntable view from behind; both with $|g|\le0.4$ and
$\lambda_{\mathrm{sm}}=5$). \textbf{(a)} Without correction, with the
luminance-only field, with the per-channel RGB field, and the source-camera
mosaic: the red and yellow cameras texture most of the back, and their boundary
crosses the T-shirt (cyan box) and the plaid shirt at the waist (red box).
\textbf{(b)} Zoom-ins of the boxed regions: the darker, greenish T-shirt patch
of the yellow camera and the brightness step on the plaid shirt are equalized by
both variants. \textbf{(c)} Input views (background masked) and the solved gain
fields: both brighten the yellow camera by up to about $1.5\times$ toward the
center of its view, and the three channels of the RGB field nearly coincide.
The fields stay smooth and low-frequency and shift each camera's colors as a
whole.}
\label{fig:supp_own_seam}
\end{figure*}

\paragraph{In-house capture.}
\cref{fig:supp_own_seam} shows the foreground-only in-house capture of the main
paper (Sec.~4.4; five Orbbec Femto Bolt sensors), whose cameras run independent
auto exposure and auto white balance, so camera boundaries may differ in color
cast as well as in brightness. We run both variants with a wider gain bound and
weaker smoothing than the Panoptic default ($|g|\le0.4$, i.e.\ gains in
$[0.67,1.49]$, and $\lambda_{\mathrm{sm}}=5$); the $24\times16$ grid, the
solver, and all other seam-correction parameters are as in \cref{tab:supp_hyper}.
The foreground-only reconstruction uses \SI{0.01}{\meter} voxels in a
\SI{2}{\meter} cube, a \SI{0.10}{\meter} depth tolerance, a four-voxel
truncation band, a $100$-face component threshold, and at most $20$ inpainting
iterations. The view is a
turntable render of one frame seen from behind; no held-out camera exists for
this capture, so the comparison is qualitative. The sequence was recorded by
the authors; the person shown took part voluntarily and gave informed consent
to being recorded and to the use of the recordings in this paper and its
supplementary video.

Two cameras texture most of the back (red and yellow in the source-camera
mosaic), and their boundary crosses the T-shirt and the plaid shirt at the
waist. Without correction, the T-shirt region textured by the yellow camera is
darker and slightly greenish next to the bright white of the red camera, and
the plaid shirt steps in brightness across the boundary; accordingly, both
fields brighten the yellow camera by up to about $1.5\times$ toward the center
of its view. The luminance field removes the brightness steps as well as the
RGB field does: on the T-shirt fabric on both sides of the boundary, the difference of the
mean CIELAB lightness of the two sides drops from about $8$ to $3$ units with
either variant, while the distance between their mean $(a^*,b^*)$ stays at
$2$--$3$ units with and without correction, close to the visibility threshold.
(The fabric pixels are those in the T-shirt zoom box with $L^*>55$ in the
uncorrected render that lie within $25$ pixels of both cameras' regions of
the source-camera mosaic, in the $1440\times1440$ renders.) Consistently, the three channels
of the RGB field agree with their mean to within $1\%$ on average
(\cref{fig:supp_own_seam}(c)): even with independent white balance, the
exposure difference dominates on this capture, and the luminance default
suffices here too (the figure applies $0.5\exp(g_k)$ to sRGB rather than
linear values, exaggerating the tint). As on Panoptic, the fields are smooth and low-frequency;
they shift each camera's colors as a whole rather than re-texturing the
surface. The green fringes at the silhouette stem from background pixels mixed
into depth-edge faces of the input images and are present in all variants.

\section{Offline Texturing on Our Geometry}
\label{sec:supp_offtex}

The \emph{Offline} baseline of \cref{sec:experiments} pairs SPSR geometry with
mvs-texturing~\cite{Waechter2014Texturing}, so it differs from our method in both
geometry and texturing. To isolate the texturing step, we instead apply
mvs-texturing to \emph{our own} per-frame meshes and evaluate from held-out
cameras (\cref{tab:supp_offtex}). This auxiliary comparison uses a separate
mesh set (frames $400$--$429$ of each sequence) and a two-method coverage intersection; its scores support
comparisons within this experiment, rather than direct comparisons with \cref{tab:panoptic_quant}.

We verified the consistency of the offline texturing pipeline:
re-texturing the same SPSR meshes matches the reference renders to
$34$--$41$\,dB PSNR. mvs-texturing is run with hole filling enabled; we also tried both
settings of its keep-unseen-faces option.

In this auxiliary comparison, offline texturing on our geometry performs worse
than our texturing in all three metrics. Some faces unseen by the input views
remain untextured (rendered blank), while our inpainting supplies colors for
these faces. This result is specific to the dense, noisy per-frame
Marching-Cubes meshes tested here.

\begin{table}[t]
\centering
\footnotesize
\setlength{\tabcolsep}{4pt}
\renewcommand{\arraystretch}{1.05}
\caption{Offline texturing (mvs-texturing) applied to \emph{our} geometry vs.\
our real-time texturing, on identical meshes (held-out, \SI{0.02}{\meter},
frames $400$--$429$).}
\label{tab:supp_offtex}
\begin{tabular}{@{}l ccc ccc@{}}
\toprule
& \multicolumn{3}{c}{\textbf{office2}} & \multicolumn{3}{c}{\textbf{band2}} \\
\cmidrule(lr){2-4}\cmidrule(lr){5-7}
Texturing & PSNR$\uparrow$ & SSIM$\uparrow$ & LPIPS$\downarrow$
          & PSNR$\uparrow$ & SSIM$\uparrow$ & LPIPS$\downarrow$ \\
\midrule
Offline (mvs) & 13.303 & 0.674 & 0.390 & 14.605 & 0.704 & 0.360 \\
Ours & \textbf{15.290} & \textbf{0.761} & \textbf{0.311} & \textbf{16.535} & \textbf{0.782} & \textbf{0.289} \\
\bottomrule
\end{tabular}
\end{table}

\section{Voxel Size: Quality versus Runtime and Memory}
\label{sec:supp_voxel}

Section~\ref{sec:experiments} uses \SI{0.02}{\meter} voxels, the setting at which the full
pipeline stays well within the \SI{33}{\milli\second} real-time budget. Finer voxels
recover geometric detail that \SI{0.02}{\meter} loses (thin structures, small
objects, sharper silhouettes) but the dense TSDF grid grows cubically. To
quantify this tradeoff we run the identical pipeline on \textit{office2} at
\SI{0.02}{\meter}, \SI{0.01}{\meter}, and \SI{0.005}{\meter} (the
\texttt{cmu-room} volume of $5.8\times2.7\times5.8\,\mathrm{m}$, i.e.
$290\times135\times290$, $580\times270\times580$, and
$1160\times540\times1160$ voxels) and report held-out re-rendering quality,
per-stage runtime, and GPU memory (\cref{tab:supp_voxel}). All three settings
use the same pipeline and the same shared valid-pixel mask (the intersection of
the three runs' coverage), so the rows are directly comparable to each other.
The comparison in \cref{sec:experiments} includes other reconstruction methods and uses a
different frame subset and coverage intersection (\cref{sec:supp_protocol});
its absolute scores therefore should not be compared directly with this ablation.

\begin{table}[t]
\centering
\footnotesize
\setlength{\tabcolsep}{3pt}
\renewcommand{\arraystretch}{1.05}
\caption{Voxel size versus held-out re-rendering quality, runtime, and GPU
memory on \textit{office2} (same pipeline, shared mask, frames 400--547 at stride
3, cameras 1,3,5,7,9, \SI{10}{\centi\meter} depth tolerance). $\Delta$ columns are
relative to the \SI{0.02}{\meter} row, from unrounded values. Runtime is the per-frame processing
time (low-95\% mean) over 2000 frames (400--2399); the
\SI{0.02}{\meter} row uses the same five-input timing trace as \cref{tab:runtime_breakdown}. Memory is the
peak device memory over these 2000 frames (static grid and mesh buffers plus
per-frame allocations). Quality uses the 50-frame subset specified above.}
\label{tab:supp_voxel}
\resizebox{\linewidth}{!}{%
\begin{tabular}{@{}l ccc ccc c@{}}
\toprule
Voxel & PSNR$\uparrow$ & SSIM$\uparrow$ & LPIPS$\downarrow$ & $\Delta$PSNR & $\Delta$LPIPS & ms/frame & GPU mem \\
\midrule
\SI{0.02}{\meter}  & \textbf{15.821} & \textbf{0.795} & \textbf{0.271} & --     & --     & \textbf{13.1} & \textbf{2.7}~GB \\
\SI{0.01}{\meter}  & 15.704 & 0.768 & 0.308 & $-0.12$ & $+0.038$ & 34.6 & 5.7~GB \\
\SI{0.005}{\meter} & 14.571 & 0.625 & 0.445 & $-1.25$ & $+0.175$ & 138.9 & 24.1~GB \\
\bottomrule
\end{tabular}%
}
\end{table}

\begin{figure*}[t]
\centering
\setlength{\tabcolsep}{1.5pt}
\begin{tabular}{@{}cccc@{}}
  \includegraphics[width=0.245\linewidth]{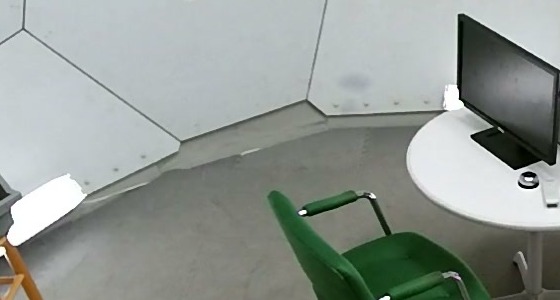} &
  \includegraphics[width=0.245\linewidth]{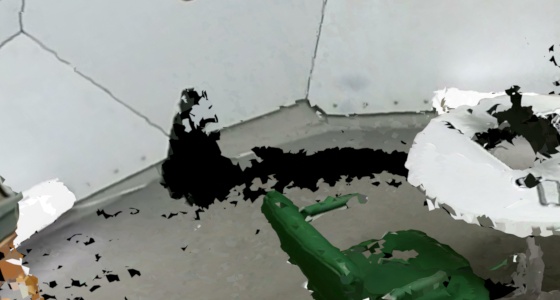} &
  \includegraphics[width=0.245\linewidth]{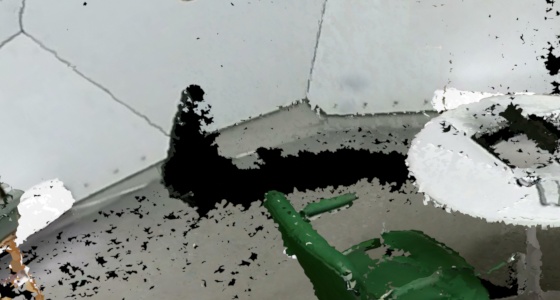} &
  \includegraphics[width=0.245\linewidth]{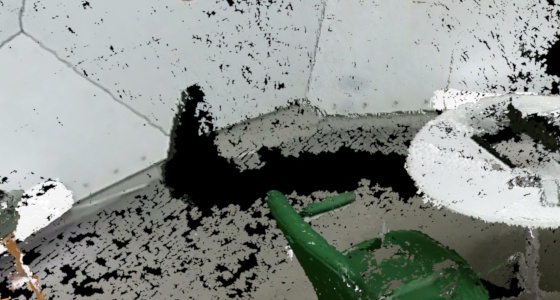} \\
  GT & \SI{0.02}{\meter} & \SI{0.01}{\meter} & \SI{0.005}{\meter} \\
\end{tabular}
\caption{Voxel size versus reconstruction detail on \textit{office2}, held-out
camera 7, frame 445 (crop). Finer voxels sharpen silhouettes (chair, table
rim) but leave more small holes and noise on the wall and floor, where
each voxel receives fewer depth samples; the truncation band also shrinks with
the voxel (four voxels: 8, 4, and \SI{2}{\centi\meter}), so voxel size is not
the only change. Overall the held-out metrics decrease (see text). Thin, dark
structures such as the stool remain missing at every setting because the depth
sensors do not observe them reliably.}
\label{fig:supp_voxel}
\end{figure*}

\paragraph{Quality.}
Halving the voxel size to \SI{0.01}{\meter} changes PSNR, SSIM, and LPIPS
by $-0.12$~dB, $-0.027$, and $+0.038$, respectively.
\cref{fig:supp_voxel} shows the sharper chair and table rim and the additional
holes on the wall and floor. At \SI{0.005}{\meter}, the changes relative to \SI{0.02}{\meter}
are $-1.25$~dB, $-0.170$, and $+0.175$. The
truncation band is four voxels, i.e. \SI{2}{\centi\meter}, at this fine setting.
On \textit{band2}, halving the voxel size changes the same metrics by
$-0.37$~dB, $-0.050$, and $+0.061$.
\SI{0.02}{\meter} remains the setting with a clear margin under the real-time
budget; \SI{0.01}{\meter} slightly exceeds it (see below).

\paragraph{Runtime and memory.}
Halving the voxel size multiplies the cell count by eight and the frame time
by $2.6$--$4.0\times$: \SI{13.1}{\milli\second} at \SI{0.02}{\meter},
\SI{34.6}{\milli\second} at \SI{0.01}{\meter}, and \SI{138.9}{\milli\second}
at \SI{0.005}{\meter} over the same 2000 frames (\textit{band2}: \SI{12.8}{} and
\SI{35.3}{\milli\second} at \SI{0.02}{} and \SI{0.01}{\meter}). The stages
scale differently. TSDF fusion and Marching Cubes grow with the cell count
($0.4\rightarrow2.3\rightarrow15.6$ and $0.5\rightarrow1.9\rightarrow10.8$~ms);
noise removal, texture mapping, inpainting, and seam correction grow with the
output face count (about $0.6$M, $2.9$M, and $12.4$M faces per frame), and
the GPU$\rightarrow$CPU download of the finished mesh becomes the largest
single stage at fine voxels ($2.9\rightarrow11.3\rightarrow45.6$~ms) because
the output itself is about \SI{0.85}{GB} per frame at \SI{0.005}{\meter}. The
dense grid, edge table, and mesh buffers take \SI{0.4}{}, \SI{2.3}{}, and
\SI{16.6}{GB} of static device memory at the three sizes (peak \SI{2.7}{},
\SI{5.7}{}, and \SI{24.1}{GB} including per-frame allocations and the input
buffers), so \SI{0.005}{\meter} needs a 24\,GiB-class GPU and, at about
\SI{7.2}{fps}, is an offline setting on this hardware, while \SI{0.01}{\meter}
fits an \SI{8}{GB} GPU at about \SI{28.9}{fps}, just short of the real-time
budget (\SI{32.4}{\milli\second} without the input upload and preprocessing).
Marching-cubes edge indices exceed
$2^{31}$ at \SI{0.005}{\meter} on this volume, so that build uses 64-bit edge
keys (a compile-time option; the default build is unchanged).

\section{Occlusion-Aware View Selection}
\label{sec:supp_occl}

The per-face view selection (\cref{sec:texture}) tests all three
vertices of a face against the chosen camera's depth image and rejects the
camera when the depth of a vertex differs from the observed depth at its
projection by more than a threshold $\tau_{\mathrm{occ}}$, in either
direction. We use $\tau_{\mathrm{occ}}=\SI{10}{\centi\meter}$ and compare it
with a loose \SI{1.1}{\meter} tolerance. The loose test admits large depth
discrepancies, allowing foreground texture to be projected onto occluded
background surfaces. The fixed \SI{10}{\centi\meter} tolerance corresponds
to five voxels at \SI{0.02}{\meter}. Rejected assignments leave faces
without a camera, which the inpainting stage fills with flat color.
On the common held-out mask, the default changes PSNR, SSIM, and LPIPS by
$+0.67$~dB, $+0.028$, and $-0.022$ on \textit{office2},
and $+0.41$~dB, $+0.028$, and $-0.021$ on \textit{band2} (\cref{tab:supp_occl}).
\cref{fig:supp_occl} shows the effect and \cref{tab:supp_occl} reports the
held-out metrics of both settings on the same build and mask.

\begin{table}[t]
\centering
\footnotesize
\setlength{\tabcolsep}{4pt}
\renewcommand{\arraystretch}{1.05}
\caption{Occlusion threshold on \textit{office2} and \textit{band2} (same build, shared mask per scene,
frames 400--547 at stride 3, cameras 1,3,5,7,9). Rejecting occluded
projections removes the ghost texture and improves all three held-out
metrics; the seam residual columns are the seam constraint residual of \cref{tab:seam} before and after seam correction.}
\label{tab:supp_occl}
\resizebox{\linewidth}{!}{%
\begin{tabular}{@{}l ccc c@{}}
\toprule
$\tau_{\mathrm{occ}}$ & PSNR$\uparrow$ & SSIM$\uparrow$ & LPIPS$\downarrow$ & seam res.\ (before$\rightarrow$after) \\
\midrule
\multicolumn{5}{@{}l}{\textit{office2}} \\
\SI{1.1}{\meter} (loose) & 14.882 & 0.737 & 0.327 & 0.328$\rightarrow$0.271 \\
\SI{10}{\centi\meter} (default) & \textbf{15.556} & \textbf{0.765} & \textbf{0.304} & 0.158$\rightarrow$0.104 \\
\midrule
\multicolumn{5}{@{}l}{\textit{band2}} \\
\SI{1.1}{\meter} (loose) & 16.110 & 0.753 & 0.310 & 0.287$\rightarrow$0.240 \\
\SI{10}{\centi\meter} (default) & \textbf{16.521} & \textbf{0.781} & \textbf{0.288} & 0.152$\rightarrow$0.104 \\
\bottomrule
\end{tabular}%
}
\end{table}

\begin{figure}[t]
\centering
\setlength{\tabcolsep}{1.5pt}
\begin{tabular}{@{}ccc@{}}
  \includegraphics[width=0.325\linewidth]{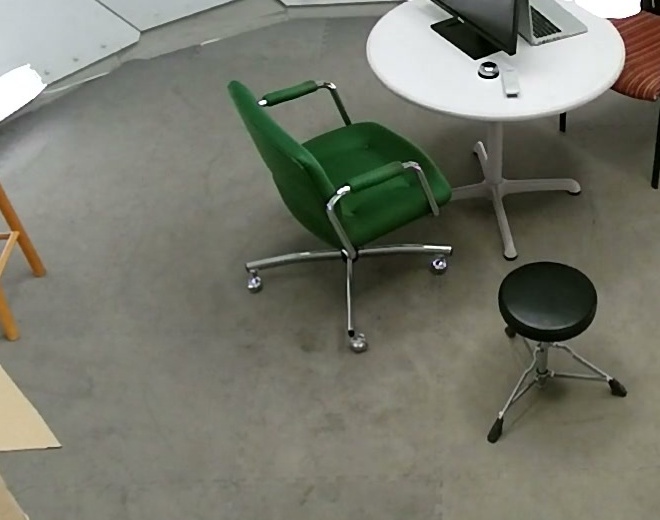} &
  \includegraphics[width=0.325\linewidth]{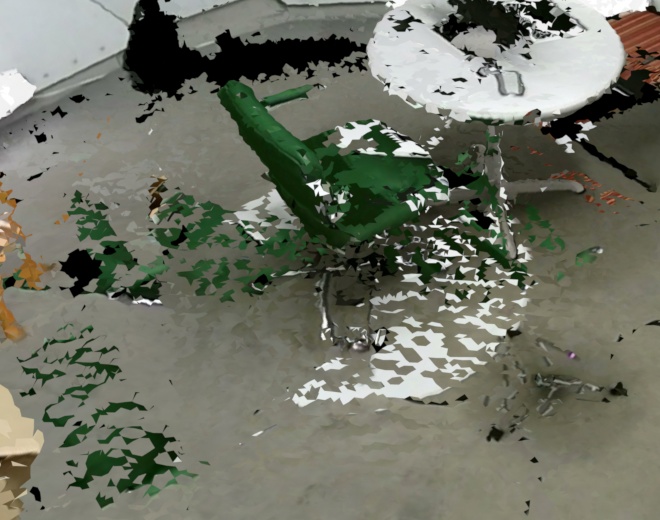} &
  \includegraphics[width=0.325\linewidth]{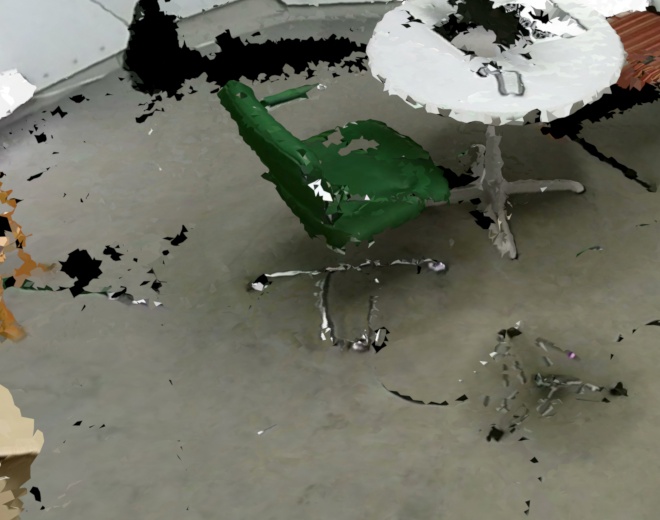} \\
  GT & $\tau_{\mathrm{occ}}=\SI{1.1}{\meter}$ & $\tau_{\mathrm{occ}}=\SI{10}{\centi\meter}$ \\
\end{tabular}
\caption{Occlusion-aware view selection on \textit{office2} (held-out camera 7,
frame 405, crop). With the loose \SI{1.1}{\meter} tolerance, the green chair's
texture spills onto the floor and the table's white texture onto the chair.
The \SI{10}{\centi\meter} test rejects these occluded projections; inpainting
fills the affected faces. Both use the same reconstruction settings.}
\label{fig:supp_occl}
\end{figure}

\section{Application to SfM+MVS}
\label{sec:supp_sfm}

\begin{figure}[t]
\centering
\begin{subfigure}[t]{0.47\linewidth}
\centering
\includegraphics[width=\linewidth]{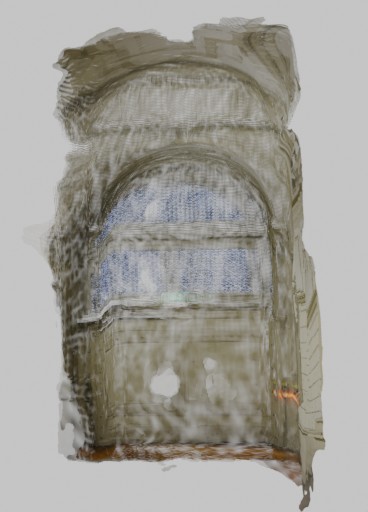}
\caption{\footnotesize VGGT point cloud}
\end{subfigure}
\hfill
\begin{subfigure}[t]{0.47\linewidth}
\centering
\includegraphics[width=\linewidth]{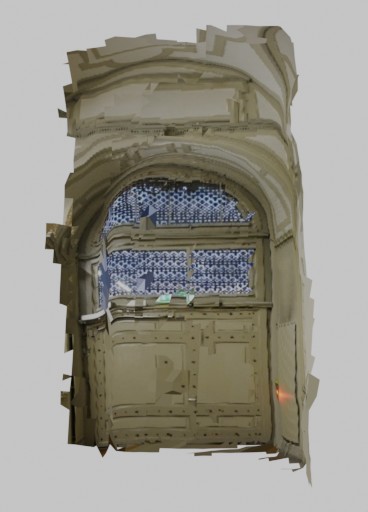}
\caption{\footnotesize Textured mesh}
\end{subfigure}
\caption{Application with SfM+MVS-style input. \textbf{(a)} Point-cloud output of
VGGT~\cite{wang2025vggt}. \textbf{(b)} Textured mesh obtained by applying our
method to the VGGT-derived geometry and camera views.}
\label{fig:supp_vggt}
\end{figure}

Our method can also texture surfaces reconstructed from image-based depth and
camera estimates rather than RGB-D sensors. \cref{fig:supp_vggt} shows an example
using the \textit{door} sequence of ETH3D~\cite{schoeps2017cvpr}: we take the
point-cloud output of VGGT~\cite{wang2025vggt} and apply our textured-meshing
pipeline to the derived geometry and camera views. This example is qualitative,
but it illustrates that the texturing stage is not tied to Kinect-style RGB-D
input.

\section{Use of Large Language Models}
\label{sec:supp_llm}

In line with the WACV 2027 policy on large language models (LLMs), we
disclose that an LLM-based coding assistant was used in two ways.
It helped polish the English of this paper, and it facilitated the experiments: it assisted in writing and
running the scripts that drive the reconstruction and evaluation campaigns,
compute the metrics, generate the plots and the supplementary video, and
check the reported numbers against the measured data, as well as in
refactoring the code for release. The method, its CUDA implementation, the
experimental design, and all claims are the author's; every statement and
number was verified by the author, who takes full responsibility for the
content.

\end{document}